\documentclass[11pt]{article}

\usepackage[final]{acl}

\usepackage{times}
\usepackage{latexsym}

\usepackage[T1]{fontenc}
\usepackage[utf8]{inputenc}

\usepackage{tabularx}

\usepackage{microtype}

\usepackage{inconsolata}

\usepackage{graphicx}

\usepackage{booktabs}

\usepackage{amssymb}

\usepackage{amsmath}

\usepackage{longtable}

\usepackage{url}

\usepackage{listings}
\lstdefinestyle{prompt}{
  basicstyle=\footnotesize\ttfamily,
  breaklines=true,
  breakatwhitespace=false,
  frame=single,
  framerule=0.3pt,
  xleftmargin=6pt,
  xrightmargin=6pt,
  aboveskip=4pt,
  belowskip=4pt,
  columns=flexible,
  keepspaces=true,
}

\lstdefinestyle{promptinner}{%
  basicstyle=\footnotesize\ttfamily,
  breaklines=true,
  breakatwhitespace=false,
  frame=none,
  xleftmargin=0pt,
  xrightmargin=0pt,
  aboveskip=0pt,
  belowskip=0pt,
  columns=flexible,
  keepspaces=true,
  showstringspaces=false,
}

\usepackage{tcolorbox}
\tcbuselibrary{skins, breakable}

\definecolor{promptblue}{RGB}{55, 90, 145}
\definecolor{promptbg}{RGB}{240, 245, 255}

\newtcolorbox{promptbox}[2][]{%
  title={#2},
  colback=promptbg,
  colframe=promptblue,
  colbacktitle=promptblue,
  coltitle=white,
  fonttitle=\bfseries\small,
  boxrule=0.8pt,
  arc=3pt,
  breakable,
  enhanced,
  left=4pt, right=4pt, top=3pt, bottom=3pt,
  #1
}

\title{Structured but Silent:\\ Probing Capability Requirements in LLM Hidden States}

\author{
  \textbf{Kyojun Choo,
  Minsoo Song,
  Yunju Kang,
  Chanjun Park\textsuperscript{$\dagger$}} \\
  Soongsil University \\
  \texttt{\{kjun5644, ecoses042, gardengnosis\}@soongsil.ac.kr} \\
  \texttt{chanjun.park@ssu.ac.kr}
}

\begin{document}
\maketitle

\begingroup
\renewcommand{\thefootnote}{\fnsymbol{footnote}}
\footnotetext[2]{Corresponding author.}
\endgroup

\begin{abstract}
Reliable tool use requires more than triggering a mechanism or matching a query to an API description. Before selecting a specific tool, an agent must first infer the capability requirements implied by the user query. In this paper, we investigate whether these query-side capability requirements are linearly decodable from LLM hidden representations prior to generation, and how this hidden-state accessibility compares with explicit verbal classification. We introduce \textbf{TACIT}, a framework that decomposes external requirements along three fundamental axes: Source, Transformation, and World Effect, defining eight structurally distinct capability classes. Using 1,600 balanced training queries from benchmarks, synthetic examples, and new domain scenarios, we train linear probes on pre-generation hidden states from four open-weight LLM families. Our empirical results demonstrate that fine-grained capability structures are linearly decodable with high accuracy across all models. Crucially, however, we expose a representation-to-verbalization gap: these same models are significantly less reliable when asked to explicitly classify the same queries in natural language. This disconnect indicates that information about required external capabilities is linearly accessible in LLM hidden representations but not reliably expressed—a phenomenon we define as \textit{structured but silent}.
\end{abstract}
 
\section{Introduction}

Large language models (LLMs) are increasingly deployed as autonomous agents capable of utilizing external tools \cite{qin2024toolllm, wang2024toolsanywaysurveylanguage}. While current paradigms focus heavily on whether to use a tool and which concrete tool to call. They overlook a more fundamental prerequisite that the model must first infer the underlying capability requirement implied by a user query. This query-side inference serves as an intermediate abstraction layer; for instance, distinguishing whether a query requires external retrieval, computation, or state-changing action is distinct from merely matching a query to a specific tool description. 

Despite its importance, existing methods optimize tool selection superficially via prompt engineering \cite{Alazraki_2025}, tool description tuning \cite{hsieh2023tooldocumentationenableszeroshot}, or output filtering \cite{yakovlev2024toolkenimprovingllmtool}. These behavior-centric approaches treat the model as a black box and fail to investigate whether fine-grained capability structures are encoded within the model's internal representations prior to generation. While recent representation probing reveals that hidden states capture latent attributes like truthfulness or factual knowledge \cite{marks2024geometrytruthemergentlinear, orgad2025llmsknowshowintrinsic}, internal analysis for tool use remains strictly limited to binary judgments, such as predicting tool triggering \cite{huang2024metatool,li2025adaptive} or detection of tool hallucinations \cite{healy2026internalrepresentationsindicatorshallucinations}. Whether the multidimensional capability requirements of tool-augmented tasks are encoded in LLM hidden states prior to generation remains an open question.

In this work, we investigate whether the external capability requirements of user queries are linearly decodable from LLM hidden representations before generation, and compare probe predictions with explicit verbal classifications of the same requirements. To formalize this question, we introduce \textbf{TACIT}, a three-axis capability probing framework that decomposes query requirements into \textbf{Source} (available vs. external retrieval), \textbf{Transformation} (direct retrieval vs. transform), and \textbf{World Effect} (read-only vs. state-changing action), defining eight structurally distinct capability classes.

Using 1,600 balanced training queries from benchmarks, synthetic examples, and new domain scenarios, we train linear probes on pre-generation hidden states from four open-weight LLM families: Llama, Mistral, Qwen, and Gemma. We evaluate them on 920 test queries: 720 domain-scenario queries, including 400 from domains held out within that component, and 200 queries from a benchmark-source held-out set. On this combined test set, probe joint accuracy ranges from 92.50\% to 94.78\% across models. On a separate deceptive evaluation with misleading surface cues, all four probes outperform the strongest lexical or sentence-embedding baseline by 19.55--23.64 percentage points in joint accuracy. However, this strong probe performance does not translate into equally accurate verbal classification: on the 920-query test set, all four models perform substantially worse under four-shot prompting than their corresponding probes. We call this gap \textbf{\textit{structured but silent}}: capability labels are more reliably accessible through linear probes than through explicit self-classification under this evaluation protocol.

Our main contributions are as follows:
\begin{itemize}
    \item We propose a three-axis, eight-class taxonomy encompassing Source, Transformation, and World Effect to formalize external capability requirements in user queries.
    \item We demonstrate that multi-dimensional tool-use capability structures are linearly decodable from hidden representations before generation, consistently across four LLM families.
    \item We expose a systematic disconnect where internal probing outperforms explicit verbalization. This structured but silent gap provides a diagnostic lens for examining discrepancies between linearly decodable capability information and explicit capability classification.
\end{itemize}

\section{Related Work}

\begin{figure*}[t]
    \centering
    \includegraphics[width=0.85\textwidth]{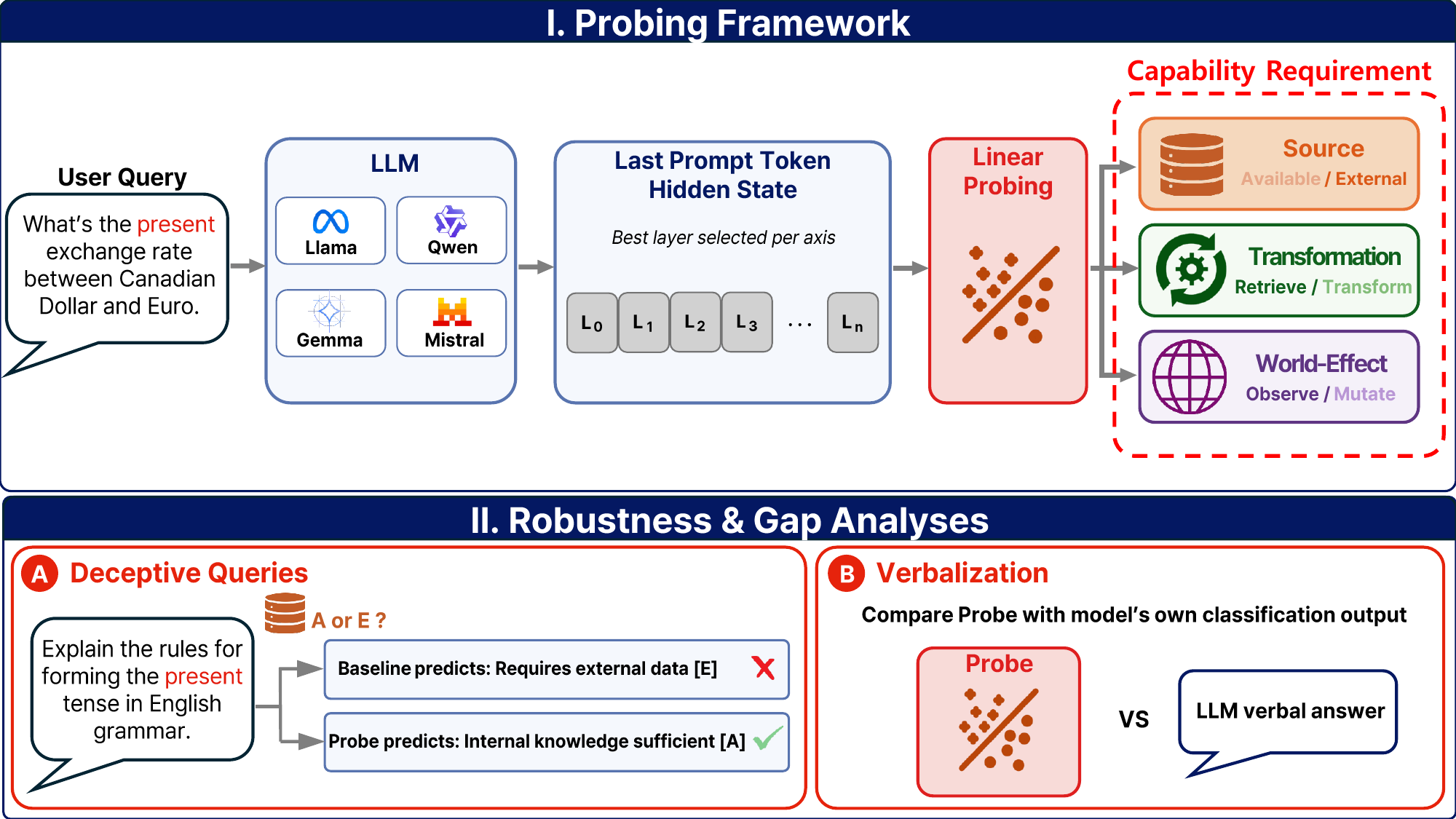}
    \caption{Overview of the TACIT probing framework. \textbf{(I)} Given a user query, we extract hidden states at the final prompt token before generation. Separate linear probes, with a layer selected on validation data for each axis, predict Source, Transformation, and World Effect, jointly defining eight capability classes. \textbf{(II)} We examine these predictions through two analyses: \textbf{(A)} robustness to misleading surface cues, comparing probes with lexical and sentence-embedding baselines on deceptive queries; and \textbf{(B)} the representation-to-verbalization gap, comparing probe predictions with the same model's explicit verbal classifications.}
    \label{fig:overview}
\end{figure*}

\subsection{Internal Representations of Language Models}
Linear probing has served as a primary diagnostic tool to dissect the informational landscape encoded within LLM hidden representations. Prior studies have successfully uncovered geometric structures corresponding to truthfulness \cite{liu2024universaltruthfulnesshyperplaneinside, bürger2024truthuniversalrobustdetection, zou2025representationengineeringtopdownapproach}, localized factual knowledge within intermediate layers \cite{gottesman2024estimatingknowledgelargelanguage}, and truth-consistent latent spaces extractable via unsupervised contrastive learning \cite{burns2024discoveringlatentknowledgelanguage}. Furthermore, recent literature demonstrates that internal states inherently reveal model errors \cite{orgad2025llmsknowshowintrinsic} and can be leveraged for real-time hallucination detection \cite{ji2024llminternalstatesreveal, su2024unsupervisedrealtimehallucinationdetection}.

Recently, this analytical paradigm has been extended to the domain of tool-augmented language models. For instance, \citet{healy2026internalrepresentationsindicatorshallucinations} show that hidden states signal hallucination risks during tool selection, while \citet{sun2026llm} demonstrate that tool necessity is linearly decodable prior to generation. Similarly, MeCo \cite{li2025adaptive} and Chain-of-Tools \cite{wu2025chainoftoolsutilizingmassiveunseen} utilize hidden representations to optimize binary tool-triggering mechanisms. While these studies primarily scrutinize binary execution choices  or downstream failure detection, our work addresses a complementary and more fundamental representational question: \textit{whether hidden states encapsulate the fine-grained capability structures required to fulfill a query}. By doing so, we shift the focus from operational tool-use decisions to the semantic and functional requirements that precede concrete tool selection.

\subsection{The Disconnect Between Knowledge and Generation}
A growing body of work demonstrates that meaningful signals internalized within hidden states do not always manifest in the model's generated outputs. \citet{azaria2023internalstatellmknows} observe that internal states accurately track the veracity of a statement even when the generated output exhibits overconfidence in a falsehood. Similarly, \citet{orgad2025llmsknowshowintrinsic} illustrate that models encode internal error signals while concurrently verbalizing incorrect answers, and \citet{gekhman2025insideouthiddenfactualknowledge} show that factual knowledge residing in intermediate hidden states frequently fails to emerge during token generation. Collectively, these findings imply that the generation process can act as a lossy channel, bottlenecking or distorting information natively accessible within the representation space.

However, existing literature on this representation-to-generation disconnect has predominantly focused on single-dimensional properties, such as factual correctness, calibration, and truthfulness. To our knowledge, no prior work has investigated whether this gap extends to multi-dimensional, structured properties defined over abstract functional taxonomies. Our work bridges this gap by evaluating external capability requirements, quantifying the precise empirical discrepancy between pre-generation probe predictions and explicit verbalized classifications across three binary capability axes and four distinct LLM families.

\section{The TACIT Framework}
We introduce \textbf{TACIT} (\textbf{T}hree-\textbf{A}xis \textbf{C}apability \textbf{I}mplicit \textbf{T}axonomy), a diagnostic framework designed to analyze whether query-side capability requirements are linearly decodable from LLM hidden states prior to text generation. As illustrated in Figure~\ref{fig:overview}, TACIT decomposes user queries along three binary capability axes, yielding eight capability classes. We evaluate pre-generation linear probes against lexical and sentence-embedding baselines and compare their predictions with four-shot verbalization.

\begin{figure*}[t]
    \centering
    \includegraphics[width=0.96\textwidth]{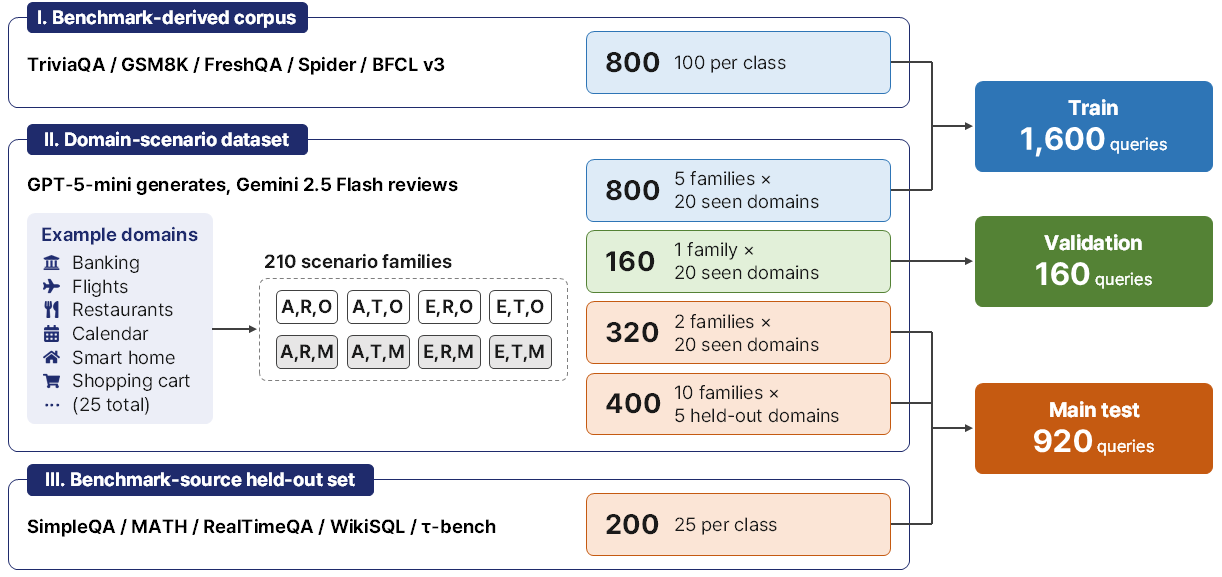}
    \caption{Overview of TACIT dataset composition and splits. Training combines 800 queries sampled from the benchmark-derived corpus with 800 domain-scenario queries. The domain-scenario dataset supplies 160 validation and 720 test queries; another 200 benchmark-source held-out queries complete the 920-query
    main test set. Each scenario family contains one query per TACIT class and remains within a single split.}
    \label{fig:Data_construct_process}
\end{figure*}

\subsection{Three-Axis Capability Taxonomy}
\label{sec:taxonomy}
To formalize the functional demands of a query, we decompose capability requirements into three binary capability axes. Our taxonomy adapts the taxonomy of perception, computation, and action proposed by \citet{wang2024toolsanywaysurveylanguage}, which categorizes tools by their explicit API actions. We project these tool-level definitions onto query-level intrinsic properties that can be inferred solely from the query text.

\paragraph{Source (S):} Inspired by the perception dimension, this axis differentiates whether the target task can be resolved natively or necessitates information retrieval from the external environment.
\begin{itemize}
    \item \textbf{Available (A):} The prerequisite information is entirely encapsulated within the query itself or resolvable via the model's parametric knowledge.
    \item \textbf{External (E):} The target objective depends on information that must be retrieved from an external source, such as real-time, search results, or dynamically changing information.
\end{itemize}

\paragraph{Transformation (T):} Adapted from the computation dimension, this axis assesses the degree of processing or reasoning required after the core data is acquired.
\begin{itemize}
    \item \textbf{Retrieve-only (R):} The query can be satisfied via direct data fetching or factual lookup without secondary operations.
    \item \textbf{Transform (T):} The query demands explicit logical or mathematical processing, such as arithmetic, multi-entity comparison, conditional branching, or structural aggregation.
\end{itemize}

\paragraph{World Effect (W):} Rooted in the action dimension, this axis determines whether the execution of the query alters the state of an external system.
\begin{itemize}
    \item \textbf{Observe (O):} The request is strictly information-seeking and leaves external environments unaltered.
    \item \textbf{Mutate (M):} The query explicitly commands a state-changing operation such as booking, database deletion, or transactional execution.
\end{itemize}

To ensure annotation rigor for World Effect, we enforce a \textit{Literal Verb Rule}: classification depends strictly on the explicit verbs within the query rather than downstream pragmatic intents. For example, `find a hotel' is annotated as \textit{Observe} even if the user may ultimately intend to book one. When conflicting verbs co-occur, the state-changing \textit{Mutate} verb takes precedence. For example, `verify availability and book' is annotated as \textit{Mutate}.

The Cartesian product of these three binary axes yields $2^3 = 8$ distinct capability classes. Table~\ref{tab:8class} provides representative examples for each class, and the full annotation definitions with boundary-case guidance are provided in Appendix~\ref{sec:taxonomy_defs}.
\begin{table*}[t]
\centering
\small
\begin{tabularx}{\textwidth}{l l l l X}
\toprule
\textbf{Class} & \textbf{S} & \textbf{T} & \textbf{W} & \textbf{Representative Example Query} \\
\midrule
(A,R,O) & A & R & O & Which chemical element has the symbol Fe? \\
(A,T,O) & A & T & O & Ruel has four books of 10 stamps and six books of 15 stamps. How many stamps does Ruel have? \\
(E,R,O) & E & R & O & What is Cristiano Ronaldo's current club? \\
(E,T,O) & E & T & O & Which department has the largest number of employees? \\
(A,R,M) & A & R & M & Can you set a new alarm for me at 17:15? \\
(A,T,M) & A & T & M & Calculate 18\% tip on my \$85 dinner bill and add the total amount to my expense report. \\
(E,R,M) & E & R & M & I need to purchase a one-way Economy class flight ticket from LA to New York for March 14th. \\
(E,T,M) & E & T & M & Look for all utility bills due this month and pay the highest-amount bill first from my primary account. \\
\bottomrule
\end{tabularx}
\caption{Taxonomy overview and representative example queries for the eight TACIT capability classes.}
\label{tab:8class}
\end{table*}

\subsection{Dataset Construction}
\label{sec:dataset_construction}
As shown in Figure~\ref{fig:Data_construct_process}, we combine benchmark-derived and domain-scenario queries for training. First, we sample 800 queries, with 100 per class, from an original benchmark-derived corpus. Its Observe queries draw on TriviaQA~\cite{joshi2017triviaqalargescaledistantly}, GSM8K~\cite{cobbe2021trainingverifierssolvemath}, FreshQA~\cite{vu2023freshllmsrefreshinglargelanguage}, and Spider~\cite{yu2019spiderlargescalehumanlabeleddataset}; its Mutate queries draw on BFCL v3~\cite{patil2025the}, with GPT-4.1-mini~\cite{openai2024gpt4technicalreport} supplementation where benchmark coverage is insufficient. After deduplication, Gemini 2.5 Flash~\cite{comanici2025gemini25pushingfrontier} reviews the capability labels and query quality, with flagged and ambiguous cases inspected before finalization.

Second, we construct a domain-scenario dataset containing 1,680 queries in 210 scenario families across 25 domains. Each family contains one query for each TACIT class within a shared domain and broad scenario, reducing associations between particular domains and capability labels. GPT-5-mini~\footnote{https://developers.openai.com/api/docs/models/gpt-5-mini} generates new queries from domain and scenario specifications rather than sampling benchmark utterances. Gemini 2.5 Flash reviews individual-query labels and quality as well as family-level coherence and class coverage; rejected queries are revised and revalidated.

The final training set combines 800 queries from each component, yielding 1,600 balanced queries with 200 per class. A separate domain-scenario validation set contains 160 queries. Twenty domains contribute training, validation, and in-domain test families, while five domains are reserved for testing; all eight queries in a scenario family remain in the same split. Source rationales, domain inventories, and review procedures
are provided in Appendix~\ref{sec:dataset_details}. Generation prompts are provided in Appendices~\ref{sec:synthesis_prompts} and~\ref{sec:domain_scenario_prompts}. Evaluation sets are described in Section~\ref{sec:evaluation_datasets}.

\subsection{Pre-generation Representation Extraction}
\label{sec:representation_extraction}
For each query $q$ in our benchmark, we extract its latent representation via a single forward pass without generating output tokens. Let $P = [t_1, t_2, \dots, t_T]$ denote the sequence of prompt tokens formatted via the model's canonical chat template. We isolate the hidden state vector at the final prompt token $t_T$, denoted as $h_T \in \mathbb{R}^d$, where $d$ represents the model's hidden dimension. Under causal self-attention, the hidden state at this terminal prompt position can attend to the entire preceding query before the first response token is generated.

For a given language model with $L$ transformer layers, this procedure yields a multi-layer representation tensor $H \in \mathbb{R}^{N \times L \times d}$, where $N$ signifies the total number of queries. For Qwen 3, we disable its optional thinking mode and use the standard chat template to maintain a consistent representation extraction procedure across models.

\subsection{Axis-wise Linear Probing}
\label{sec:probe_training}
To evaluate the linearly decodable structure of each functional property, we train independent $\ell_2$-regularized logistic regression probes for each capability axis $a \in \{S, T, W\}$ across every individual transformer layer $l \in \{1, \dots, L\}$. 

Let $x^{(l)} \in \mathbb{R}^d$ be the activation vector at layer $l$ for a given instance. The probe minimizes the standard cross-entropy loss with an $\ell_2$ regularization penalty:
\begin{equation}
\scalebox{0.75}{%
$\displaystyle
    \min_{w, b} \sum_{i=1}^{N_{\text{train}}} \log \left( 1 + \exp\left( -y_i \left( w^{\top} x_i^{(l)} + b \right) \right) \right) + \lambda \|w\|_2^2
$%
}
\end{equation}

where $y_i \in \{-1,1\}$ is the binary label for axis $a$, and $\lambda$ controls the regularization penalty. We use logistic regression with a fixed inverse regularization strength of $C=1.0$, the L-BFGS solver, and a maximum of 1,000 iterations. Activation features are standardized to zero mean and unit variance using training-split statistics only.

\section{Experimental Setup}

\subsection{Evaluation Datasets}
\label{sec:evaluation_datasets}
We evaluate generalization using a main test set of 920 queries and robustness to misleading surface cues using a separate deceptive set.

\paragraph{Main Test Set}
The main test set comprises three complementary subsets. The domain-scenario dataset contributes 320 queries from new scenario families within the 20 training domains and 400 queries from five additional domains reserved for testing in this component. Scenario families do not overlap across training, validation, and test splits.

The remaining 200 queries assess generalization across benchmark sources while preserving the same capability classes. We construct this subset from benchmarks different from those used for the original training component: SimpleQA \cite{wei2024measuringshortformfactualitylarge} for (A,R,O), MATH \cite{hendrycks2021measuringmathematicalproblemsolving} for (A,T,O), RealTimeQA \cite{kasai2024realtimeqawhatsanswer} for (E,R,O), and WikiSQL \cite{zhong2017seq2sqlgeneratingstructuredqueries} for (E,T,O). For the four \textit{Mutate} classes, we curate queries from the airline and retail environments of $\tau$-bench \cite{yao2024taubenchbenchmarktoolagentuserinteraction}, supplementing gaps in class coverage with GPT-4.1-mini-generated examples. This subset contains 25 queries per class. All three test subsets are balanced across the eight TACIT classes. Detailed source mappings and construction procedures appear in Appendix~\ref{sec:dataset_details}.

\paragraph{Deceptive Evaluation}
To examine sensitivity to misleading surface cues, we construct a separate set of 220 deceptive queries. Each query targets one capability axis and contains wording that suggests an incorrect label on that axis while retaining a determinate gold label under the TACIT definitions. The set contains 77 queries targeting Source, 70 targeting Transformation, and 73 targeting World Effect.

For example, one query asks: ``The report titled `Current Stock' lists 44 USB-C cables. How many cables does it list?" Although `Current Stock' may suggest a need to access live inventory, the requested value is explicitly supplied in the query. Its gold labels are therefore Source=A, Transformation=R, and World Effect=O.

We generate candidate queries using GPT-5~\footnote{https://openai.com/ko-KR/index/introducing-gpt-5/}. Claude Sonnet 5~\footnote{https://www.anthropic.com/news/claude-sonnet-5} independently predicts the three capability labels without access to the proposed labels and checks naturalness, ambiguity, grounding, and the presence of an identifiable misleading cue. We retain 220 queries that pass these checks, freeze the set before evaluation, and use it exclusively for robustness analysis. Construction details and prompts are provided in Appendices~\ref{sec:deceptive_details} and~\ref{sec:deceptive_prompts}.

\subsection{Models}
\label{sec:models}
We probe four instruction-tuned open-weight LLM families spanning diverse architectures, layer depths, and pretraining distributions: Llama 3.1 8B Instruct \cite{grattafiori2024llama3herdmodels}, Qwen 3 8B \cite{yang2025qwen3technicalreport}, Gemma 2 9B IT \cite{gemmateam2024gemma2improvingopen}, and Mistral 7B Instruct v0.3 \cite{jiang2023mistral7b}. We designate Llama 3.1 8B Instruct as our primary subject for in-depth architectural analysis, utilizing the remaining three models to establish cross-model generalization in Section~\ref{sec:crossmodel}.

\begin{table}[t]
\centering
\small
\scalebox{0.83}{%
\begin{tabular}{lccc}
\toprule
\textbf{Probe Axis} & \textbf{Best Layer} & \textbf{Relative Depth} & \textbf{Accuracy (\%)} \\
\midrule
Source         & L14 & 43.8\% & 97.28 \\
Transformation & L7 & 21.9\% & 97.28 \\
World Effect   & L16 & 50.0\% & 99.24 \\
\midrule
\textbf{Joint (3-axis)} & \multicolumn{2}{c}{\textit{per-axis best layers}} & \textbf{94.02} \\
\bottomrule
\end{tabular}%
}
\caption{Best-layer linear probing accuracy on Llama 3.1 8B Instruct (main test set, $N = 920$). Relative depth is $l/32$. Joint accuracy requires all three axis predictions to be correct.}
\label{tab:probe_accuracy}
\end{table}

\begin{table}[t]
\centering
\small
\begin{tabular}{l ccc c}
\toprule
\textbf{Capability Class} & \textbf{S} & \textbf{T} & \textbf{W} & \textbf{Joint Accuracy (\%)} \\
\midrule
(A,R,O) & A & R & O & 91.30 \\
(A,T,O) & A & T & O & 98.26 \\
(E,R,O) & E & R & O & 96.52 \\
(E,T,O) & E & T & O & 90.43 \\
(A,R,M) & A & R & M & 93.91 \\
(A,T,M) & A & T & M & 94.78 \\
(E,R,M) & E & R & M & 99.13 \\
(E,T,M) & E & T & M & 87.83 \\
\bottomrule
\end{tabular}
\caption{Per-class joint accuracy on Llama 3.1 8B Instruct (main test set, $N = 115$ per class).}
\label{tab:per_class_acc}
\end{table}

\subsection{Baselines}
\label{sec:baselines}
We compare hidden-state probes with lexical and sentence-embedding baselines to assess how effectively different query representations support capability classification. We use a TF-IDF Bag-of-Words (BoW) classifier over word unigrams and bigrams with $\ell_2$-regularized logistic regression. A word and character-level N-gram classifier extends these features with character 3--5-gram TF-IDF features and selects the regularization strength on the validation set. We also evaluate an E5 sentence-embedding baseline \cite{wang2022text}, using a frozen E5-large-v2 encoder followed by logistic regression.

Each baseline trains a separate classifier for each capability axis using the same 1,600 training queries as the hidden-state probes. All baselines are evaluated on the same main and deceptive test sets as the probes.

\subsection{Evaluation Metrics}
\label{sec:metrics}
We evaluate performance on the 920-query main test set described in Section~\ref{sec:evaluation_datasets}. We report both \textit{per-axis accuracy}, which measures classification accuracy for each capability axis, and \textit{joint three-axis accuracy}, which requires all three predicted labels to match the ground truth. For each model, we select the \textit{best layer} for each axis by maximizing validation accuracy. When multiple layers achieve the same maximum, we select the combination with the highest joint validation accuracy. These selected layers are used for all test evaluations.

For deceptive evaluation, we report \textit{target-axis accuracy} on each query's designated axis, aggregated across 220 queries by weighting each axis's accuracy by its query count (77 Source, 70 Transformation, and 73 World Effect).

For main-test accuracy estimates and paired comparisons between methods, we compute 95\% percentile bootstrap confidence intervals using 10,000 resamples. Resampling is stratified by test subset, with scenario families as the sampling unit for domain-scenario queries and individual queries for the 200-query benchmark-source subset. For method comparisons, we resample the per-query correctness differences using the same sampling units.

\subsection{Verbalization Setup}
\label{sec:verbal_setup}
We compare linear decodability from hidden states with four-shot verbalization for all four models on the same 920 test queries. The prompts specify the TACIT taxonomy, including the Literal Verb Rule, and request labels for all three capability axes. Four fixed labeled examples are shared across models. The full prompt is provided in Appendix~\ref{sec:verbal_prompt}.

\section{Results}

\subsection{RQ1: Are External Capability Requirements Linearly Encoded Before Generation?}
\label{sec:decodability}
We first investigate whether the three core capability axes defined by TACIT are linearly decodable from the model's latent representations before any response token is produced. To answer this, we evaluate the best-layer probing accuracy on the Llama 3.1 8B Instruct model across the main test set ($N=920$). Full per-layer validation accuracies are reported in Appendix Table~\ref{tab:layer_sweep_all}.

As reported in Table~\ref{tab:probe_accuracy}, the selected probing layers span early and intermediate depths of the network. Specifically, the best layers are Layer 14 (43.8\% relative depth) for Source, Layer 7 (21.9\%) for Transformation, and Layer 16 (50.0\%) for World Effect.

The axis-wise linear probes achieve a joint three-axis accuracy of 94.02\% (95\% CI: 92.50--95.43\%). This indicates that the three-axis capability profile of a user query can be reliably decoded before token generation begins.

To more closely examine class-level performance, Table~\ref{tab:per_class_acc} reports joint accuracy across the eight TACIT classes. Seven of the eight classes achieve at least 90\% accuracy, with \textit{(E,R,M)} reaching 99.13\%. Performance is lower for the \textit{(E,T,M)} class at 87.83\%, indicating that this class remains comparatively more challenging for the probes, although further analysis is needed to determine the source of this difficulty.

\begin{table}[t]
\centering
\small
\begin{tabular}{lc}
\toprule
\textbf{Axis Pair} & $\mathbf{\cos(w_i,\, w_j)}$ \\
\midrule
Source $\times$ Transformation       & $\phantom{-}0.016$ \\
Source $\times$ World Effect         & $-0.009$ \\
Transformation $\times$ World Effect & $-0.030$ \\
\bottomrule
\end{tabular}
\caption{Cosine similarity between best-layer probe weight vectors on Llama 3.1 8B Instruct.}
\label{tab:cosine_sim}
\end{table}

\begin{table}[t]
\centering
\small
\resizebox{\columnwidth}{!}{%
\begin{tabular}{lcccc}
\toprule
 & \multicolumn{3}{c}{\textbf{Target-axis Accuracy}} & \\
\cmidrule(lr){2-4}
\textbf{Method} & Source & Trans. & W-Eff. & \textbf{Joint} \\
\midrule
BoW                  & 61.04 & 44.29 & 79.45 & 38.64 \\
N-gram               & 64.94 & 44.29 & 76.71 & 35.91 \\
E5                   & 55.84 & 55.71 & 75.34 & 37.27 \\
\midrule
Llama 3.1 8B probe   & \textbf{72.73} & \textbf{67.14} & \textbf{97.26} & \textbf{60.45} \\
\bottomrule
\end{tabular}%
}
\caption{Accuracy on the deceptive evaluation set (\%). Source, Transformation, and World Effect columns evaluate queries targeting the respective axis ($N=77$, $70$, and $73$). Joint accuracy is computed over all 220 queries and requires all three axis predictions to be correct.}
\label{tab:deceptive}
\end{table}

\begin{table}[t]
\centering
\small
\begin{tabular}{lccc}
\toprule
\textbf{Accuracy} & \textbf{Probe} & \textbf{four-shot} & \textbf{Gap (pp)} \\
\midrule
Source         & 97.28 & 78.80 & +18.48 \\
Transformation & 97.28 & 88.80 & +8.48  \\
World Effect   & 99.24 & 98.04 & +1.20  \\
\midrule
\textbf{Joint (3-axis)} & \textbf{94.02} & \textbf{68.04} & \textbf{+25.98} \\
\bottomrule
\end{tabular}
\caption{Probe and four-shot verbalization accuracy on Llama 3.1 8B Instruct (main test set, $N=920$). The gap is probe accuracy minus verbalization accuracy in percentage points.}
\label{tab:verbalization}
\end{table}

\begin{table*}[t]
\centering
\small
\setlength{\tabcolsep}{4pt}
\begin{tabular}{l ccccc cccc}
\toprule
 & \multicolumn{5}{c}{\textbf{Main Test Joint Accuracy (\%)}} &
   \multicolumn{4}{c}{\textbf{Deceptive-Set Joint Accuracy (\%)}} \\
\cmidrule(lr){2-6}\cmidrule(lr){7-10}
\textbf{Model} & Probe & BoW & N-gram & E5 & four-shot
               & Probe & BoW & N-gram & E5 \\
\midrule
Llama 3.1 8B    & \textbf{94.02} & 83.26 & 85.98 & 84.13 & 68.04
                & \textbf{60.45} & 38.64 & 35.91 & 37.27 \\
Qwen 3 8B       & \textbf{92.50} & 83.26 & 85.98 & 84.13 & 59.67
                & \textbf{62.27} & 38.64 & 35.91 & 37.27 \\
Gemma 2 9B      & \textbf{94.78} & 83.26 & 85.98 & 84.13 & 69.46
                & \textbf{60.91} & 38.64 & 35.91 & 37.27 \\
Mistral 7B v0.3 & \textbf{92.50} & 83.26 & 85.98 & 84.13 & 54.35
                & \textbf{58.18} & 38.64 & 35.91 & 37.27 \\
\bottomrule
\end{tabular}
\caption{Cross-model joint three-axis accuracy on the main test set ($N=920$) and the deceptive evaluation set ($N=220$). The text-only baselines have the same scores for each model because they use the same query data independently of the probed LLM. The four-shot column reports each model's verbal classification on the main test set.}
\label{tab:crossmodel}
\end{table*}

To examine the relationship between the learned probe directions, we compute the cosine similarity between the weight vectors ($w_i, w_j$) of the respective best-layer probes. As summarized in Table~\ref{tab:cosine_sim}, all pairwise cosine similarities are close to zero. These near-zero similarities do not by themselves establish
semantic independence among the three axes.

\noindent\textbf{Takeaway:} Query-side capability requirements are reliably linearly decodable from pre-generation representations at early and intermediate transformer layers, with near-zero pairwise cosine similarities between the selected probe weight vectors.

\subsection{RQ2: Do Probes Capture Abstract Taxonomy or Surface Lexical Shortcuts?}
\label{sec:deceptive}
To test whether probe performance reflects superficial lexical correlations, we evaluate robustness to misleading surface cues on the deceptive set described in Section~\ref{sec:evaluation_datasets}. We report results on all 220 deceptive queries.

As shown in Table~\ref{tab:deceptive}, the Llama 3.1 8B hidden-state probes outperform the BoW baseline across all three targeted axes. The largest difference is observed for Transformation ($\Delta = +22.86$ percentage points), followed by World Effect ($+17.81$) and Source ($+11.69$). We further compare against stronger text-based baselines, including an n-gram classifier and E5 sentence embeddings. The probes achieve 79.09\% target-axis accuracy and 60.45\% joint accuracy, compared with 61.82--62.27\% and 35.91--38.64\%, respectively, for the three baselines. These results provide evidence that the probes capture capability information beyond the surface cues exploited by the tested baselines.

\noindent\textbf{Takeaway:} Hidden-state probes outperform lexical and sentence-embedding baselines under misleading surface cues, supporting capability decoding beyond the patterns captured by these baselines.

\subsection{RQ3: Does Linearly Decodable Capability Information Surface in Explicit Generation?}
\label{sec:verbalization}
Having shown that capability requirements are linearly decodable from hidden states, we next examine the representation-to-verbalization gap: whether models can explicitly classify the same capability requirements under direct prompting.

As shown in Table~\ref{tab:verbalization}, the Llama 3.1 8B hidden-state probes achieve higher accuracy than four-shot verbal classification across all three axes. The difference is particularly large in joint three-axis accuracy, with the probes achieving 94.02\% compared with 68.04\% for verbalized classification ($\Delta = 25.98$ percentage points; 95\% CI: 23.37--28.59). This comparison indicates a substantial gap between information that is linearly accessible from hidden states and what the model explicitly expresses under the tested prompting conditions.

\begin{figure}[t]
  \centering
  \includegraphics[width=\columnwidth]{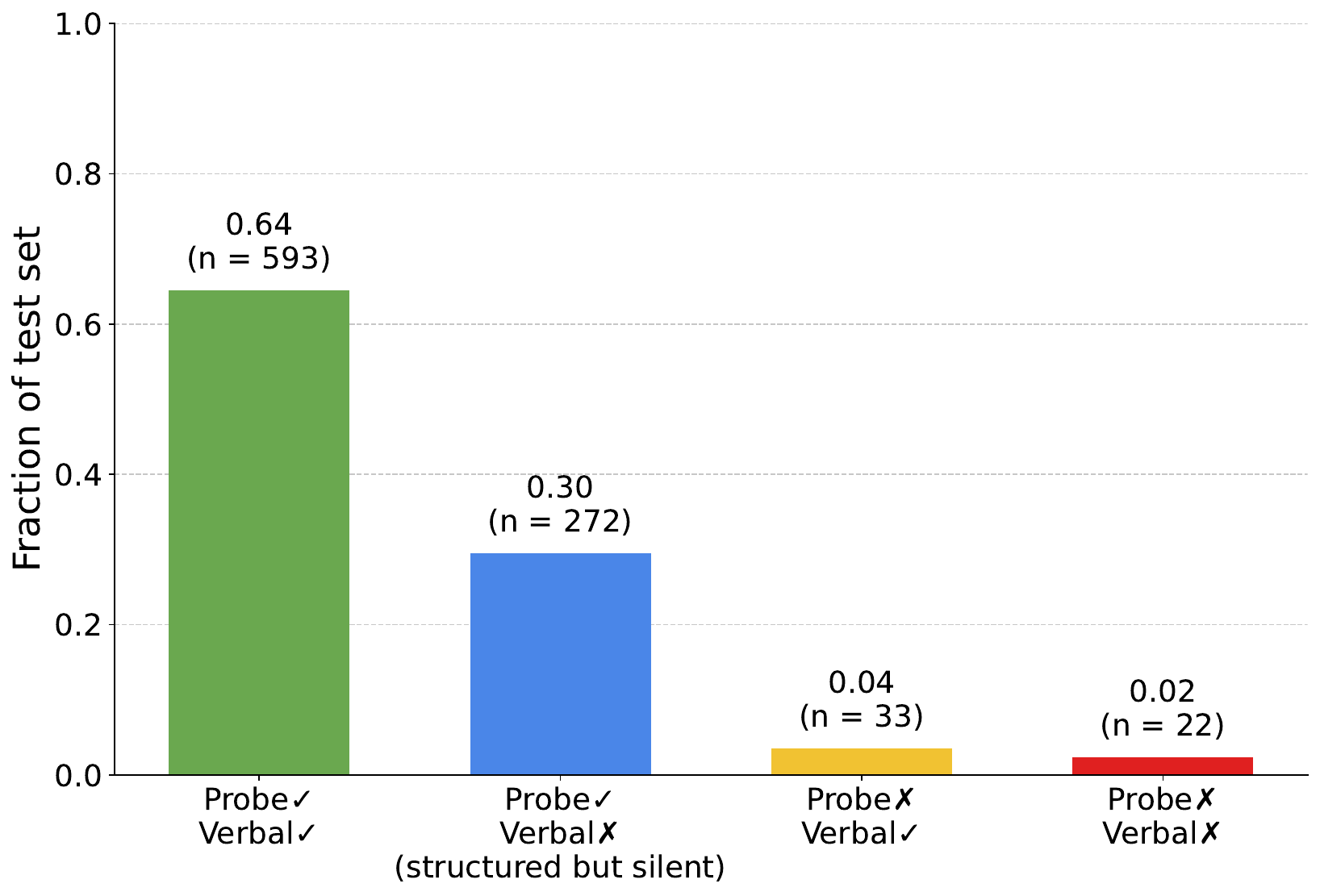}
  \caption{Four-way breakdown of joint three-axis correctness for the probe and four-shot verbalization on Llama 3.1 8B Instruct ($N=920$). Bars show the fraction of test queries in each outcome, with counts shown above them.}
  \label{fig:verbal}
\end{figure}

To further examine the representation-to-verbalization gap, Figure~\ref{fig:verbal} reports a four-way breakdown of probe and verbalized classification outcomes. In 29.6\% of queries, the probe predicts all three labels correctly while verbalized classification does not. The reverse pattern, where verbalized classification is jointly correct but the probe is not, occurs in only 3.6\% of queries. This asymmetry shows that disagreements between the two methods are predominantly cases in which the probe succeeds while explicit classification fails under the tested prompting conditions.

\noindent\textbf{Takeaway:} We observe a clear and asymmetric representation-to-verbalization gap. Capability requirements can be linearly decoded from hidden states even when the model fails to classify them correctly through four-shot verbalization.

\subsection{RQ4: Does the "Structured but Silent" Phenomenon Generalize Across Models?}
\label{sec:crossmodel}
To examine whether our findings are specific to Llama or generalize across model families, we extend the same experimental setup to three additional models: Qwen 3 8B, Gemma 2 9B IT, and Mistral 7B Instruct v0.3. Table~\ref{tab:crossmodel} summarizes the resulting cross-model comparison.

High linear decodability of capability requirements is observed across all four model families. Joint three-axis probe accuracy on the main test set ranges from 92.50\% to 94.78\%, compared with 83.26--85.98\% for the BoW, n-gram, and E5 baselines. The selected best layers differ across models and are reported alongside the full layer-wise validation results in Appendix~\ref{sec:appendix_layer_sweep}.

On the deceptive evaluation set, the probes achieve 58.18--62.27\% joint accuracy, compared with 38.64\% for BoW, 35.91\% for n-gram, and 37.27\% for E5. The \textbf{\textit{structured but silent}} pattern is likewise observed across all four models: four-shot verbalization achieves 54.35--69.46\% joint accuracy on the main test set, leaving a representation-to-verbalization gap of 25.33--38.15 percentage points.

The corresponding 95\% confidence intervals for the probe--verbalization gaps are 23.37--28.59, 29.78--35.76, 22.39--28.26, and 35.00--41.30 percentage points for Llama, Qwen, Gemma, and Mistral, respectively.

\noindent\textbf{Takeaway:} Across four model families, capability requirements remain linearly decodable and comparatively robust to deceptive wording, while four-shot verbalization is consistently less accurate than the corresponding probe.

\section{Conclusion}
We introduced TACIT, a three-axis framework for analyzing query-side capability requirements relevant to tool use. Across four open-weight LLM families, these requirements are linearly decodable from pre-generation hidden states, with 92.50--94.78\% joint accuracy on the 920-query main test set. The probes also outperform lexical and sentence-embedding baselines on deceptive queries, although their accuracy decreases under misleading wording.

We observe a consistent gap between probe accuracy and explicit four-shot classification. Across the four models, probe joint accuracy exceeds verbalization by 25.33--38.15 percentage points. We call this pattern \textit{structured but silent}: capability labels are more reliably accessible through linear probes than expressed through generation under the evaluated prompting protocol.

Future work should test whether these decodable representations causally influence tool selection and whether they can improve behavior in agents that actually execute tools.

\section*{Limitations}
TACIT addresses capability requirements implied by a user query, not the full process of tool use. Our experiments do not test whether models use the probed representations when selecting or executing tools. Tool availability, API schemas, permissions, multi-step planning, and execution feedback are outside the present evaluation. Establishing causal use would require interventions on the representations and tests of downstream tool behavior.

Dataset construction remains a potential source of artifacts. In the original corpus, some capability classes are associated with particular benchmarks and query styles, and underrepresented Mutate classes require synthetic supplementation. The domain-scenario dataset places all eight classes within each scenario and includes unseen-domain tests, but this design cannot rule out every stylistic shortcut. Because all eight classes appear in training, the results also do not establish generalization to an entirely unseen capability class.

The deceptive evaluation exposes remaining sensitivity to misleading cues: probe joint accuracy falls to 58.18--62.27\% on this set, with errors particularly on queries targeting Source and Transformation. These queries were generated and screened with language models. The deceptive results should therefore be interpreted as a targeted stress test rather than an exhaustive measure of robustness.

Our probe--verbalization comparison also uses different forms of access to the model. The probes are supervised classifiers trained on hidden states, whereas verbalization is measured through a fixed four-shot prompt and output parser. The observed gap describes these evaluation conditions; it does not show that the models could never express the same labels under other prompts or training procedures.

Finally, our empirical scope is limited to four open-weight models in the 7--9B parameter range. Whether the observed patterns generalize to substantially larger models, multilingual settings, or models with explicit reasoning modes remains an open empirical question.

\section*{Ethical Considerations}
This study complies with standard ethical guidelines in natural language processing. Our work does not involve human participants, and no personally identifiable information (PII) was collected, processed, or utilized. All datasets leveraged in our benchmarks—including TriviaQA, GSM8K, FreshQA, Spider, SimpleQA, MATH, RealTimeQA, WikiSQL, BFCL v3, and Tau-bench—are publicly accessible and utilized in strict adherence to their respective licensing agreements.

Proprietary models were used for data generation, provisional labeling, and automated review. Model-generated data may introduce distributional or stylistic biases. We address these concerns through explicit taxonomy definitions, query- and family-level review, and targeted revision. These procedures provide automated quality checks.

As a diagnostic evaluation requiring direct access to intermediate activations, our methodology is inherently tailored to open-weight models and cannot be deployed in black-box configurations. The primary intent of this research is to advance the transparency and reliability of autonomous agent systems by diagnosing routing failures. While probing techniques could theoretically be adapted to extract unintended or sensitive latent attributes, the linear probes optimized in this study are strictly bounded to abstract, task-level capability labels and possess no utility beyond functional diagnostic applications.

\section*{Acknowledgements}
This work was supported by the Korea Internet \& Security Agency (KISA) grant funded by the Korea government (PIPC) (No.~RS-2026-25526342, Development of Technologies for Preventing Sensitive Information Inference and Risk Assessment in Foundation Model Operations). This research was also supported by the Culture, Sports and Tourism R\&D Program through the Korea Creative Content Agency grant funded by the Ministry of Culture, Sports and Tourism in 2026 (Project Name: Develop AI agent technology to connect knowledge through public cultural facility-based discussion and communication, Project Number: RS-2026-25520645). Further support was provided by the National Research Foundation of Korea (NRF) grant funded by the Korea government (MSIT) (RS-2026-25483747). This work was additionally supported by the Institute of Information \& communications Technology Planning \& Evaluation (IITP, AI Computing Support Project for R\&D) grant funded by the Korea government (MSIT) (High-Performance Research AI Computing Infrastructure Support at the 2~PFLOPS Scale, RS-2026-25505492).

\bibliography{custom}

@inproceedings{qin2024toolllm,
  title={Toolllm: Facilitating large language models to master 16000+ real-world apis},
  author={Qin, Yujia and Liang, Shihao and Ye, Yining and Zhu, Kunlun and Yan, Lan and Lu, Yaxi and Lin, Yankai and Cong, Xin and Tang, Xiangru and Qian, Bill and others},
  booktitle={International Conference on Learning Representations},
  volume={2024},
  pages={9695--9717},
  year={2024}
}

@inproceedings{huang2024metatool,
  title={Metatool benchmark for large language models: Deciding whether to use tools and which to use},
  author={Huang, Yue and Shi, Jiawen and Li, Yuan and Fan, Chenrui and Wu, Siyuan and Zhang, Qihui and Liu, Yixin and Zhou, Pan and Wan, Yao and Gong, Neil and others},
  booktitle={International Conference on Learning Representations},
  volume={2024},
  pages={42978--43007},
  year={2024}
}

@inproceedings{li2025adaptive,
  title={Adaptive tool use in large language models with meta-cognition trigger},
  author={Li, Wenjun and Li, Dexun and Dong, Kuicai and Zhang, Cong and Zhang, Hao and Liu, Weiwen and Wang, Yasheng and Tang, Ruiming and Liu, Yong},
  booktitle={Proceedings of the 63rd Annual Meeting of the Association for Computational Linguistics (Volume 1: Long Papers)},
  pages={13346--13370},
  year={2025}
}

@misc{wang2024toolsanywaysurveylanguage,
      title={What Are Tools Anyway? A Survey from the Language Model Perspective}, 
      author={Zhiruo Wang and Zhoujun Cheng and Hao Zhu and Daniel Fried and Graham Neubig},
      year={2024},
      eprint={2403.15452},
      archivePrefix={arXiv},
      primaryClass={cs.CL},
      url={https://arxiv.org/abs/2403.15452}, 
}

@misc{marks2024geometrytruthemergentlinear,
      title={The Geometry of Truth: Emergent Linear Structure in Large Language Model Representations of True/False Datasets}, 
      author={Samuel Marks and Max Tegmark},
      year={2024},
      eprint={2310.06824},
      archivePrefix={arXiv},
      primaryClass={cs.AI},
      url={https://arxiv.org/abs/2310.06824}, 
}

@misc{orgad2025llmsknowshowintrinsic,
      title={LLMs Know More Than They Show: On the Intrinsic Representation of LLM Hallucinations}, 
      author={Hadas Orgad and Michael Toker and Zorik Gekhman and Roi Reichart and Idan Szpektor and Hadas Kotek and Yonatan Belinkov},
      year={2025},
      eprint={2410.02707},
      archivePrefix={arXiv},
      primaryClass={cs.CL},
      url={https://arxiv.org/abs/2410.02707}, 
}

@misc{healy2026internalrepresentationsindicatorshallucinations,
      title={Internal Representations as Indicators of Hallucinations in Agent Tool Selection}, 
      author={Kait Healy and Bharathi Srinivasan and Visakh Madathil and Jing Wu},
      year={2026},
      eprint={2601.05214},
      archivePrefix={arXiv},
      primaryClass={cs.AI},
      url={https://arxiv.org/abs/2601.05214}, 
}

@misc{yakovlev2024toolkenimprovingllmtool,
      title={Toolken+: Improving LLM Tool Usage with Reranking and a Reject Option}, 
      author={Konstantin Yakovlev and Sergey Nikolenko and Andrey Bout},
      year={2024},
      eprint={2410.12004},
      archivePrefix={arXiv},
      primaryClass={cs.CL},
      url={https://arxiv.org/abs/2410.12004}, 
}

@inproceedings{Alazraki_2025,
   title={Meta-Reasoning Improves Tool Use in Large Language Models},
   url={http://dx.doi.org/10.18653/v1/2025.findings-naacl.440},
   DOI={10.18653/v1/2025.findings-naacl.440},
   booktitle={Findings of the Association for Computational Linguistics: NAACL 2025},
   publisher={Association for Computational Linguistics},
   author={Alazraki, Lisa and Rei, Marek},
   year={2025},
   pages={7885–7897} }

@misc{hsieh2023tooldocumentationenableszeroshot,
      title={Tool Documentation Enables Zero-Shot Tool-Usage with Large Language Models}, 
      author={Cheng-Yu Hsieh and Si-An Chen and Chun-Liang Li and Yasuhisa Fujii and Alexander Ratner and Chen-Yu Lee and Ranjay Krishna and Tomas Pfister},
      year={2023},
      eprint={2308.00675},
      archivePrefix={arXiv},
      primaryClass={cs.CL},
      url={https://arxiv.org/abs/2308.00675}, 
}

@misc{burns2024discoveringlatentknowledgelanguage,
      title={Discovering Latent Knowledge in Language Models Without Supervision}, 
      author={Collin Burns and Haotian Ye and Dan Klein and Jacob Steinhardt},
      year={2024},
      eprint={2212.03827},
      archivePrefix={arXiv},
      primaryClass={cs.CL},
      url={https://arxiv.org/abs/2212.03827}, 
}

@misc{liu2024universaltruthfulnesshyperplaneinside,
      title={On the Universal Truthfulness Hyperplane Inside LLMs}, 
      author={Junteng Liu and Shiqi Chen and Yu Cheng and Junxian He},
      year={2024},
      eprint={2407.08582},
      archivePrefix={arXiv},
      primaryClass={cs.CL},
      url={https://arxiv.org/abs/2407.08582}, 
}

@misc{bürger2024truthuniversalrobustdetection,
      title={Truth is Universal: Robust Detection of Lies in LLMs}, 
      author={Lennart Bürger and Fred A. Hamprecht and Boaz Nadler},
      year={2024},
      eprint={2407.12831},
      archivePrefix={arXiv},
      primaryClass={cs.CL},
      url={https://arxiv.org/abs/2407.12831}, 
}

@misc{zou2025representationengineeringtopdownapproach,
      title={Representation Engineering: A Top-Down Approach to AI Transparency}, 
      author={Andy Zou and Long Phan and Sarah Chen and James Campbell and Phillip Guo and Richard Ren and Alexander Pan and Xuwang Yin and Mantas Mazeika and Ann-Kathrin Dombrowski and Shashwat Goel and Nathaniel Li and Michael J. Byun and Zifan Wang and Alex Mallen and Steven Basart and Sanmi Koyejo and Dawn Song and Matt Fredrikson and J. Zico Kolter and Dan Hendrycks},
      year={2025},
      eprint={2310.01405},
      archivePrefix={arXiv},
      primaryClass={cs.LG},
      url={https://arxiv.org/abs/2310.01405}, 
}

@misc{gottesman2024estimatingknowledgelargelanguage,
      title={Estimating Knowledge in Large Language Models Without Generating a Single Token}, 
      author={Daniela Gottesman and Mor Geva},
      year={2024},
      eprint={2406.12673},
      archivePrefix={arXiv},
      primaryClass={cs.CL},
      url={https://arxiv.org/abs/2406.12673}, 
}

@misc{ji2024llminternalstatesreveal,
      title={LLM Internal States Reveal Hallucination Risk Faced With a Query}, 
      author={Ziwei Ji and Delong Chen and Etsuko Ishii and Samuel Cahyawijaya and Yejin Bang and Bryan Wilie and Pascale Fung},
      year={2024},
      eprint={2407.03282},
      archivePrefix={arXiv},
      primaryClass={cs.CL},
      url={https://arxiv.org/abs/2407.03282}, 
}

@misc{su2024unsupervisedrealtimehallucinationdetection,
      title={Unsupervised Real-Time Hallucination Detection based on the Internal States of Large Language Models}, 
      author={Weihang Su and Changyue Wang and Qingyao Ai and Yiran HU and Zhijing Wu and Yujia Zhou and Yiqun Liu},
      year={2024},
      eprint={2403.06448},
      archivePrefix={arXiv},
      primaryClass={cs.CL},
      url={https://arxiv.org/abs/2403.06448}, 
}

@article{sun2026llm,
  title={LLM Agents Already Know When to Call Tools--Even Without Reasoning},
  author={Sun, Chung-En and Liu, Linbo and Yan, Ge and Wang, Zimo and Weng, Tsui-Wei},
  journal={arXiv preprint arXiv:2605.09252},
  year={2026}
}

@misc{wu2025chainoftoolsutilizingmassiveunseen,
      title={Chain-of-Tools: Utilizing Massive Unseen Tools in the CoT Reasoning of Frozen Language Models}, 
      author={Mengsong Wu and Tong Zhu and Han Han and Xiang Zhang and Wenbiao Shao and Wenliang Chen},
      year={2025},
      eprint={2503.16779},
      archivePrefix={arXiv},
      primaryClass={cs.CL},
      url={https://arxiv.org/abs/2503.16779}, 
}

@misc{azaria2023internalstatellmknows,
      title={The Internal State of an LLM Knows When It's Lying}, 
      author={Amos Azaria and Tom Mitchell},
      year={2023},
      eprint={2304.13734},
      archivePrefix={arXiv},
      primaryClass={cs.CL},
      url={https://arxiv.org/abs/2304.13734}, 
}

@misc{gekhman2025insideouthiddenfactualknowledge,
      title={Inside-Out: Hidden Factual Knowledge in LLMs}, 
      author={Zorik Gekhman and Eyal Ben David and Hadas Orgad and Eran Ofek and Yonatan Belinkov and Idan Szpektor and Jonathan Herzig and Roi Reichart},
      year={2025},
      eprint={2503.15299},
      archivePrefix={arXiv},
      primaryClass={cs.CL},
      url={https://arxiv.org/abs/2503.15299}, 
}

@misc{joshi2017triviaqalargescaledistantly,
      title={TriviaQA: A Large Scale Distantly Supervised Challenge Dataset for Reading Comprehension}, 
      author={Mandar Joshi and Eunsol Choi and Daniel S. Weld and Luke Zettlemoyer},
      year={2017},
      eprint={1705.03551},
      archivePrefix={arXiv},
      primaryClass={cs.CL},
      url={https://arxiv.org/abs/1705.03551}, 
}

@misc{cobbe2021trainingverifierssolvemath,
      title={Training Verifiers to Solve Math Word Problems}, 
      author={Karl Cobbe and Vineet Kosaraju and Mohammad Bavarian and Mark Chen and Heewoo Jun and Lukasz Kaiser and Matthias Plappert and Jerry Tworek and Jacob Hilton and Reiichiro Nakano and Christopher Hesse and John Schulman},
      year={2021},
      eprint={2110.14168},
      archivePrefix={arXiv},
      primaryClass={cs.LG},
      url={https://arxiv.org/abs/2110.14168}, 
}

@misc{vu2023freshllmsrefreshinglargelanguage,
      title={FreshLLMs: Refreshing Large Language Models with Search Engine Augmentation}, 
      author={Tu Vu and Mohit Iyyer and Xuezhi Wang and Noah Constant and Jerry Wei and Jason Wei and Chris Tar and Yun-Hsuan Sung and Denny Zhou and Quoc Le and Thang Luong},
      year={2023},
      eprint={2310.03214},
      archivePrefix={arXiv},
      primaryClass={cs.CL},
      url={https://arxiv.org/abs/2310.03214}, 
}

@misc{yu2019spiderlargescalehumanlabeleddataset,
      title={Spider: A Large-Scale Human-Labeled Dataset for Complex and Cross-Domain Semantic Parsing and Text-to-SQL Task}, 
      author={Tao Yu and Rui Zhang and Kai Yang and Michihiro Yasunaga and Dongxu Wang and Zifan Li and James Ma and Irene Li and Qingning Yao and Shanelle Roman and Zilin Zhang and Dragomir Radev},
      year={2019},
      eprint={1809.08887},
      archivePrefix={arXiv},
      primaryClass={cs.CL},
      url={https://arxiv.org/abs/1809.08887}, 
}

@misc{wei2024measuringshortformfactualitylarge,
      title={Measuring short-form factuality in large language models}, 
      author={Jason Wei and Nguyen Karina and Hyung Won Chung and Yunxin Joy Jiao and Spencer Papay and Amelia Glaese and John Schulman and William Fedus},
      year={2024},
      eprint={2411.04368},
      archivePrefix={arXiv},
      primaryClass={cs.CL},
      url={https://arxiv.org/abs/2411.04368}, 
}

@misc{hendrycks2021measuringmathematicalproblemsolving,
      title={Measuring Mathematical Problem Solving With the MATH Dataset}, 
      author={Dan Hendrycks and Collin Burns and Saurav Kadavath and Akul Arora and Steven Basart and Eric Tang and Dawn Song and Jacob Steinhardt},
      year={2021},
      eprint={2103.03874},
      archivePrefix={arXiv},
      primaryClass={cs.LG},
      url={https://arxiv.org/abs/2103.03874}, 
}

@misc{kasai2024realtimeqawhatsanswer,
      title={RealTime QA: What's the Answer Right Now?}, 
      author={Jungo Kasai and Keisuke Sakaguchi and Yoichi Takahashi and Ronan Le Bras and Akari Asai and Xinyan Yu and Dragomir Radev and Noah A. Smith and Yejin Choi and Kentaro Inui},
      year={2024},
      eprint={2207.13332},
      archivePrefix={arXiv},
      primaryClass={cs.CL},
      url={https://arxiv.org/abs/2207.13332}, 
}

@misc{zhong2017seq2sqlgeneratingstructuredqueries,
      title={Seq2SQL: Generating Structured Queries from Natural Language using Reinforcement Learning}, 
      author={Victor Zhong and Caiming Xiong and Richard Socher},
      year={2017},
      eprint={1709.00103},
      archivePrefix={arXiv},
      primaryClass={cs.CL},
      url={https://arxiv.org/abs/1709.00103}, 
}

@misc{yao2024taubenchbenchmarktoolagentuserinteraction,
      title={$\tau$-bench: A Benchmark for Tool-Agent-User Interaction in Real-World Domains}, 
      author={Shunyu Yao and Noah Shinn and Pedram Razavi and Karthik Narasimhan},
      year={2024},
      eprint={2406.12045},
      archivePrefix={arXiv},
      primaryClass={cs.AI},
      url={https://arxiv.org/abs/2406.12045}, 
}

@inproceedings{rastogi2020towards,
  title={Towards scalable multi-domain conversational agents: The schema-guided dialogue dataset},
  author={Rastogi, Abhinav and Zang, Xiaoxue and Sunkara, Srinivas and Gupta, Raghav and Khaitan, Pranav},
  booktitle={Proceedings of the AAAI conference on artificial intelligence},
  volume={34},
  pages={8689--8696},
  year={2020}
}

@inproceedings{patil2025the,
    title={The Berkeley Function Calling Leaderboard ({BFCL}): From Tool Use to Agentic Evaluation of Large Language Models},
    author={Shishir G Patil and Huanzhi Mao and Fanjia Yan and Charlie Cheng-Jie Ji and Vishnu Suresh and Ion Stoica and Joseph E. Gonzalez},
    booktitle={Forty-second International Conference on Machine Learning},
    year={2025},
    url={https://openreview.net/forum?id=2GmDdhBdDk}
}

@article{wang2022text,
  title={Text embeddings by weakly-supervised contrastive pre-training},
  author={Wang, Liang and Yang, Nan and Huang, Xiaolong and Jiao, Binxing and Yang, Linjun and Jiang, Daxin and Majumder, Rangan and Wei, Furu},
  journal={arXiv preprint arXiv:2212.03533},
  year={2022}
}

@misc{openai2024gpt4technicalreport,
      title={GPT-4 Technical Report}, 
      author={OpenAI and Josh Achiam and Steven Adler and Sandhini Agarwal and Lama Ahmad and Ilge Akkaya and Florencia Leoni Aleman and Diogo Almeida and Janko Altenschmidt and Sam Altman and Shyamal Anadkat and Red Avila and Igor Babuschkin and Suchir Balaji and Valerie Balcom and Paul Baltescu and Haiming Bao and Mohammad Bavarian and Jeff Belgum and Irwan Bello and Jake Berdine and Gabriel Bernadett-Shapiro and Christopher Berner and Lenny Bogdonoff and Oleg Boiko and Madelaine Boyd and Anna-Luisa Brakman and Greg Brockman and Tim Brooks and Miles Brundage and Kevin Button and Trevor Cai and Rosie Campbell and Andrew Cann and Brittany Carey and Chelsea Carlson and Rory Carmichael and Brooke Chan and Che Chang and Fotis Chantzis and Derek Chen and Sully Chen and Ruby Chen and Jason Chen and Mark Chen and Ben Chess and Chester Cho and Casey Chu and Hyung Won Chung and Dave Cummings and Jeremiah Currier and Yunxing Dai and Cory Decareaux and Thomas Degry and Noah Deutsch and Damien Deville and Arka Dhar and David Dohan and Steve Dowling and Sheila Dunning and Adrien Ecoffet and Atty Eleti and Tyna Eloundou and David Farhi and Liam Fedus and Niko Felix and Simón Posada Fishman and Juston Forte and Isabella Fulford and Leo Gao and Elie Georges and Christian Gibson and Vik Goel and Tarun Gogineni and Gabriel Goh and Rapha Gontijo-Lopes and Jonathan Gordon and Morgan Grafstein and Scott Gray and Ryan Greene and Joshua Gross and Shixiang Shane Gu and Yufei Guo and Chris Hallacy and Jesse Han and Jeff Harris and Yuchen He and Mike Heaton and Johannes Heidecke and Chris Hesse and Alan Hickey and Wade Hickey and Peter Hoeschele and Brandon Houghton and Kenny Hsu and Shengli Hu and Xin Hu and Joost Huizinga and Shantanu Jain and Shawn Jain and Joanne Jang and Angela Jiang and Roger Jiang and Haozhun Jin and Denny Jin and Shino Jomoto and Billie Jonn and Heewoo Jun and Tomer Kaftan and Łukasz Kaiser and Ali Kamali and Ingmar Kanitscheider and Nitish Shirish Keskar and Tabarak Khan and Logan Kilpatrick and Jong Wook Kim and Christina Kim and Yongjik Kim and Jan Hendrik Kirchner and Jamie Kiros and Matt Knight and Daniel Kokotajlo and Łukasz Kondraciuk and Andrew Kondrich and Aris Konstantinidis and Kyle Kosic and Gretchen Krueger and Vishal Kuo and Michael Lampe and Ikai Lan and Teddy Lee and Jan Leike and Jade Leung and Daniel Levy and Chak Ming Li and Rachel Lim and Molly Lin and Stephanie Lin and Mateusz Litwin and Theresa Lopez and Ryan Lowe and Patricia Lue and Anna Makanju and Kim Malfacini and Sam Manning and Todor Markov and Yaniv Markovski and Bianca Martin and Katie Mayer and Andrew Mayne and Bob McGrew and Scott Mayer McKinney and Christine McLeavey and Paul McMillan and Jake McNeil and David Medina and Aalok Mehta and Jacob Menick and Luke Metz and Andrey Mishchenko and Pamela Mishkin and Vinnie Monaco and Evan Morikawa and Daniel Mossing and Tong Mu and Mira Murati and Oleg Murk and David Mély and Ashvin Nair and Reiichiro Nakano and Rajeev Nayak and Arvind Neelakantan and Richard Ngo and Hyeonwoo Noh and Long Ouyang and Cullen O'Keefe and Jakub Pachocki and Alex Paino and Joe Palermo and Ashley Pantuliano and Giambattista Parascandolo and Joel Parish and Emy Parparita and Alex Passos and Mikhail Pavlov and Andrew Peng and Adam Perelman and Filipe de Avila Belbute Peres and Michael Petrov and Henrique Ponde de Oliveira Pinto and Michael and Pokorny and Michelle Pokrass and Vitchyr H. Pong and Tolly Powell and Alethea Power and Boris Power and Elizabeth Proehl and Raul Puri and Alec Radford and Jack Rae and Aditya Ramesh and Cameron Raymond and Francis Real and Kendra Rimbach and Carl Ross and Bob Rotsted and Henri Roussez and Nick Ryder and Mario Saltarelli and Ted Sanders and Shibani Santurkar and Girish Sastry and Heather Schmidt and David Schnurr and John Schulman and Daniel Selsam and Kyla Sheppard and Toki Sherbakov and Jessica Shieh and Sarah Shoker and Pranav Shyam and Szymon Sidor and Eric Sigler and Maddie Simens and Jordan Sitkin and Katarina Slama and Ian Sohl and Benjamin Sokolowsky and Yang Song and Natalie Staudacher and Felipe Petroski Such and Natalie Summers and Ilya Sutskever and Jie Tang and Nikolas Tezak and Madeleine B. Thompson and Phil Tillet and Amin Tootoonchian and Elizabeth Tseng and Preston Tuggle and Nick Turley and Jerry Tworek and Juan Felipe Cerón Uribe and Andrea Vallone and Arun Vijayvergiya and Chelsea Voss and Carroll Wainwright and Justin Jay Wang and Alvin Wang and Ben Wang and Jonathan Ward and Jason Wei and CJ Weinmann and Akila Welihinda and Peter Welinder and Jiayi Weng and Lilian Weng and Matt Wiethoff and Dave Willner and Clemens Winter and Samuel Wolrich and Hannah Wong and Lauren Workman and Sherwin Wu and Jeff Wu and Michael Wu and Kai Xiao and Tao Xu and Sarah Yoo and Kevin Yu and Qiming Yuan and Wojciech Zaremba and Rowan Zellers and Chong Zhang and Marvin Zhang and Shengjia Zhao and Tianhao Zheng and Juntang Zhuang and William Zhuk and Barret Zoph},
      year={2024},
      eprint={2303.08774},
      archivePrefix={arXiv},
      primaryClass={cs.CL},
      url={https://arxiv.org/abs/2303.08774}, 
}

@misc{comanici2025gemini25pushingfrontier,
      title={Gemini 2.5: Pushing the Frontier with Advanced Reasoning, Multimodality, Long Context, and Next Generation Agentic Capabilities}, 
      author={Gheorghe Comanici and Eric Bieber and Mike Schaekermann and Ice Pasupat and Noveen Sachdeva and Inderjit Dhillon and Marcel Blistein and Ori Ram and Dan Zhang and Evan Rosen and Luke Marris and Sam Petulla and Colin Gaffney and Asaf Aharoni and Nathan Lintz and Tiago Cardal Pais and Henrik Jacobsson and Idan Szpektor and Nan-Jiang Jiang and Krishna Haridasan and Ahmed Omran and Nikunj Saunshi and Dara Bahri and Gaurav Mishra and Eric Chu and Toby Boyd and Brad Hekman and Aaron Parisi and Chaoyi Zhang and Kornraphop Kawintiranon and Tania Bedrax-Weiss and Oliver Wang and Ya Xu and Ollie Purkiss and Uri Mendlovic and Ilaï Deutel and Nam Nguyen and Adam Langley and Flip Korn and Lucia Rossazza and Alexandre Ramé and Sagar Waghmare and Helen Miller and Nathan Byrd and Ashrith Sheshan and Raia Hadsell and Sangnie Bhardwaj and Pawel Janus and Tero Rissa and Dan Horgan and Alvin Abdagic and Lior Belenki and James Allingham and Anima Singh and Theo Guidroz and Srivatsan Srinivasan and Herman Schmit and Kristen Chiafullo and Andre Elisseeff and Nilpa Jha and Prateek Kolhar and Leonard Berrada and Frank Ding and Xiance Si and Shrestha Basu Mallick and Franz Och and Sofia Erell and Eric Ni and Tejasi Latkar and Sherry Yang and Petar Sirkovic and Ziqiang Feng and Robert Leland and Rachel Hornung and Gang Wu and Charles Blundell and Hamidreza Alvari and Po-Sen Huang and Cathy Yip and Sanja Deur and Li Liu and Gabriela Surita and Pablo Duque and Dima Damen and Johnson Jia and Arthur Guez and Markus Mircea and Animesh Sinha and Alberto Magni and Paweł Stradomski and Tal Marian and Vlado Galić and Wenhu Chen and Hisham Husain and Achintya Singhal and Dominik Grewe and François-Xavier Aubet and Shuang Song and Lorenzo Blanco and Leland Rechis and Lewis Ho and Rich Munoz and Kelvin Zheng and Jessica Hamrick and Kevin Mather and Hagai Taitelbaum and Eliza Rutherford and Yun Lei and Kuangyuan Chen and Anand Shukla and Erica Moreira and Eric Doi and Berivan Isik and Nir Shabat and Dominika Rogozińska and Kashyap Kolipaka and Jason Chang and Eugen Vušak and Srinivasan Venkatachary and Shadi Noghabi and Tarun Bharti and Younghoon Jun and Aleksandr Zaks and Simon Green and Jeshwanth Challagundla and William Wong and Muqthar Mohammad and Dean Hirsch and Yong Cheng and Iftekhar Naim and Lev Proleev and Damien Vincent and Aayush Singh and Maxim Krikun and Dilip Krishnan and Zoubin Ghahramani and Aviel Atias and Rajeev Aggarwal and Christo Kirov and Dimitrios Vytiniotis and Christy Koh and Alexandra Chronopoulou and Pawan Dogra and Vlad-Doru Ion and Gladys Tyen and Jason Lee and Felix Weissenberger and Trevor Strohman and Ashwin Balakrishna and Jack Rae and Marko Velic and Raoul de Liedekerke and Oded Elyada and Wentao Yuan and Canoee Liu and Lior Shani and Sergey Kishchenko and Bea Alessio and Yandong Li and Richard Song and Sam Kwei and Orion Jankowski and Aneesh Pappu and Youhei Namiki and Yenai Ma and Nilesh Tripuraneni and Colin Cherry and Marissa Ikonomidis and Yu-Cheng Ling and Colin Ji and Beka Westberg and Auriel Wright and Da Yu and David Parkinson and Swaroop Ramaswamy and Jerome Connor and Soheil Hassas Yeganeh and Snchit Grover and George Kenwright and Lubo Litchev and Chris Apps and Alex Tomala and Felix Halim and Alex Castro-Ros and Zefei Li and Anudhyan Boral and Pauline Sho and Michal Yarom and Eric Malmi and David Klinghoffer and Rebecca Lin and Alan Ansell and Pradeep Kumar S and Shubin Zhao and Siqi Zuo and Adam Santoro and Heng-Tze Cheng and Solomon Demmessie and Yuchi Liu and Nicole Brichtova and Allie Culp and Nathaniel Braun and Dan Graur and Will Ng and Nikhil Mehta and Aaron Phillips and Patrik Sundberg and Varun Godbole and Fangyu Liu and Yash Katariya and David Rim and Mojtaba Seyedhosseini and Sean Ammirati and Jonas Valfridsson and Mahan Malihi and Timothy Knight and Andeep Toor and Thomas Lampe and Abe Ittycheriah and Lewis Chiang and Chak Yeung and Alexandre Fréchette and Jinmeng Rao and Huisheng Wang and Himanshu Srivastava and Richard Zhang and Rocky Rhodes and Ariel Brand and Dean Weesner and Ilya Figotin and Felix Gimeno and Rachana Fellinger and Pierre Marcenac and José Leal and Eyal Marcus and Victor Cotruta and Rodrigo Cabrera and Sheryl Luo and Dan Garrette and Vera Axelrod and Sorin Baltateanu and David Barker and Dongkai Chen and Horia Toma and Ben Ingram and Jason Riesa and Chinmay Kulkarni and Yujing Zhang and Hongbin Liu and Chao Wang and Martin Polacek and Will Wu and Kai Hui and Adrian N Reyes and Yi Su and Megan Barnes and Ishaan Malhi and Anfal Siddiqui and Qixuan Feng and Mihai Damaschin and Daniele Pighin and Andreas Steiner and Samuel Yang and Ramya Sree Boppana and Simeon Ivanov and Arun Kandoor and Aditya Shah and Asier Mujika and Da Huang and Christopher A. Choquette-Choo and Mohak Patel and Tianhe Yu and Toni Creswell and Jerry and Liu and Catarina Barros and Yasaman Razeghi and Aurko Roy and Phil Culliton and Binbin Xiong and Jiaqi Pan and Thomas Strohmann and Tolly Powell and Babi Seal and Doug DeCarlo and Pranav Shyam and Kaan Katircioglu and Xuezhi Wang and Cassidy Hardin and Immanuel Odisho and Josef Broder and Oscar Chang and Arun Nair and Artem Shtefan and Maura O'Brien and Manu Agarwal and Sahitya Potluri and Siddharth Goyal and Amit Jhindal and Saksham Thakur and Yury Stuken and James Lyon and Kristina Toutanova and Fangxiaoyu Feng and Austin Wu and Ben Horn and Alek Wang and Alex Cullum and Gabe Taubman and Disha Shrivastava and Chongyang Shi and Hamish Tomlinson and Roma Patel and Tao Tu and Ada Maksutaj Oflazer and Francesco Pongetti and Mingyao Yang and Adrien Ali Taïga and Vincent Perot and Nuo Wang Pierse and Feng Han and Yoel Drori and Iñaki Iturrate and Ayan Chakrabarti and Legg Yeung and Dave Dopson and Yi-ting Chen and Apoorv Kulshreshtha and Tongfei Guo and Philip Pham and Tal Schuster and Junquan Chen and Alex Polozov and Jinwei Xing and Huanjie Zhou and Praneeth Kacham and Doron Kukliansky and Antoine Miech and Sergey Yaroshenko and Ed Chi and Sholto Douglas and Hongliang Fei and Mathieu Blondel and Preethi Myla and Lior Madmoni and Xing Wu and Daniel Keysers and Kristian Kjems and Isabela Albuquerque and Lijun Yu and Joel D'sa and Michelle Plantan and Vlad Ionescu and Jaume Sanchez Elias and Abhirut Gupta and Manish Reddy Vuyyuru and Fred Alcober and Tong Zhou and Kaiyang Ji and Florian Hartmann and Subha Puttagunta and Hugo Song and Ehsan Amid and Anca Stefanoiu and Andrew Lee and Paul Pucciarelli and Emma Wang and Amit Raul and Slav Petrov and Isaac Tian and Valentin Anklin and Nana Nti and Victor Gomes and Max Schumacher and Grace Vesom and Alex Panagopoulos and Konstantinos Bousmalis and Daniel Andor and Josh Jacob and Yuan Zhang and Bill Rosgen and Matija Kecman and Matthew Tung and Alexandra Belias and Noah Goodman and Paul Covington and Brian Wieder and Nikita Saxena and Elnaz Davoodi and Muhuan Huang and Sharath Maddineni and Vincent Roulet and Folawiyo Campbell-Ajala and Pier Giuseppe Sessa and Xintian and Wu and Guangda Lai and Paul Collins and Alex Haig and Vytenis Sakenas and Xiaowei Xu and Marissa Giustina and Laurent El Shafey and Pichi Charoenpanit and Shefali Garg and Joshua Ainslie and Boone Severson and Montse Gonzalez Arenas and Shreya Pathak and Sujee Rajayogam and Jie Feng and Michiel Bakker and Sheng Li and Nevan Wichers and Jamie Rogers and Xinyang Geng and Yeqing Li and Rolf Jagerman and Chao Jia and Nadav Olmert and David Sharon and Matthew Mauger and Sandeep Mariserla and Hongxu Ma and Megha Mohabey and Kyuyeun Kim and Alek Andreev and Scott Pollom and Juliette Love and Vihan Jain and Priyanka Agrawal and Yannick Schroecker and Alisa Fortin and Manfred Warmuth and Ji Liu and Andrew Leach and Irina Blok and Ganesh Poomal Girirajan and Roee Aharoni and Benigno Uria and Andrei Sozanschi and Dan Goldberg and Lucian Ionita and Marco Tulio Ribeiro and Martin Zlocha and Vighnesh Birodkar and Sami Lachgar and Liangzhe Yuan and Himadri Choudhury and Matt Ginsberg and Fei Zheng and Gregory Dibb and Emily Graves and Swachhand Lokhande and Gabriel Rasskin and George-Cristian Muraru and Corbin Quick and Sandeep Tata and Pierre Sermanet and Aditya Chawla and Itay Karo and Yan Wang and Susan Zhang and Orgad Keller and Anca Dragan and Guolong Su and Ian Chou and Xi Liu and Yiqing Tao and Shruthi Prabhakara and Marc Wilson and Ruibo Liu and Shibo Wang and Georgie Evans and David Du and Alfonso Castaño and Gautam Prasad and Mona El Mahdy and Sebastian Gerlach and Machel Reid and Jarrod Kahn and Amir Zait and Thanumalayan Sankaranarayana Pillai and Thatcher Ulrich and Guanyu Wang and Jan Wassenberg and Efrat Farkash and Kiran Yalasangi and Congchao Wang and Maria Bauza and Simon Bucher and Ting Liu and Jun Yan and Gary Leung and Vikas Sindhwani and Parker Barnes and Avi Singh and Ivan Jurin and Jichuan Chang and Niket Kumar Bhumihar and Sivan Eiger and Gui Citovsky and Ben Withbroe and Zhang Li and Siyang Xue and Niccolò Dal Santo and Georgi Stoyanov and Yves Raimond and Steven Zheng and Yilin Gao and Vít Listík and Sławek Kwasiborski and Rachel Saputro and Adnan Ozturel and Ganesh Mallya and Kushal Majmundar and Ross West and Paul Caron and Jinliang Wei and Lluis Castrejon and Sharad Vikram and Deepak Ramachandran and Nikhil Dhawan and Jiho Park and Sara Smoot and George van den Driessche and Yochai Blau and Chase Malik and Wei Liang and Roy Hirsch and Cicero Nogueira dos Santos and Eugene Weinstein and Aäron van den Oord and Sid Lall and Nicholas FitzGerald and Zixuan Jiang and Xuan Yang and Dale Webster and Ali Elqursh and Aedan Pope and Georges Rotival and David Raposo and Wanzheng Zhu and Jeff Dean and Sami Alabed and Dustin Tran and Arushi Gupta and Zach Gleicher and Jessica Austin and Edouard Rosseel and Megh Umekar and Dipanjan Das and Yinghao Sun and Kai Chen and Karolis Misiunas and Xiang Zhou and Yixian Di and Alyssa Loo and Josh Newlan and Bo Li and Vinay Ramasesh and Ying Xu and Alex Chen and Sudeep Gandhe and Radu Soricut and Nikita Gupta and Shuguang Hu and Seliem El-Sayed and Xavier Garcia and Idan Brusilovsky and Pu-Chin Chen and Andrew Bolt and Lu Huang and Alex Gurney and Zhiying Zhang and Alexander Pritzel and Jarek Wilkiewicz and Bryan Seybold and Bhargav Kanagal Shamanna and Felix Fischer and Josef Dean and Karan Gill and Ross Mcilroy and Abhishek Bhowmick and Jeremy Selier and Antoine Yang and Derek Cheng and Vladimir Magay and Jie Tan and Dhriti Varma and Christian Walder and Tomas Kocisky and Ryo Nakashima and Paul Natsev and Mike Kwong and Ionel Gog and Chiyuan Zhang and Sander Dieleman and Thomas Jimma and Andrey Ryabtsev and Siddhartha Brahma and David Steiner and Dayou Du and Ante Žužul and Mislav Žanić and Mukund Raghavachari and Willi Gierke and Zeyu Zheng and Dessie Petrova and Yann Dauphin and Yuchuan Liu and Ido Kessler and Steven Hand and Chris Duvarney and Seokhwan Kim and Hyo Lee and Léonard Hussenot and Jeffrey Hui and Josh Smith and Deepali Jain and Jiawei Xia and Gaurav Singh Tomar and Keyvan Amiri and Du Phan and Fabian Fuchs and Tobias Weyand and Nenad Tomasev and Alexandra Cordell and Xin Liu and Jonathan Mallinson and Pankaj Joshi and Andy Crawford and Arun Suggala and Steve Chien and Nick Fernando and Mariella Sanchez-Vargas and Duncan Williams and Phil Crone and Xiyang Luo and Igor Karpov and Jyn Shan and Terry Thurk and Robin Strudel and Paul Voigtlaender and Piyush Patil and Tim Dozat and Ali Khodaei and Sahil Singla and Piotr Ambroszczyk and Qiyin Wu and Yifan Chang and Brian Roark and Chaitra Hegde and Tianli Ding and Angelos Filos and Zhongru Wu and André Susano Pinto and Shuang Liu and Saarthak Khanna and Aditya Pandey and Siobhan Mcloughlin and Qiujia Li and Sam Haves and Allan Zhou and Elena Buchatskaya and Isabel Leal and Peter de Boursac and Nami Akazawa and Nina Anderson and Terry Chen and Krishna Somandepalli and Chen Liang and Sheela Goenka and Stephanie Winkler and Alexander Grushetsky and Yifan Ding and Jamie Smith and Fan Ye and Jordi Pont-Tuset and Eric Li and Ruichao Li and Tomer Golany and Dawid Wegner and Tao Jiang and Omer Barak and Yuan Shangguan and Eszter Vértes and Renee Wong and Jörg Bornschein and Alex Tudor and Michele Bevilacqua and Tom Schaul and Ankit Singh Rawat and Yang Zhao and Kyriakos Axiotis and Lei Meng and Cory McLean and Jonathan Lai and Jennifer Beattie and Nate Kushman and Yaxin Liu and Blair Kutzman and Fiona Lang and Jingchen Ye and Praneeth Netrapalli and Pushkar Mishra and Myriam Khan and Megha Goel and Rob Willoughby and David Tian and Honglei Zhuang and JD Chen and Zak Tsai and Tasos Kementsietsidis and Arjun Khare and James Keeling and Keyang Xu and Nathan Waters and Florent Altché and Ashok Popat and Bhavishya Mittal and David Saxton and Dalia El Badawy and Michael Mathieu and Zheng Zheng and Hao Zhou and Nishant Ranka and Richard Shin and Qingnan Duan and Tim Salimans and Ioana Mihailescu and Uri Shaham and Ming-Wei Chang and Yannis Assael and Nishanth Dikkala and Martin Izzard and Vincent Cohen-Addad and Cat Graves and Vlad Feinberg and Grace Chung and DJ Strouse and Danny Karmon and Sahand Sharifzadeh and Zoe Ashwood and Khiem Pham and Jon Blanton and Alex Vasiloff and Jarred Barber and Mark Geller and Aurick Zhou and Fedir Zubach and Tzu-Kuo Huang and Lei Zhang and Himanshu Gupta and Matt Young and Julia Proskurnia and Ronny Votel and Valentin Gabeur and Gabriel Barcik and Aditya Tripathi and Hongkun Yu and Geng Yan and Beer Changpinyo and Filip Pavetić and Amy Coyle and Yasuhisa Fujii and Jorge Gonzalez Mendez and Tianhao Zhou and Harish Rajamani and Blake Hechtman and Eddie Cao and Da-Cheng Juan and Yi-Xuan Tan and Valentin Dalibard and Yilun Du and Natalie Clay and Kaisheng Yao and Wenhao Jia and Dimple Vijaykumar and Yuxiang Zhou and Xinyi Bai and Wei-Chih Hung and Steven Pecht and Georgi Todorov and Nikhil Khadke and Pramod Gupta and Preethi Lahoti and Arnaud Autef and Karthik Duddu and James Lee-Thorp and Alexander Bykovsky and Tautvydas Misiunas and Sebastian Flennerhag and Santhosh Thangaraj and Jed McGiffin and Zack Nado and Markus Kunesch and Andreas Noever and Amir Hertz and Marco Liang and Victor Stone and Evan Palmer and Samira Daruki and Arijit Pramanik and Siim Põder and Austin Kyker and Mina Khan and Evgeny Sluzhaev and Marvin Ritter and Avraham Ruderman and Wenlei Zhou and Chirag Nagpal and Kiran Vodrahalli and George Necula and Paul Barham and Ellie Pavlick and Jay Hartford and Izhak Shafran and Long Zhao and Maciej Mikuła and Tom Eccles and Hidetoshi Shimokawa and Kanav Garg and Luke Vilnis and Hanwen Chen and Ilia Shumailov and Kuang-Huei Lee and Abdelrahman Abdelhamed and Meiyan Xie and Vered Cohen and Ester Hlavnova and Dan Malkin and Chawin Sitawarin and James Lottes and Pauline Coquinot and Tianli Yu and Sandeep Kumar and Jingwei Zhang and Aroma Mahendru and Zafarali Ahmed and James Martens and Tao Chen and Aviel Boag and Daiyi Peng and Coline Devin and Arseniy Klimovskiy and Mary Phuong and Danny Vainstein and Jin Xie and Bhuvana Ramabhadran and Nathan Howard and Xinxin Yu and Gitartha Goswami and Jingyu Cui and Sam Shleifer and Mario Pinto and Chih-Kuan Yeh and Ming-Hsuan Yang and Sara Javanmardi and Dan Ethier and Chace Lee and Jordi Orbay and Suyog Kotecha and Carla Bromberg and Pete Shaw and James Thornton and Adi Gerzi Rosenthal and Shane Gu and Matt Thomas and Ian Gemp and Aditya Ayyar and Asahi Ushio and Aarush Selvan and Joel Wee and Chenxi Liu and Maryam Majzoubi and Weiren Yu and Jake Abernethy and Tyler Liechty and Renke Pan and Hoang Nguyen and Qiong and Hu and Sarah Perrin and Abhinav Arora and Emily Pitler and Weiyi Wang and Kaushik Shivakumar and Flavien Prost and Ben Limonchik and Jing Wang and Yi Gao and Timothee Cour and Shyamal Buch and Huan Gui and Maria Ivanova and Philipp Neubeck and Kelvin Chan and Lucy Kim and Huizhong Chen and Naman Goyal and Da-Woon Chung and Lu Liu and Yao Su and Anastasia Petrushkina and Jiajun Shen and Armand Joulin and Yuanzhong Xu and Stein Xudong Lin and Yana Kulizhskaya and Ciprian Chelba and Shobha Vasudevan and Eli Collins and Vasilisa Bashlovkina and Tony Lu and Doug Fritz and Jongbin Park and Yanqi Zhou and Chen Su and Richard Tanburn and Mikhail Sushkov and Mitchelle Rasquinha and Jinning Li and Jennifer Prendki and Yiming Li and Pallavi LV and Shriya Sharma and Hen Fitoussi and Hui Huang and Andrew Dai and Phuong Dao and Mike Burrows and Henry Prior and Danfeng Qin and Golan Pundak and Lars Lowe Sjoesund and Art Khurshudov and Zhenkai Zhu and Albert Webson and Elizabeth Kemp and Tat Tan and Saurabh Agrawal and Susie Sargsyan and Liqun Cheng and Jim Stephan and Tom Kwiatkowski and David Reid and Arunkumar Byravan and Assaf Hurwitz Michaely and Nicolas Heess and Luowei Zhou and Sonam Goenka and Viral Carpenter and Anselm Levskaya and Bo Wang and Reed Roberts and Rémi Leblond and Sharat Chikkerur and Stav Ginzburg and Max Chang and Robert Riachi and Chuqiao and Xu and Zalán Borsos and Michael Pliskin and Julia Pawar and Morgane Lustman and Hannah Kirkwood and Ankit Anand and Aditi Chaudhary and Norbert Kalb and Kieran Milan and Sean Augenstein and Anna Goldie and Laurel Prince and Karthik Raman and Yanhua Sun and Vivian Xia and Aaron Cohen and Zhouyuan Huo and Josh Camp and Seher Ellis and Lukas Zilka and David Vilar Torres and Lisa Patel and Sho Arora and Betty Chan and Jonas Adler and Kareem Ayoub and Jacky Liang and Fayaz Jamil and Jiepu Jiang and Simon Baumgartner and Haitian Sun and Yael Karov and Yaroslav Akulov and Hui Zheng and Irene Cai and Claudio Fantacci and James Rubin and Alex Rav Acha and Mengchao Wang and Nina D'Souza and Rohit Sathyanarayana and Shengyang Dai and Simon Rowe and Andrey Simanovsky and Omer Goldman and Yuheng Kuang and Xiaoyue Pan and Andrew Rosenberg and Tania Rojas-Esponda and Praneet Dutta and Amy Zeng and Irina Jurenka and Greg Farquhar and Yamini Bansal and Shariq Iqbal and Becca Roelofs and Ga-Young Joung and Parker Beak and Changwan Ryu and Ryan Poplin and Yan Wu and Jean-Baptiste Alayrac and Senaka Buthpitiya and Olaf Ronneberger and Caleb Habtegebriel and Wei Li and Paul Cavallaro and Aurora Wei and Guy Bensky and Timo Denk and Harish Ganapathy and Jeff Stanway and Pratik Joshi and Francesco Bertolini and Jessica Lo and Olivia Ma and Zachary Charles and Geta Sampemane and Himanshu Sahni and Xu Chen and Harry Askham and David Gaddy and Peter Young and Jiewen Tan and Matan Eyal and Arthur Bražinskas and Li Zhong and Zhichun Wu and Mark Epstein and Kai Bailey and Andrew Hard and Kamyu Lee and Sasha Goldshtein and Alex Ruiz and Mohammed Badawi and Matthias Lochbrunner and JK Kearns and Ashley Brown and Fabio Pardo and Theophane Weber and Haichuan Yang and Pan-Pan Jiang and Berkin Akin and Zhao Fu and Marcus Wainwright and Chi Zou and Meenu Gaba and Pierre-Antoine Manzagol and Wendy Kan and Yang Song and Karina Zainullina and Rui Lin and Jeongwoo Ko and Salil Deshmukh and Apoorv Jindal and James Svensson and Divya Tyam and Heri Zhao and Christine Kaeser-Chen and Scott Baird and Pooya Moradi and Jamie Hall and Qiuchen Guo and Vincent Tsang and Bowen Liang and Fernando Pereira and Suhas Ganesh and Ivan Korotkov and Jakub Adamek and Sridhar Thiagarajan and Vinh Tran and Charles Chen and Chris Tar and Sanil Jain and Ishita Dasgupta and Taylan Bilal and David Reitter and Kai Zhao and Giulia Vezzani and Yasmin Gehman and Pulkit Mehta and Lauren Beltrone and Xerxes Dotiwalla and Sergio Guadarrama and Zaheer Abbas and Stefani Karp and Petko Georgiev and Chun-Sung Ferng and Marc Brockschmidt and Liqian Peng and Christoph Hirnschall and Vikas Verma and Yingying Bi and Ying Xiao and Avigail Dabush and Kelvin Xu and Phil Wallis and Randall Parker and Qifei Wang and Yang Xu and Ilkin Safarli and Dinesh Tewari and Yin Zhang and Seungyeon Kim and Andrea Gesmundo and Mackenzie Thomas and Sergey Levi and Ahmed Chowdhury and Kanishka Rao and Peter Garst and Sam Conway-Rahman and Helen Ran and Kay McKinney and Zhisheng Xiao and Wenhao Yu and Rohan Agrawal and Axel Stjerngren and Catalin Ionescu and Jingjing Chen and Vivek Sharma and Justin Chiu and Fei Liu and Ken Franko and Clayton Sanford and Xingyu Cai and Paul Michel and Sanjay Ganapathy and Jane Labanowski and Zachary Garrett and Ben Vargas and Sean Sun and Bryan Gale and Thomas Buschmann and Guillaume Desjardins and Nimesh Ghelani and Palak Jain and Mudit Verma and Chulayuth Asawaroengchai and Julian Eisenschlos and Jitendra Harlalka and Hideto Kazawa and Don Metzler and Joshua Howland and Ying Jian and Jake Ades and Viral Shah and Tynan Gangwani and Seungji Lee and Roman Ring and Steven M. Hernandez and Dean Reich and Amer Sinha and Ashutosh Sathe and Joe Kovac and Ashleah Gill and Ajay Kannan and Andrea D'olimpio and Martin Sevenich and Jay Whang and Been Kim and Khe Chai Sim and Jilin Chen and Jiageng Zhang and Shuba Lall and Yossi Matias and Bill Jia and Abe Friesen and Sara Nasso and Ashish Thapliyal and Bryan Perozzi and Ting Yu and Anna Shekhawat and Safeen Huda and Peter Grabowski and Eric Wang and Ashwin Sreevatsa and Hilal Dib and Mehadi Hassen and Parker Schuh and Vedrana Milutinovic and Chris Welty and Michael Quinn and Ali Shah and Bangju Wang and Gabe Barth-Maron and Justin Frye and Natalie Axelsson and Tao Zhu and Yukun Ma and Irene Giannoumis and Hanie Sedghi and Chang Ye and Yi Luan and Kevin Aydin and Bilva Chandra and Vivek Sampathkumar and Ronny Huang and Victor Lavrenko and Ahmed Eleryan and Zhi Hong and Steven Hansen and Sara Mc Carthy and Bidisha Samanta and Domagoj Ćevid and Xin Wang and Fangtao Li and Michael Voznesensky and Matt Hoffman and Andreas Terzis and Vikash Sehwag and Gil Fidel and Luheng He and Mu Cai and Yanzhang He and Alex Feng and Martin Nikoltchev and Samrat Phatale and Jason Chase and Rory Lawton and Ming Zhang and Tom Ouyang and Manuel Tragut and Mehdi Hafezi Manshadi and Arjun Narayanan and Jiaming Shen and Xu Gao and Tolga Bolukbasi and Nick Roy and Xin Li and Daniel Golovin and Liviu Panait and Zhen Qin and Guangxing Han and Thomas Anthony and Sneha Kudugunta and Viorica Patraucean and Aniket Ray and Xinyun Chen and Xiaochen Yang and Tanuj Bhatia and Pranav Talluri and Alex Morris and Andrija Ražnatović and Bethanie Brownfield and James An and Sheng Peng and Patrick Kane and Ce Zheng and Nico Duduta and Joshua Kessinger and James Noraky and Siqi Liu and Keran Rong and Petar Veličković and Keith Rush and Alex Goldin and Fanny Wei and Shiva Mohan Reddy Garlapati and Caroline Pantofaru and Okwan Kwon and Jianmo Ni and Eric Noland and Julia Di Trapani and Françoise Beaufays and Abhijit Guha Roy and Yinlam Chow and Aybuke Turker and Geoffrey Cideron and Lantao Mei and Jon Clark and Qingyun Dou and Matko Bošnjak and Ralph Leith and Yuqing Du and Amir Yazdanbakhsh and Milad Nasr and Chester Kwak and Suraj Satishkumar Sheth and Alex Kaskasoli and Ankesh Anand and Balaji Lakshminarayanan and Sammy Jerome and David Bieber and Chun-Te Chu and Alexandre Senges and Tianxiao Shen and Mukund Sridhar and Ndaba Ndebele and Benjamin Beyret and Shakir Mohamed and Mia Chen and Markus Freitag and Jiaxian Guo and Luyang Liu and Paul Roit and Heng Chen and Shen Yan and Tom Stone and JD Co-Reyes and Jeremy Cole and Salvatore Scellato and Shekoofeh Azizi and Hadi Hashemi and Alicia Jin and Anand Iyer and Marcella Valentine and András György and Arun Ahuja and Daniel Hernandez Diaz and Chen-Yu Lee and Nathan Clement and Weize Kong and Drew Garmon and Ishaan Watts and Kush Bhatia and Khyatti Gupta and Matt Miecnikowski and Hugo Vallet and Ankur Taly and Edward Loper and Saket Joshi and James Atwood and Jo Chick and Mark Collier and Fotis Iliopoulos and Ryan Trostle and Beliz Gunel and Ramiro Leal-Cavazos and Arnar Mar Hrafnkelsson and Michael Guzman and Xiaoen Ju and Andy Forbes and Jesse Emond and Kushal Chauhan and Ben Caine and Li Xiao and Wenjun Zeng and Alexandre Moufarek and Daniel Murphy and Maya Meng and Nitish Gupta and Felix Riedel and Anil Das and Elijah Lawal and Shashi Narayan and Tiberiu Sosea and James Swirhun and Linda Friso and Behnam Neyshabur and Jing Lu and Sertan Girgin and Michael Wunder and Edouard Yvinec and Aroonalok Pyne and Victor Carbune and Shruti Rijhwani and Yang Guo and Tulsee Doshi and Anton Briukhov and Max Bain and Ayal Hitron and Xuanhui Wang and Ashish Gupta and Ke Chen and Cosmo Du and Weiyang Zhang and Dhruv Shah and Arjun Akula and Max Dylla and Ashyana Kachra and Weicheng Kuo and Tingting Zou and Lily Wang and Luyao Xu and Jifan Zhu and Justin Snyder and Sachit Menon and Orhan Firat and Igor Mordatch and Yuan Yuan and Natalia Ponomareva and Rory Blevins and Lawrence Moore and Weijun Wang and Phil Chen and Martin Scholz and Artur Dwornik and Jason Lin and Sicheng Li and Diego Antognini and Te I and Xiaodan Song and Matt Miller and Uday Kalra and Adam Raveret and Oscar Akerlund and Felix Wu and Andrew Nystrom and Namrata Godbole and Tianqi Liu and Hannah DeBalsi and Jewel Zhao and Buhuang Liu and Avi Caciularu and Lauren Lax and Urvashi Khandelwal and Victoria Langston and Eric Bailey and Silvio Lattanzi and Yufei Wang and Neel Kovelamudi and Sneha Mondal and Guru Guruganesh and Nan Hua and Ofir Roval and Paweł Wesołowski and Rishikesh Ingale and Jonathan Halcrow and Tim Sohn and Christof Angermueller and Bahram Raad and Eli Stickgold and Eva Lu and Alec Kosik and Jing Xie and Timothy Lillicrap and Austin Huang and Lydia Lihui Zhang and Dominik Paulus and Clement Farabet and Alex Wertheim and Bing Wang and Rishabh Joshi and Chu-ling Ko and Yonghui Wu and Shubham Agrawal and Lily Lin and XiangHai Sheng and Peter Sung and Tyler Breland-King and Christina Butterfield and Swapnil Gawde and Sumeet Singh and Qiao Zhang and Raj Apte and Shilpa Shetty and Adrian Hutter and Tao Li and Elizabeth Salesky and Federico Lebron and Jonni Kanerva and Michela Paganini and Arthur Nguyen and Rohith Vallu and Jan-Thorsten Peter and Sarmishta Velury and David Kao and Jay Hoover and Anna Bortsova and Colton Bishop and Shoshana Jakobovits and Alessandro Agostini and Alekh Agarwal and Chang Liu and Charles Kwong and Sasan Tavakkol and Ioana Bica and Alex Greve and Anirudh GP and Jake Marcus and Le Hou and Tom Duerig and Rivka Moroshko and Dave Lacey and Andy Davis and Julien Amelot and Guohui Wang and Frank Kim and Theofilos Strinopoulos and Hui Wan and Charline Le Lan and Shankar Krishnan and Haotian Tang and Peter Humphreys and Junwen Bai and Idan Heimlich Shtacher and Diego Machado and Chenxi Pang and Ken Burke and Dangyi Liu and Renga Aravamudhan and Yue Song and Ed Hirst and Abhimanyu Singh and Brendan Jou and Liang Bai and Francesco Piccinno and Chuyuan Kelly Fu and Robin Alazard and Barak Meiri and Daniel Winter and Charlie Chen and Mingda Zhang and Jens Heitkaemper and John Lambert and Jinhyuk Lee and Alexander Frömmgen and Sergey Rogulenko and Pranav Nair and Paul Niemczyk and Anton Bulyenov and Bibo Xu and Hadar Shemtov and Morteza Zadimoghaddam and Serge Toropov and Mateo Wirth and Hanjun Dai and Sreenivas Gollapudi and Daniel Zheng and Alex Kurakin and Chansoo Lee and Kalesha Bullard and Nicolas Serrano and Ivana Balazevic and Yang Li and Johan Schalkwyk and Mark Murphy and Mingyang Zhang and Kevin Sequeira and Romina Datta and Nishant Agrawal and Charles Sutton and Nithya Attaluri and Mencher Chiang and Wael Farhan and Gregory Thornton and Kate Lin and Travis Choma and Hung Nguyen and Kingshuk Dasgupta and Dirk Robinson and Iulia Comşa and Michael Riley and Arjun Pillai and Basil Mustafa and Ben Golan and Amir Zandieh and Jean-Baptiste Lespiau and Billy Porter and David Ross and Sujeevan Rajayogam and Mohit Agarwal and Subhashini Venugopalan and Bobak Shahriari and Qiqi Yan and Hao Xu and Taylor Tobin and Pavel Dubov and Hongzhi Shi and Adrià Recasens and Anton Kovsharov and Sebastian Borgeaud and Lucio Dery and Shanthal Vasanth and Elena Gribovskaya and Linhai Qiu and Mahdis Mahdieh and Wojtek Skut and Elizabeth Nielsen and CJ Zheng and Adams Yu and Carrie Grimes Bostock and Shaleen Gupta and Aaron Archer and Chris Rawles and Elinor Davies and Alexey Svyatkovskiy and Tomy Tsai and Yoni Halpern and Christian Reisswig and Bartek Wydrowski and Bo Chang and Joan Puigcerver and Mor Hazan Taege and Jian Li and Eva Schnider and Xinjian Li and Dragos Dena and Yunhan Xu and Umesh Telang and Tianze Shi and Heiga Zen and Kyle Kastner and Yeongil Ko and Neesha Subramaniam and Aviral Kumar and Pete Blois and Zhuyun Dai and John Wieting and Yifeng Lu and Yoel Zeldes and Tian Xie and Anja Hauth and Alexandru Ţifrea and Yuqi Li and Sam El-Husseini and Dan Abolafia and Howard Zhou and Wen Ding and Sahra Ghalebikesabi and Carlos Guía and Andrii Maksai and Ágoston Weisz and Sercan Arik and Nick Sukhanov and Aga Świetlik and Xuhui Jia and Luo Yu and Weiyue Wang and Mark Brand and Dawn Bloxwich and Sean Kirmani and Zhe Chen and Alec Go and Pablo Sprechmann and Nithish Kannen and Alen Carin and Paramjit Sandhu and Isabel Edkins and Leslie Nooteboom and Jai Gupta and Loren Maggiore and Javad Azizi and Yael Pritch and Pengcheng Yin and Mansi Gupta and Danny Tarlow and Duncan Smith and Desi Ivanov and Mohammad Babaeizadeh and Ankita Goel and Satish Kambala and Grace Chu and Matej Kastelic and Michelle Liu and Hagen Soltau and Austin Stone and Shivani Agrawal and Min Kim and Kedar Soparkar and Srinivas Tadepalli and Oskar Bunyan and Rachel Soh and Arvind Kannan and DY Kim and Blake JianHang Chen and Afief Halumi and Sudeshna Roy and Yulong Wang and Olcan Sercinoglu and Gena Gibson and Sijal Bhatnagar and Motoki Sano and Daniel von Dincklage and Qingchun Ren and Blagoj Mitrevski and Mirek Olšák and Jennifer She and Carl Doersch and Jilei and Wang and Bingyuan Liu and Qijun Tan and Tamar Yakar and Tris Warkentin and Alex Ramirez and Carl Lebsack and Josh Dillon and Rajiv Mathews and Tom Cobley and Zelin Wu and Zhuoyuan Chen and Jon Simon and Swaroop Nath and Tara Sainath and Alexei Bendebury and Ryan Julian and Bharath Mankalale and Daria Ćurko and Paulo Zacchello and Adam R. Brown and Kiranbir Sodhia and Heidi Howard and Sergi Caelles and Abhinav Gupta and Gareth Evans and Anna Bulanova and Lesley Katzen and Roman Goldenberg and Anton Tsitsulin and Joe Stanton and Benoit Schillings and Vitaly Kovalev and Corey Fry and Rushin Shah and Kuo Lin and Shyam Upadhyay and Cheng Li and Soroush Radpour and Marcello Maggioni and Jing Xiong and Lukas Haas and Jenny Brennan and Aishwarya Kamath and Nikolay Savinov and Arsha Nagrani and Trevor Yacovone and Ryan Kappedal and Kostas Andriopoulos and Li Lao and YaGuang Li and Grigory Rozhdestvenskiy and Kazuma Hashimoto and Andrew Audibert and Sophia Austin and Daniel Rodriguez and Anian Ruoss and Garrett Honke and Deep Karkhanis and Xi Xiong and Qing Wei and James Huang and Zhaoqi Leng and Vittal Premachandran and Stan Bileschi and Georgios Evangelopoulos and Thomas Mensink and Jay Pavagadhi and Denis Teplyashin and Paul Chang and Linting Xue and Garrett Tanzer and Sally Goldman and Kaushal Patel and Shixin Li and Jeremy Wiesner and Ivy Zheng and Ian Stewart-Binks and Jie Han and Zhi Li and Liangchen Luo and Karel Lenc and Mario Lučić and Fuzhao Xue and Ryan Mullins and Alexey Guseynov and Chung-Ching Chang and Isaac Galatzer-Levy and Adam Zhang and Garrett Bingham and Grace Hu and Ale Hartman and Yue Ma and Jordan Griffith and Alex Irpan and Carey Radebaugh and Summer Yue and Lijie Fan and Victor Ungureanu and Christina Sorokin and Hannah Teufel and Peiran Li and Rohan Anil and Dimitris Paparas and Todd Wang and Chu-Cheng Lin and Hui Peng and Megan Shum and Goran Petrovic and Demetra Brady and Richard Nguyen and Klaus Macherey and Zhihao Li and Harman Singh and Madhavi Yenugula and Mariko Iinuma and Xinyi Chen and Kavya Kopparapu and Alexey Stern and Shachi Dave and Chandu Thekkath and Florence Perot and Anurag Kumar and Fangda Li and Yang Xiao and Matthew Bilotti and Mohammad Hossein Bateni and Isaac Noble and Lisa Lee and Amelio Vázquez-Reina and Julian Salazar and Xiaomeng Yang and Boyu Wang and Ela Gruzewska and Anand Rao and Sindhu Raghuram and Zheng Xu and Eyal Ben-David and Jieru Mei and Sid Dalmia and Zhaoyi Zhang and Yuchen Liu and Gagan Bansal and Helena Pankov and Steven Schwarcz and Andrea Burns and Christine Chan and Sumit Sanghai and Ricky Liang and Ethan Liang and Antoine He and Amy Stuart and Arun Narayanan and Yukun Zhu and Christian Frank and Bahar Fatemi and Amit Sabne and Oran Lang and Indro Bhattacharya and Shane Settle and Maria Wang and Brendan McMahan and Andrea Tacchetti and Livio Baldini Soares and Majid Hadian and Serkan Cabi and Timothy Chung and Nikita Putikhin and Gang Li and Jeremy Chen and Austin Tarango and Henryk Michalewski and Mehran Kazemi and Hussain Masoom and Hila Sheftel and Rakesh Shivanna and Archita Vadali and Ramona Comanescu and Doug Reid and Joss Moore and Arvind Neelakantan and Michaël Sander and Jonathan Herzig and Aviv Rosenberg and Mostafa Dehghani and JD Choi and Michael Fink and Reid Hayes and Eric Ge and Shitao Weng and Chia-Hua Ho and John Karro and Kalpesh Krishna and Lam Nguyen Thiet and Amy Skerry-Ryan and Daniel Eppens and Marco Andreetto and Navin Sarma and Silvano Bonacina and Burcu Karagol Ayan and Megha Nawhal and Zhihao Shan and Mike Dusenberry and Shantanu Thakoor and Sagar Gubbi and Duc Dung Nguyen and Reut Tsarfaty and Samuel Albanie and Jovana Mitrović and Meet Gandhi and Bo-Juen Chen and Alessandro Epasto and Georgi Stephanov and Ye Jin and Samuel Gehman and Aida Amini and Jack Weber and Feryal Behbahani and Shawn Xu and Miltos Allamanis and Xi Chen and Myle Ott and Claire Sha and Michal Jastrzebski and Hang Qi and David Greene and Xinyi Wu and Abodunrinwa Toki and Daniel Vlasic and Jane Shapiro and Ragha Kotikalapudi and Zhe Shen and Takaaki Saeki and Sirui Xie and Albin Cassirer and Shikhar Bharadwaj and Tatsuya Kiyono and Srinadh Bhojanapalli and Elan Rosenfeld and Sam Ritter and Jieming Mao and João Gabriel Oliveira and Zoltan Egyed and Bernd Bandemer and Emilio Parisotto and Keisuke Kinoshita and Juliette Pluto and Petros Maniatis and Steve Li and Yaohui Guo and Golnaz Ghiasi and Jean Tarbouriech and Srimon Chatterjee and Julie Jin and Katrina and Xu and Jennimaria Palomaki and Séb Arnold and Madhavi Sewak and Federico Piccinini and Mohit Sharma and Ben Albrecht and Sean Purser-haskell and Ashwin Vaswani and Chongyan Chen and Matheus Wisniewski and Qin Cao and John Aslanides and Nguyet Minh Phu and Maximilian Sieb and Lauren Agubuzu and Anne Zheng and Daniel Sohn and Marco Selvi and Anders Andreassen and Krishan Subudhi and Prem Eruvbetine and Oliver Woodman and Tomas Mery and Sebastian Krause and Xiaoqi Ren and Xiao Ma and Jincheng Luo and Dawn Chen and Wei Fan and Henry Griffiths and Christian Schuler and Alice Li and Shujian Zhang and Jean-Michel Sarr and Shixin Luo and Riccardo Patana and Matthew Watson and Dani Naboulsi and Michael Collins and Sailesh Sidhwani and Emiel Hoogeboom and Sharon Silver and Emily Caveness and Xiaokai Zhao and Mikel Rodriguez and Maxine Deines and Libin Bai and Patrick Griffin and Marco Tagliasacchi and Emily Xue and Spandana Raj Babbula and Bo Pang and Nan Ding and Gloria Shen and Elijah Peake and Remi Crocker and Shubha Srinivas Raghvendra and Danny Swisher and Woohyun Han and Richa Singh and Ling Wu and Vladimir Pchelin and Tsendsuren Munkhdalai and Dana Alon and Geoff Bacon and Efren Robles and Jannis Bulian and Melvin Johnson and George Powell and Felipe Tiengo Ferreira and Yaoyiran Li and Frederik Benzing and Mihajlo Velimirović and Hubert Soyer and William Kong and Tony and Nguyên and Zhen Yang and Jeremiah Liu and Joost van Amersfoort and Daniel Gillick and Baochen Sun and Nathalie Rauschmayr and Katie Zhang and Serena Zhan and Tao Zhou and Alexey Frolov and Chengrun Yang and Denis Vnukov and Louis Rouillard and Hongji Li and Amol Mandhane and Nova Fallen and Rajesh Venkataraman and Clara Huiyi Hu and Jennifer Brennan and Jenny Lee and Jerry Chang and Martin Sundermeyer and Zhufeng Pan and Rosemary Ke and Simon Tong and Alex Fabrikant and William Bono and Jindong Gu and Ryan Foley and Yiran Mao and Manolis Delakis and Dhruva Bhaswar and Roy Frostig and Nick Li and Avital Zipori and Cath Hope and Olga Kozlova and Swaroop Mishra and Josip Djolonga and Craig Schiff and Majd Al Merey and Eleftheria Briakou and Peter Morgan and Andy Wan and Avinatan Hassidim and RJ Skerry-Ryan and Kuntal Sengupta and Mary Jasarevic and Praveen Kallakuri and Paige Kunkle and Hannah Brennan and Tom Lieber and Hassan Mansoor and Julian Walker and Bing Zhang and Annie Xie and Goran Žužić and Adaeze Chukwuka and Alex Druinsky and Donghyun Cho and Rui Yao and Ferjad Naeem and Shiraz Butt and Eunyoung Kim and Zhipeng Jia and Mandy Jordan and Adam Lelkes and Mark Kurzeja and Sophie Wang and James Zhao and Andrew Over and Abhishek Chakladar and Marcel Prasetya and Neha Jha and Sriram Ganapathy and Yale Cong and Prakash Shroff and Carl Saroufim and Sobhan Miryoosefi and Mohamed Hammad and Tajwar Nasir and Weijuan Xi and Yang Gao and Young Maeng and Ben Hora and Chin-Yi Cheng and Parisa Haghani and Yoad Lewenberg and Caden Lu and Martin Matysiak and Naina Raisinghani and Huiyu Wang and Lexi Baugher and Rahul Sukthankar and Minh Giang and John Schultz and Noah Fiedel and Minmin Chen and Cheng-Chun Lee and Tapomay Dey and Hao Zheng and Shachi Paul and Celine Smith and Andy Ly and Yicheng Wang and Rishabh Bansal and Bartek Perz and Susanna Ricco and Stasha Blank and Vaishakh Keshava and Deepak Sharma and Marvin Chow and Kunal Lad and Komal Jalan and Simon Osindero and Craig Swanson and Jacob Scott and Anastasija Ilić and Xiaowei Li and Siddhartha Reddy Jonnalagadda and Afzal Shama Soudagar and Yan Xiong and Bat-Orgil Batsaikhan and Daniel Jarrett and Naveen Kumar and Maulik Shah and Matt Lawlor and Austin Waters and Mark Graham and Rhys May and Sabela Ramos and Sandra Lefdal and Zeynep Cankara and Nacho Cano and Brendan O'Donoghue and Jed Borovik and Frederick Liu and Jordan Grimstad and Mahmoud Alnahlawi and Katerina Tsihlas and Tom Hudson and Nikolai Grigorev and Yiling Jia and Terry Huang and Tobenna Peter Igwe and Sergei Lebedev and Xiaodan Tang and Igor Krivokon and Frankie Garcia and Melissa Tan and Eric Jia and Peter Stys and Shikhar Vashishth and Yu Liang and Balaji Venkatraman and Chenjie Gu and Anastasios Kementsietsidis and Chen Zhu and Junehyuk Jung and Yunfei Bai and Mohammad Javad Hosseini and Faruk Ahmed and Aditya Gupta and Xin Yuan and Shereen Ashraf and Shitij Nigam and Gautam Vasudevan and Pranjal Awasthi and Adi Mayrav Gilady and Zelda Mariet and Ramy Eskander and Haiguang Li and Hexiang Hu and Guillermo Garrido and Philippe Schlattner and George Zhang and Rohun Saxena and Petar Dević and Kritika Muralidharan and Ashwin Murthy and Yiqian Zhou and Min Choi and Arissa Wongpanich and Zhengdong Wang and Premal Shah and Yuntao Xu and Yiling Huang and Stephen Spencer and Alice Chen and James Cohan and Junjie Wang and Jonathan Tompson and Junru Wu and Ruba Haroun and Haiqiong Li and Blanca Huergo and Fan Yang and Tongxin Yin and James Wendt and Michael Bendersky and Rahma Chaabouni and Javier Snaider and Johan Ferret and Abhishek Jindal and Tara Thompson and Andrew Xue and Will Bishop and Shubham Milind Phal and Archit Sharma and Yunhsuan Sung and Prabakar Radhakrishnan and Mo Shomrat and Reeve Ingle and Roopali Vij and Justin Gilmer and Mihai Dorin Istin and Sam Sobell and Yang Lu and Emily Nottage and Dorsa Sadigh and Jeremiah Willcock and Tingnan Zhang and Steve Xu and Sasha Brown and Katherine Lee and Gary Wang and Yun Zhu and Yi Tay and Cheolmin Kim and Audrey Gutierrez and Abhanshu Sharma and Yongqin Xian and Sungyong Seo and Claire Cui and Elena Pochernina and Cip Baetu and Krzysztof Jastrzębski and Mimi Ly and Mohamed Elhawaty and Dan Suh and Eren Sezener and Pidong Wang and Nancy Yuen and George Tucker and Jiahao Cai and Zuguang Yang and Cindy Wang and Alex Muzio and Hai Qian and Jae Yoo and Derek Lockhart and Kevin R. McKee and Mandy Guo and Malika Mehrotra and Artur Mendonça and Sanket Vaibhav Mehta and Sherry Ben and Chetan Tekur and Jiaqi Mu and Muye Zhu and Victoria Krakovna and Hongrae Lee and AJ Maschinot and Sébastien Cevey and HyunJeong Choe and Aijun Bai and Hansa Srinivasan and Derek Gasaway and Nick Young and Patrick Siegler and Dan Holtmann-Rice and Vihari Piratla and Kate Baumli and Roey Yogev and Alex Hofer and Hado van Hasselt and Svetlana Grant and Yuri Chervonyi and David Silver and Andrew Hogue and Ayushi Agarwal and Kathie Wang and Preeti Singh and Four Flynn and Josh Lipschultz and Robert David and Lizzetth Bellot and Yao-Yuan Yang and Long Le and Filippo Graziano and Kate Olszewska and Kevin Hui and Akanksha Maurya and Nikos Parotsidis and Weijie Chen and Tayo Oguntebi and Joe Kelley and Anirudh Baddepudi and Johannes Mauerer and Gregory Shaw and Alex Siegman and Lin Yang and Shravya Shetty and Subhrajit Roy and Yunting Song and Wojciech Stokowiec and Ryan Burnell and Omkar Savant and Robert Busa-Fekete and Jin Miao and Samrat Ghosh and Liam MacDermed and Phillip Lippe and Mikhail Dektiarev and Zach Behrman and Fabian Mentzer and Kelvin Nguyen and Meng Wei and Siddharth Verma and Chris Knutsen and Sudeep Dasari and Zhipeng Yan and Petr Mitrichev and Xingyu Wang and Virat Shejwalkar and Jacob Austin and Srinivas Sunkara and Navneet Potti and Yan Virin and Christian Wright and Gaël Liu and Oriana Riva and Etienne Pot and Greg Kochanski and Quoc Le and Gargi Balasubramaniam and Arka Dhar and Yuguo Liao and Adam Bloniarz and Divyansh Shukla and Elizabeth Cole and Jong Lee and Sheng Zhang and Sushant Kafle and Siddharth Vashishtha and Parsa Mahmoudieh and Grace Chen and Raphael Hoffmann and Pranesh Srinivasan and Agustin Dal Lago and Yoav Ben Shalom and Zi Wang and Michael Elabd and Anuj Sharma and Junhyuk Oh and Suraj Kothawade and Maigo Le and Marianne Monteiro and Shentao Yang and Kaiz Alarakyia and Robert Geirhos and Diana Mincu and Håvard Garnes and Hayato Kobayashi and Soroosh Mariooryad and Kacper Krasowiak and Zhixin and Lai and Shibl Mourad and Mingqiu Wang and Fan Bu and Ophir Aharoni and Guanjie Chen and Abhimanyu Goyal and Vadim Zubov and Ankur Bapna and Elahe Dabir and Nisarg Kothari and Kay Lamerigts and Nicola De Cao and Jeremy Shar and Christopher Yew and Nitish Kulkarni and Dre Mahaarachchi and Mandar Joshi and Zhenhai Zhu and Jared Lichtarge and Yichao Zhou and Hannah Muckenhirn and Vittorio Selo and Oriol Vinyals and Peter Chen and Anthony Brohan and Vaibhav Mehta and Sarah Cogan and Ruth Wang and Ty Geri and Wei-Jen Ko and Wei Chen and Fabio Viola and Keshav Shivam and Lisa Wang and Madeleine Clare Elish and Raluca Ada Popa and Sébastien Pereira and Jianqiao Liu and Raphael Koster and Donnie Kim and Gufeng Zhang and Sayna Ebrahimi and Partha Talukdar and Yanyan Zheng and Petra Poklukar and Ales Mikhalap and Dale Johnson and Anitha Vijayakumar and Mark Omernick and Matt Dibb and Ayush Dubey and Qiong Hu and Apurv Suman and Vaibhav Aggarwal and Ilya Kornakov and Fei Xia and Wing Lowe and Alexey Kolganov and Ted Xiao and Vitaly Nikolaev and Steven Hemingray and Bonnie Li and Joana Iljazi and Mikołaj Rybiński and Ballie Sandhu and Peggy Lu and Thang Luong and Rodolphe Jenatton and Vineetha Govindaraj and Hui and Li and Gabriel Dulac-Arnold and Wonpyo Park and Henry Wang and Abhinit Modi and Jean Pouget-Abadie and Kristina Greller and Rahul Gupta and Robert Berry and Prajit Ramachandran and Jinyu Xie and Liam McCafferty and Jianling Wang and Kilol Gupta and Hyeontaek Lim and Blaž Bratanič and Andy Brock and Ilia Akolzin and Jim Sproch and Dan Karliner and Duhyeon Kim and Adrian Goedeckemeyer and Noam Shazeer and Cordelia Schmid and Daniele Calandriello and Parul Bhatia and Krzysztof Choromanski and Ceslee Montgomery and Dheeru Dua and Ana Ramalho and Helen King and Yue Gao and Lynn Nguyen and David Lindner and Divya Pitta and Oleaser Johnson and Khalid Salama and Diego Ardila and Michael Han and Erin Farnese and Seth Odoom and Ziyue Wang and Xiangzhuo Ding and Norman Rink and Ray Smith and Harshal Tushar Lehri and Eden Cohen and Neera Vats and Tong He and Parthasarathy Gopavarapu and Adam Paszke and Miteyan Patel and Wouter Van Gansbeke and Lucia Loher and Luis Castro and Maria Voitovich and Tamara von Glehn and Nelson George and Simon Niklaus and Zach Eaton-Rosen and Nemanja Rakićević and Erik Jue and Sagi Perel and Carrie Zhang and Yuval Bahat and Angéline Pouget and Zhi Xing and Fantine Huot and Ashish Shenoy and Taylor Bos and Vincent Coriou and Bryan Richter and Natasha Noy and Yaqing Wang and Santiago Ontanon and Siyang Qin and Gleb Makarchuk and Demis Hassabis and Zhuowan Li and Mandar Sharma and Kumaran Venkatesan and Iurii Kemaev and Roxanne Daniel and Shiyu Huang and Saloni Shah and Octavio Ponce and Warren and Chen and Manaal Faruqui and Jialin Wu and Slavica Andačić and Szabolcs Payrits and Daniel McDuff and Tom Hume and Yuan Cao and MH Tessler and Qingze Wang and Yinan Wang and Ivor Rendulic and Eirikur Agustsson and Matthew Johnson and Tanya Lando and Andrew Howard and Sri Gayatri Sundara Padmanabhan and Mayank Daswani and Andrea Banino and Michael Kilgore and Jonathan Heek and Ziwei Ji and Alvaro Caceres and Conglong Li and Nora Kassner and Alexey Vlaskin and Zeyu Liu and Alex Grills and Yanhan Hou and Roykrong Sukkerd and Gowoon Cheon and Nishita Shetty and Larisa Markeeva and Piotr Stanczyk and Tejas Iyer and Yuan Gong and Shawn Gao and Keerthana Gopalakrishnan and Tim Blyth and Malcolm Reynolds and Avishkar Bhoopchand and Misha Bilenko and Dero Gharibian and Vicky Zayats and Aleksandra Faust and Abhinav Singh and Min Ma and Hongyang Jiao and Sudheendra Vijayanarasimhan and Lora Aroyo and Vikas Yadav and Sarah Chakera and Ashwin Kakarla and Vilobh Meshram and Karol Gregor and Gabriela Botea and Evan Senter and Dawei Jia and Geza Kovacs and Neha Sharma and Sebastien Baur and Kai Kang and Yifan He and Lin Zhuo and Marija Kostelac and Itay Laish and Songyou Peng and Louis O'Bryan and Daniel Kasenberg and Girish Ramchandra Rao and Edouard Leurent and Biao Zhang and Sage Stevens and Ana Salazar and Ye Zhang and Ivan Lobov and Jake Walker and Allen Porter and Morgan Redshaw and Han Ke and Abhishek Rao and Alex Lee and Hoi Lam and Michael Moffitt and Jaeyoun Kim and Siyuan Qiao and Terry Koo and Robert Dadashi and Xinying Song and Mukund Sundararajan and Peng Xu and Chizu Kawamoto and Yan Zhong and Clara Barbu and Apoorv Reddy and Mauro Verzetti and Leon Li and George Papamakarios and Hanna Klimczak-Plucińska and Mary Cassin and Koray Kavukcuoglu and Rigel Swavely and Alain Vaucher and Jeffrey Zhao and Ross Hemsley and Michael Tschannen and Heming Ge and Gaurav Menghani and Yang Yu and Natalie Ha and Wei He and Xiao Wu and Maggie Song and Rachel Sterneck and Stefan Zinke and Dan A. Calian and Annie Marsden and Alejandro Cruzado Ruiz and Matteo Hessel and Almog Gueta and Benjamin Lee and Brian Farris and Manish Gupta and Yunjie Li and Mohammad Saleh and Vedant Misra and Kefan Xiao and Piermaria Mendolicchio and Gavin Buttimore and Varvara Krayvanova and Nigamaa Nayakanti and Matthew Wiethoff and Yash Pande and Azalia Mirhoseini and Ni Lao and Jasmine Liu and Yiqing Hua and Angie Chen and Yury Malkov and Dmitry Kalashnikov and Shubham Gupta and Kartik Audhkhasi and Yuexiang Zhai and Sudhindra Kopalle and Prateek Jain and Eran Ofek and Clemens Meyer and Khuslen Baatarsukh and Hana Strejček and Jun Qian and James Freedman and Ricardo Figueira and Michal Sokolik and Olivier Bachem and Raymond Lin and Dia Kharrat and Chris Hidey and Pingmei Xu and Dennis Duan and Yin Li and Muge Ersoy and Richard Everett and Kevin Cen and Rebeca Santamaria-Fernandez and Amir Taubenfeld and Ian Mackinnon and Linda Deng and Polina Zablotskaia and Shashank Viswanadha and Shivanker Goel and Damion Yates and Yunxiao Deng and Peter Choy and Mingqing Chen and Abhishek Sinha and Alex Mossin and Yiming Wang and Arthur Szlam and Susan Hao and Paul Kishan Rubenstein and Metin Toksoz-Exley and Miranda Aperghis and Yin Zhong and Junwhan Ahn and Michael Isard and Olivier Lacombe and Florian Luisier and Chrysovalantis Anastasiou and Yogesh Kalley and Utsav Prabhu and Emma Dunleavy and Shaan Bijwadia and Justin Mao-Jones and Kelly Chen and Rama Pasumarthi and Emily Wood and Adil Dostmohamed and Nate Hurley and Jiri Simsa and Alicia Parrish and Mantas Pajarskas and Matt Harvey and Ondrej Skopek and Yony Kochinski and Javier Rey and Verena Rieser and Denny Zhou and Sun Jae Lee and Trilok Acharya and Guowang Li and Joe Jiang and Xiaofan Zhang and Bryant Gipson and Ethan Mahintorabi and Marco Gelmi and Nima Khajehnouri and Angel Yeh and Kayi Lee and Loic Matthey and Leslie Baker and Trang Pham and Han Fu and Alex Pak and Prakhar Gupta and Cristina Vasconcelos and Adam Sadovsky and Brian Walker and Sissie Hsiao and Patrik Zochbauer and Andreea Marzoca and Noam Velan and Junhao Zeng and Gilles Baechler and Danny Driess and Divya Jain and Yanping Huang and Lizzie Tao and John Maggs and Nir Levine and Jon Schneider and Erika Gemzer and Samuel Petit and Shan Han and Zach Fisher and Dustin Zelle and Courtney Biles and Eugene Ie and Asya Fadeeva and Casper Liu and Juliana Vicente Franco and Adrian Collister and Hao Zhang and Renshen Wang and Ruizhe Zhao and Leandro Kieliger and Kurt Shuster and Rui Zhu and Boqing Gong and Lawrence Chan and Ruoxi Sun and Sujoy Basu and Roland Zimmermann and Jamie Hayes and Abhishek Bapna and Jasper Snoek and Weel Yang and Puranjay Datta and Jad Al Abdallah and Kevin Kilgour and Lu Li and SQ Mah and Yennie Jun and Morgane Rivière and Abhijit Karmarkar and Tammo Spalink and Tao Huang and Lucas Gonzalez and Duc-Hieu Tran and Averi Nowak and John Palowitch and Martin Chadwick and Ellie Talius and Harsh Mehta and Thibault Sellam and Philipp Fränken and Massimo Nicosia and Kyle He and Aditya Kini and David Amos and Sugato Basu and Harrison Jobe and Eleni Shaw and Qiantong Xu and Colin Evans and Daisuke Ikeda and Chaochao Yan and Larry Jin and Lun Wang and Sachin Yadav and Ilia Labzovsky and Ramesh Sampath and Ada Ma and Candice Schumann and Aditya Siddhant and Rohin Shah and John Youssef and Rishabh Agarwal and Natalie Dabney and Alessio Tonioni and Moran Ambar and Jing Li and Isabelle Guyon and Benny Li and David Soergel and Boya Fang and Georgi Karadzhov and Cristian Udrescu and Trieu Trinh and Vikas Raunak and Seb Noury and Dee Guo and Sonal Gupta and Mara Finkelstein and Denis Petek and Lihao Liang and Greg Billock and Pei Sun and David Wood and Yiwen Song and Xiaobin Yu and Tatiana Matejovicova and Regev Cohen and Kalyan Andra and David D'Ambrosio and Zhiwei Deng and Vincent Nallatamby and Ebrahim Songhori and Rumen Dangovski and Andrew Lampinen and Pankil Botadra and Adam Hillier and Jiawei Cao and Nagabhushan Baddi and Adhi Kuncoro and Toshihiro Yoshino and Ankit Bhagatwala and Marcáurelio Ranzato and Rylan Schaeffer and Tianlin Liu and Shuai Ye and Obaid Sarvana and John Nham and Chenkai Kuang and Isabel Gao and Jinoo Baek and Shubham Mittal and Ayzaan Wahid and Anita Gergely and Bin Ni and Josh Feldman and Carrie Muir and Pascal Lamblin and Wolfgang Macherey and Ethan Dyer and Logan Kilpatrick and Víctor Campos and Mukul Bhutani and Stanislav Fort and Yanif Ahmad and Aliaksei Severyn and Kleopatra Chatziprimou and Oleksandr Ferludin and Mason Dimarco and Aditya Kusupati and Joe Heyward and Dan Bahir and Kevin Villela and Katie Millican and Dror Marcus and Sanaz Bahargam and Caglar Unlu and Nicholas Roth and Zichuan Wei and Siddharth Gopal and Deepanway Ghoshal and Edward Lee and Sharon Lin and Jennie Lees and Dayeong Lee and Anahita Hosseini and Connie Fan and Seth Neel and Marcus Wu and Yasemin Altun and Honglong Cai and Enrique Piqueras and Josh Woodward and Alessandro Bissacco and Salem Haykal and Mahyar Bordbar and Prasha Sundaram and Sarah Hodkinson and Daniel Toyama and George Polovets and Austin Myers and Anu Sinha and Tomer Levinboim and Kashyap Krishnakumar and Rachita Chhaparia and Tatiana Sholokhova and Nitesh Bharadwaj Gundavarapu and Ganesh Jawahar and Haroon Qureshi and Jieru Hu and Nikola Momchev and Matthew Rahtz and Renjie Wu and Aishwarya P S and Kedar Dhamdhere and Meiqi Guo and Umang Gupta and Ali Eslami and Mariano Schain and Michiel Blokzijl and David Welling and Dave Orr and Levent Bolelli and Nicolas Perez-Nieves and Mikhail Sirotenko and Aman Prasad and Arjun Kar and Borja De Balle Pigem and Tayfun Terzi and Gellért Weisz and Dipankar Ghosh and Aditi Mavalankar and Dhruv Madeka and Kaspar Daugaard and Hartwig Adam and Viraj Shah and Dana Berman and Maggie Tran and Steven Baker and Ewa Andrejczuk and Grishma Chole and Ganna Raboshchuk and Mahdi Mirzazadeh and Thais Kagohara and Shimu Wu and Christian Schallhart and Bernett Orlando and Chen Wang and Alban Rrustemi and Hao Xiong and Hao Liu and Arpi Vezer and Nolan Ramsden and Shuo-yiin Chang and Sidharth Mudgal and Yan Li and Nino Vieillard and Yedid Hoshen and Farooq Ahmad and Ambrose Slone and Amy Hua and Natan Potikha and Mirko Rossini and Jon Stritar and Sushant Prakash and Zifeng Wang and Xuanyi Dong and Alireza Nazari and Efrat Nehoran and Kaan Tekelioglu and Yinxiao Li and Kartikeya Badola and Tom Funkhouser and Yuanzhen Li and Varun Yerram and Ramya Ganeshan and Daniel Formoso and Karol Langner and Tian Shi and Huijian Li and Yumeya Yamamori and Amayika Panda and Alaa Saade and Angelo Scorza Scarpati and Chris Breaux and CJ Carey and Zongwei Zhou and Cho-Jui Hsieh and Sophie Bridgers and Alena Butryna and Nishesh Gupta and Vaibhav Tulsyan and Sanghyun Woo and Evgenii Eltyshev and Will Grathwohl and Chanel Parks and Seth Benjamin and Rina Panigrahy and Shenil Dodhia and Daniel De Freitas and Chris Sauer and Will Song and Ferran Alet and Jackson Tolins and Cosmin Paduraru and Xingyi Zhou and Brian Albert and Zizhao Zhang and Lei Shu and Mudit Bansal and Sarah Nguyen and Amir Globerson and Owen Xiao and James Manyika and Tom Hennigan and Rong Rong and Josip Matak and Anton Bakalov and Ankur Sharma and Danila Sinopalnikov and Andrew Pierson and Stephen Roller and Geoff Brown and Mingcen Gao and Toshiyuki Fukuzawa and Amin Ghafouri and Kenny Vassigh and Iain Barr and Zhicheng Wang and Anna Korsun and Rajesh Jayaram and Lijie Ren and Tim Zaman and Samira Khan and Yana Lunts and Dan Deutsch and Dave Uthus and Nitzan Katz and Masha Samsikova and Amr Khalifa and Nikhil Sethi and Jiao Sun and Luming Tang and Uri Alon and Xianghong Luo and Dian Yu and Abhishek Nayyar and Bryce Petrini and Will Truong and Vincent Hellendoorn and Nikolai Chinaev and Chris Alberti and Wei Wang and Jingcao Hu and Vahab Mirrokni and Ananth Balashankar and Avia Aharon and Aahil Mehta and Ahmet Iscen and Joseph Kready and Lucas Manning and Anhad Mohananey and Yuankai Chen and Anshuman Tripathi and Allen Wu and Igor Petrovski and Dawsen Hwang and Martin Baeuml and Shreyas Chandrakaladharan and Yuan Liu and Rey Coaguila and Maxwell Chen and Sally Ma and Pouya Tafti and Susheel Tatineni and Terry Spitz and Jiayu Ye and Paul Vicol and Mihaela Rosca and Adrià Puigdomènech and Zohar Yahav and Sanjay Ghemawat and Hanzhao Lin and Phoebe Kirk and Zaid Nabulsi and Sergey Brin and Bernd Bohnet and Ken Caluwaerts and Aditya Srikanth Veerubhotla and Dan Zheng and Zihang Dai and Petre Petrov and Yichong Xu and Ramin Mehran and Zhuo Xu and Luisa Zintgraf and Jiho Choi and Spurthi Amba Hombaiah and Romal Thoppilan and Sashank Reddi and Lukasz Lew and Li Li and Kellie Webster and KP Sawhney and Lampros Lamprou and Siamak Shakeri and Mayank Lunayach and Jianmin Chen and Sumit Bagri and Alex Salcianu and Ying Chen and Yani Donchev and Charlotte Magister and Signe Nørly and Vitor Rodrigues and Tomas Izo and Hila Noga and Joe Zou and Thomas Köppe and Wenxuan Zhou and Kenton Lee and Xiangzhu Long and Danielle Eisenbud and Anthony Chen and Connor Schenck and Chi Ming To and Peilin Zhong and Emanuel Taropa and Minh Truong and Omer Levy and Danilo Martins and Zhiyuan Zhang and Christopher Semturs and Kelvin Zhang and Alex Yakubovich and Pol Moreno and Lara McConnaughey and Di Lu and Sam Redmond and Lotte Weerts and Yonatan Bitton and Tiziana Refice and Nicolas Lacasse and Arthur Conmy and Corentin Tallec and Julian Odell and Hannah Forbes-Pollard and Arkadiusz Socala and Jonathan Hoech and Pushmeet Kohli and Alanna Walton and Rui Wang and Mikita Sazanovich and Kexin Zhu and Andrei Kapishnikov and Rich Galt and Matthew Denton and Ben Murdoch and Caitlin Sikora and Kareem Mohamed and Wei Wei and Uri First and Tim McConnell and Luis C. Cobo and James Qin and Thi Avrahami and Daniel Balle and Yu Watanabe and Annie Louis and Adam Kraft and Setareh Ariafar and Yiming Gu and Eugénie Rives and Charles Yoon and Andrei Rusu and James Cobon-Kerr and Chris Hahn and Jiaming Luo and Yuvein and Zhu and Niharika Ahuja and Rodrigo Benenson and Raphaël Lopez Kaufman and Honglin Yu and Lloyd Hightower and Junlin Zhang and Darren Ni and Lisa Anne Hendricks and Gabby Wang and Gal Yona and Lalit Jain and Pablo Barrio and Surya Bhupatiraju and Siva Velusamy and Allan Dafoe and Sebastian Riedel and Tara Thomas and Zhe Yuan and Mathias Bellaiche and Sheena Panthaplackel and Klemen Kloboves and Sarthak Jauhari and Canfer Akbulut and Todor Davchev and Evgeny Gladchenko and David Madras and Aleksandr Chuklin and Tyrone Hill and Quan Yuan and Mukundan Madhavan and Luke Leonhard and Dylan Scandinaro and Qihang Chen and Ning Niu and Arthur Douillard and Bogdan Damoc and Yasumasa Onoe and Fabian Pedregosa and Fred Bertsch and Chas Leichner and Joseph Pagadora and Jonathan Malmaud and Sameera Ponda and Andy Twigg and Oleksii Duzhyi and Jingwei Shen and Miaosen Wang and Roopal Garg and Jing Chen and Utku Evci and Jonathan Lee and Leon Liu and Koji Kojima and Masa Yamaguchi and Arunkumar Rajendran and AJ Piergiovanni and Vinodh Kumar Rajendran and Marco Fornoni and Gabriel Ibagon and Harry Ragan and Sadh MNM Khan and John Blitzer and Andrew Bunner and Guan Sun and Takahiro Kosakai and Scott Lundberg and Ndidi Elue and Kelvin Guu and SK Park and Jane Park and Arunachalam Narayanaswamy and Chengda Wu and Jayaram Mudigonda and Trevor Cohn and Hairong Mu and Ravi Kumar and Laura Graesser and Yichi Zhang and Richard Killam and Vincent Zhuang and Mai Giménez and Wael Al Jishi and Ruy Ley-Wild and Alex Zhai and Kazuki Osawa and Diego Cedillo and Jialu Liu and Mayank Upadhyay and Marcin Sieniek and Roshan Sharma and Tom Paine and Anelia Angelova and Sravanti Addepalli and Carolina Parada and Kingshuk Majumder and Avery Lamp and Sanjiv Kumar and Xiang Deng and Artiom Myaskovsky and Tea Sabolić and Jeffrey Dudek and Sarah York and Félix de Chaumont Quitry and Jiazhong Nie and Dee Cattle and Alok Gunjan and Bilal Piot and Waleed Khawaja and Seojin Bang and Simon Wang and Siavash Khodadadeh and Raghavender R and Praynaa Rawlani and Richard Powell and Kevin Lee and Johannes Griesser and GS Oh and Cesar Magalhaes and Yujia Li and Simon Tokumine and Hadas Natalie Vogel and Dennis Hsu and Arturo BC and Disha Jindal and Matan Cohen and Zi Yang and Junwei Yuan and Dario de Cesare and Tony Bruguier and Jun Xu and Monica Roy and Alon Jacovi and Dan Belov and Rahul Arya and Phoenix Meadowlark and Shlomi Cohen-Ganor and Wenting Ye and Patrick Morris-Suzuki and Praseem Banzal and Gan Song and Pranavaraj Ponnuramu and Fred Zhang and George Scrivener and Salah Zaiem and Alif Raditya Rochman and Kehang Han and Badih Ghazi and Kate Lee and Shahar Drath and Daniel Suo and Antonious Girgis and Pradeep Shenoy and Duy Nguyen and Douglas Eck and Somit Gupta and Le Yan and Joao Carreira and Anmol Gulati and Ruoxin Sang and Daniil Mirylenka and Emma Cooney and Edward Chou and Mingyang Ling and Cindy Fan and Ben Coleman and Guilherme Tubone and Ravin Kumar and Jason Baldridge and Felix Hernandez-Campos and Angeliki Lazaridou and James Besley and Itay Yona and Neslihan Bulut and Quentin Wellens and AJ Pierigiovanni and Jasmine George and Richard Green and Pu Han and Connie Tao and Geoff Clark and Chong You and Abbas Abdolmaleki and Justin Fu and Tongzhou Chen and Ashwin Chaugule and Angad Chandorkar and Altaf Rahman and Will Thompson and Penporn Koanantakool and Mike Bernico and Jie Ren and Andrey Vlasov and Sergei Vassilvitskii and Maciej Kula and Yizhong Liang and Dahun Kim and Yangsibo Huang and Chengxi Ye and Dmitry Lepikhin and Wesley Helmholz},
      year={2025},
      eprint={2507.06261},
      archivePrefix={arXiv},
      primaryClass={cs.CL},
      url={https://arxiv.org/abs/2507.06261}, 
}

@misc{grattafiori2024llama3herdmodels,
      title={The Llama 3 Herd of Models}, 
      author={Aaron Grattafiori and Abhimanyu Dubey and Abhinav Jauhri and Abhinav Pandey and Abhishek Kadian and Ahmad Al-Dahle and Aiesha Letman and Akhil Mathur and Alan Schelten and Alex Vaughan and Amy Yang and Angela Fan and Anirudh Goyal and Anthony Hartshorn and Aobo Yang and Archi Mitra and Archie Sravankumar and Artem Korenev and Arthur Hinsvark and Arun Rao and Aston Zhang and Aurelien Rodriguez and Austen Gregerson and Ava Spataru and Baptiste Roziere and Bethany Biron and Binh Tang and Bobbie Chern and Charlotte Caucheteux and Chaya Nayak and Chloe Bi and Chris Marra and Chris McConnell and Christian Keller and Christophe Touret and Chunyang Wu and Corinne Wong and Cristian Canton Ferrer and Cyrus Nikolaidis and Damien Allonsius and Daniel Song and Danielle Pintz and Danny Livshits and Danny Wyatt and David Esiobu and Dhruv Choudhary and Dhruv Mahajan and Diego Garcia-Olano and Diego Perino and Dieuwke Hupkes and Egor Lakomkin and Ehab AlBadawy and Elina Lobanova and Emily Dinan and Eric Michael Smith and Filip Radenovic and Francisco Guzmán and Frank Zhang and Gabriel Synnaeve and Gabrielle Lee and Georgia Lewis Anderson and Govind Thattai and Graeme Nail and Gregoire Mialon and Guan Pang and Guillem Cucurell and Hailey Nguyen and Hannah Korevaar and Hu Xu and Hugo Touvron and Iliyan Zarov and Imanol Arrieta Ibarra and Isabel Kloumann and Ishan Misra and Ivan Evtimov and Jack Zhang and Jade Copet and Jaewon Lee and Jan Geffert and Jana Vranes and Jason Park and Jay Mahadeokar and Jeet Shah and Jelmer van der Linde and Jennifer Billock and Jenny Hong and Jenya Lee and Jeremy Fu and Jianfeng Chi and Jianyu Huang and Jiawen Liu and Jie Wang and Jiecao Yu and Joanna Bitton and Joe Spisak and Jongsoo Park and Joseph Rocca and Joshua Johnstun and Joshua Saxe and Junteng Jia and Kalyan Vasuden Alwala and Karthik Prasad and Kartikeya Upasani and Kate Plawiak and Ke Li and Kenneth Heafield and Kevin Stone and Khalid El-Arini and Krithika Iyer and Kshitiz Malik and Kuenley Chiu and Kunal Bhalla and Kushal Lakhotia and Lauren Rantala-Yeary and Laurens van der Maaten and Lawrence Chen and Liang Tan and Liz Jenkins and Louis Martin and Lovish Madaan and Lubo Malo and Lukas Blecher and Lukas Landzaat and Luke de Oliveira and Madeline Muzzi and Mahesh Pasupuleti and Mannat Singh and Manohar Paluri and Marcin Kardas and Maria Tsimpoukelli and Mathew Oldham and Mathieu Rita and Maya Pavlova and Melanie Kambadur and Mike Lewis and Min Si and Mitesh Kumar Singh and Mona Hassan and Naman Goyal and Narjes Torabi and Nikolay Bashlykov and Nikolay Bogoychev and Niladri Chatterji and Ning Zhang and Olivier Duchenne and Onur Çelebi and Patrick Alrassy and Pengchuan Zhang and Pengwei Li and Petar Vasic and Peter Weng and Prajjwal Bhargava and Pratik Dubal and Praveen Krishnan and Punit Singh Koura and Puxin Xu and Qing He and Qingxiao Dong and Ragavan Srinivasan and Raj Ganapathy and Ramon Calderer and Ricardo Silveira Cabral and Robert Stojnic and Roberta Raileanu and Rohan Maheswari and Rohit Girdhar and Rohit Patel and Romain Sauvestre and Ronnie Polidoro and Roshan Sumbaly and Ross Taylor and Ruan Silva and Rui Hou and Rui Wang and Saghar Hosseini and Sahana Chennabasappa and Sanjay Singh and Sean Bell and Seohyun Sonia Kim and Sergey Edunov and Shaoliang Nie and Sharan Narang and Sharath Raparthy and Sheng Shen and Shengye Wan and Shruti Bhosale and Shun Zhang and Simon Vandenhende and Soumya Batra and Spencer Whitman and Sten Sootla and Stephane Collot and Suchin Gururangan and Sydney Borodinsky and Tamar Herman and Tara Fowler and Tarek Sheasha and Thomas Georgiou and Thomas Scialom and Tobias Speckbacher and Todor Mihaylov and Tong Xiao and Ujjwal Karn and Vedanuj Goswami and Vibhor Gupta and Vignesh Ramanathan and Viktor Kerkez and Vincent Gonguet and Virginie Do and Vish Vogeti and Vítor Albiero and Vladan Petrovic and Weiwei Chu and Wenhan Xiong and Wenyin Fu and Whitney Meers and Xavier Martinet and Xiaodong Wang and Xiaofang Wang and Xiaoqing Ellen Tan and Xide Xia and Xinfeng Xie and Xuchao Jia and Xuewei Wang and Yaelle Goldschlag and Yashesh Gaur and Yasmine Babaei and Yi Wen and Yiwen Song and Yuchen Zhang and Yue Li and Yuning Mao and Zacharie Delpierre Coudert and Zheng Yan and Zhengxing Chen and Zoe Papakipos and Aaditya Singh and Aayushi Srivastava and Abha Jain and Adam Kelsey and Adam Shajnfeld and Adithya Gangidi and Adolfo Victoria and Ahuva Goldstand and Ajay Menon and Ajay Sharma and Alex Boesenberg and Alexei Baevski and Allie Feinstein and Amanda Kallet and Amit Sangani and Amos Teo and Anam Yunus and Andrei Lupu and Andres Alvarado and Andrew Caples and Andrew Gu and Andrew Ho and Andrew Poulton and Andrew Ryan and Ankit Ramchandani and Annie Dong and Annie Franco and Anuj Goyal and Aparajita Saraf and Arkabandhu Chowdhury and Ashley Gabriel and Ashwin Bharambe and Assaf Eisenman and Azadeh Yazdan and Beau James and Ben Maurer and Benjamin Leonhardi and Bernie Huang and Beth Loyd and Beto De Paola and Bhargavi Paranjape and Bing Liu and Bo Wu and Boyu Ni and Braden Hancock and Bram Wasti and Brandon Spence and Brani Stojkovic and Brian Gamido and Britt Montalvo and Carl Parker and Carly Burton and Catalina Mejia and Ce Liu and Changhan Wang and Changkyu Kim and Chao Zhou and Chester Hu and Ching-Hsiang Chu and Chris Cai and Chris Tindal and Christoph Feichtenhofer and Cynthia Gao and Damon Civin and Dana Beaty and Daniel Kreymer and Daniel Li and David Adkins and David Xu and Davide Testuggine and Delia David and Devi Parikh and Diana Liskovich and Didem Foss and Dingkang Wang and Duc Le and Dustin Holland and Edward Dowling and Eissa Jamil and Elaine Montgomery and Eleonora Presani and Emily Hahn and Emily Wood and Eric-Tuan Le and Erik Brinkman and Esteban Arcaute and Evan Dunbar and Evan Smothers and Fei Sun and Felix Kreuk and Feng Tian and Filippos Kokkinos and Firat Ozgenel and Francesco Caggioni and Frank Kanayet and Frank Seide and Gabriela Medina Florez and Gabriella Schwarz and Gada Badeer and Georgia Swee and Gil Halpern and Grant Herman and Grigory Sizov and Guangyi and Zhang and Guna Lakshminarayanan and Hakan Inan and Hamid Shojanazeri and Han Zou and Hannah Wang and Hanwen Zha and Haroun Habeeb and Harrison Rudolph and Helen Suk and Henry Aspegren and Hunter Goldman and Hongyuan Zhan and Ibrahim Damlaj and Igor Molybog and Igor Tufanov and Ilias Leontiadis and Irina-Elena Veliche and Itai Gat and Jake Weissman and James Geboski and James Kohli and Janice Lam and Japhet Asher and Jean-Baptiste Gaya and Jeff Marcus and Jeff Tang and Jennifer Chan and Jenny Zhen and Jeremy Reizenstein and Jeremy Teboul and Jessica Zhong and Jian Jin and Jingyi Yang and Joe Cummings and Jon Carvill and Jon Shepard and Jonathan McPhie and Jonathan Torres and Josh Ginsburg and Junjie Wang and Kai Wu and Kam Hou U and Karan Saxena and Kartikay Khandelwal and Katayoun Zand and Kathy Matosich and Kaushik Veeraraghavan and Kelly Michelena and Keqian Li and Kiran Jagadeesh and Kun Huang and Kunal Chawla and Kyle Huang and Lailin Chen and Lakshya Garg and Lavender A and Leandro Silva and Lee Bell and Lei Zhang and Liangpeng Guo and Licheng Yu and Liron Moshkovich and Luca Wehrstedt and Madian Khabsa and Manav Avalani and Manish Bhatt and Martynas Mankus and Matan Hasson and Matthew Lennie and Matthias Reso and Maxim Groshev and Maxim Naumov and Maya Lathi and Meghan Keneally and Miao Liu and Michael L. Seltzer and Michal Valko and Michelle Restrepo and Mihir Patel and Mik Vyatskov and Mikayel Samvelyan and Mike Clark and Mike Macey and Mike Wang and Miquel Jubert Hermoso and Mo Metanat and Mohammad Rastegari and Munish Bansal and Nandhini Santhanam and Natascha Parks and Natasha White and Navyata Bawa and Nayan Singhal and Nick Egebo and Nicolas Usunier and Nikhil Mehta and Nikolay Pavlovich Laptev and Ning Dong and Norman Cheng and Oleg Chernoguz and Olivia Hart and Omkar Salpekar and Ozlem Kalinli and Parkin Kent and Parth Parekh and Paul Saab and Pavan Balaji and Pedro Rittner and Philip Bontrager and Pierre Roux and Piotr Dollar and Polina Zvyagina and Prashant Ratanchandani and Pritish Yuvraj and Qian Liang and Rachad Alao and Rachel Rodriguez and Rafi Ayub and Raghotham Murthy and Raghu Nayani and Rahul Mitra and Rangaprabhu Parthasarathy and Raymond Li and Rebekkah Hogan and Robin Battey and Rocky Wang and Russ Howes and Ruty Rinott and Sachin Mehta and Sachin Siby and Sai Jayesh Bondu and Samyak Datta and Sara Chugh and Sara Hunt and Sargun Dhillon and Sasha Sidorov and Satadru Pan and Saurabh Mahajan and Saurabh Verma and Seiji Yamamoto and Sharadh Ramaswamy and Shaun Lindsay and Shaun Lindsay and Sheng Feng and Shenghao Lin and Shengxin Cindy Zha and Shishir Patil and Shiva Shankar and Shuqiang Zhang and Shuqiang Zhang and Sinong Wang and Sneha Agarwal and Soji Sajuyigbe and Soumith Chintala and Stephanie Max and Stephen Chen and Steve Kehoe and Steve Satterfield and Sudarshan Govindaprasad and Sumit Gupta and Summer Deng and Sungmin Cho and Sunny Virk and Suraj Subramanian and Sy Choudhury and Sydney Goldman and Tal Remez and Tamar Glaser and Tamara Best and Thilo Koehler and Thomas Robinson and Tianhe Li and Tianjun Zhang and Tim Matthews and Timothy Chou and Tzook Shaked and Varun Vontimitta and Victoria Ajayi and Victoria Montanez and Vijai Mohan and Vinay Satish Kumar and Vishal Mangla and Vlad Ionescu and Vlad Poenaru and Vlad Tiberiu Mihailescu and Vladimir Ivanov and Wei Li and Wenchen Wang and Wenwen Jiang and Wes Bouaziz and Will Constable and Xiaocheng Tang and Xiaojian Wu and Xiaolan Wang and Xilun Wu and Xinbo Gao and Yaniv Kleinman and Yanjun Chen and Ye Hu and Ye Jia and Ye Qi and Yenda Li and Yilin Zhang and Ying Zhang and Yossi Adi and Youngjin Nam and Yu and Wang and Yu Zhao and Yuchen Hao and Yundi Qian and Yunlu Li and Yuzi He and Zach Rait and Zachary DeVito and Zef Rosnbrick and Zhaoduo Wen and Zhenyu Yang and Zhiwei Zhao and Zhiyu Ma},
      year={2024},
      eprint={2407.21783},
      archivePrefix={arXiv},
      primaryClass={cs.AI},
      url={https://arxiv.org/abs/2407.21783}, 
}

@misc{yang2025qwen3technicalreport,
      title={Qwen3 Technical Report}, 
      author={An Yang and Anfeng Li and Baosong Yang and Beichen Zhang and Binyuan Hui and Bo Zheng and Bowen Yu and Chang Gao and Chengen Huang and Chenxu Lv and Chujie Zheng and Dayiheng Liu and Fan Zhou and Fei Huang and Feng Hu and Hao Ge and Haoran Wei and Huan Lin and Jialong Tang and Jian Yang and Jianhong Tu and Jianwei Zhang and Jianxin Yang and Jiaxi Yang and Jing Zhou and Jingren Zhou and Junyang Lin and Kai Dang and Keqin Bao and Kexin Yang and Le Yu and Lianghao Deng and Mei Li and Mingfeng Xue and Mingze Li and Pei Zhang and Peng Wang and Qin Zhu and Rui Men and Ruize Gao and Shixuan Liu and Shuang Luo and Tianhao Li and Tianyi Tang and Wenbiao Yin and Xingzhang Ren and Xinyu Wang and Xinyu Zhang and Xuancheng Ren and Yang Fan and Yang Su and Yichang Zhang and Yinger Zhang and Yu Wan and Yuqiong Liu and Zekun Wang and Zeyu Cui and Zhenru Zhang and Zhipeng Zhou and Zihan Qiu},
      year={2025},
      eprint={2505.09388},
      archivePrefix={arXiv},
      primaryClass={cs.CL},
      url={https://arxiv.org/abs/2505.09388}, 
}

@misc{gemmateam2024gemma2improvingopen,
      title={Gemma 2: Improving Open Language Models at a Practical Size}, 
      author={Gemma Team and Morgane Riviere and Shreya Pathak and Pier Giuseppe Sessa and Cassidy Hardin and Surya Bhupatiraju and Léonard Hussenot and Thomas Mesnard and Bobak Shahriari and Alexandre Ramé and Johan Ferret and Peter Liu and Pouya Tafti and Abe Friesen and Michelle Casbon and Sabela Ramos and Ravin Kumar and Charline Le Lan and Sammy Jerome and Anton Tsitsulin and Nino Vieillard and Piotr Stanczyk and Sertan Girgin and Nikola Momchev and Matt Hoffman and Shantanu Thakoor and Jean-Bastien Grill and Behnam Neyshabur and Olivier Bachem and Alanna Walton and Aliaksei Severyn and Alicia Parrish and Aliya Ahmad and Allen Hutchison and Alvin Abdagic and Amanda Carl and Amy Shen and Andy Brock and Andy Coenen and Anthony Laforge and Antonia Paterson and Ben Bastian and Bilal Piot and Bo Wu and Brandon Royal and Charlie Chen and Chintu Kumar and Chris Perry and Chris Welty and Christopher A. Choquette-Choo and Danila Sinopalnikov and David Weinberger and Dimple Vijaykumar and Dominika Rogozińska and Dustin Herbison and Elisa Bandy and Emma Wang and Eric Noland and Erica Moreira and Evan Senter and Evgenii Eltyshev and Francesco Visin and Gabriel Rasskin and Gary Wei and Glenn Cameron and Gus Martins and Hadi Hashemi and Hanna Klimczak-Plucińska and Harleen Batra and Harsh Dhand and Ivan Nardini and Jacinda Mein and Jack Zhou and James Svensson and Jeff Stanway and Jetha Chan and Jin Peng Zhou and Joana Carrasqueira and Joana Iljazi and Jocelyn Becker and Joe Fernandez and Joost van Amersfoort and Josh Gordon and Josh Lipschultz and Josh Newlan and Ju-yeong Ji and Kareem Mohamed and Kartikeya Badola and Kat Black and Katie Millican and Keelin McDonell and Kelvin Nguyen and Kiranbir Sodhia and Kish Greene and Lars Lowe Sjoesund and Lauren Usui and Laurent Sifre and Lena Heuermann and Leticia Lago and Lilly McNealus and Livio Baldini Soares and Logan Kilpatrick and Lucas Dixon and Luciano Martins and Machel Reid and Manvinder Singh and Mark Iverson and Martin Görner and Mat Velloso and Mateo Wirth and Matt Davidow and Matt Miller and Matthew Rahtz and Matthew Watson and Meg Risdal and Mehran Kazemi and Michael Moynihan and Ming Zhang and Minsuk Kahng and Minwoo Park and Mofi Rahman and Mohit Khatwani and Natalie Dao and Nenshad Bardoliwalla and Nesh Devanathan and Neta Dumai and Nilay Chauhan and Oscar Wahltinez and Pankil Botarda and Parker Barnes and Paul Barham and Paul Michel and Pengchong Jin and Petko Georgiev and Phil Culliton and Pradeep Kuppala and Ramona Comanescu and Ramona Merhej and Reena Jana and Reza Ardeshir Rokni and Rishabh Agarwal and Ryan Mullins and Samaneh Saadat and Sara Mc Carthy and Sarah Cogan and Sarah Perrin and Sébastien M. R. Arnold and Sebastian Krause and Shengyang Dai and Shruti Garg and Shruti Sheth and Sue Ronstrom and Susan Chan and Timothy Jordan and Ting Yu and Tom Eccles and Tom Hennigan and Tomas Kocisky and Tulsee Doshi and Vihan Jain and Vikas Yadav and Vilobh Meshram and Vishal Dharmadhikari and Warren Barkley and Wei Wei and Wenming Ye and Woohyun Han and Woosuk Kwon and Xiang Xu and Zhe Shen and Zhitao Gong and Zichuan Wei and Victor Cotruta and Phoebe Kirk and Anand Rao and Minh Giang and Ludovic Peran and Tris Warkentin and Eli Collins and Joelle Barral and Zoubin Ghahramani and Raia Hadsell and D. Sculley and Jeanine Banks and Anca Dragan and Slav Petrov and Oriol Vinyals and Jeff Dean and Demis Hassabis and Koray Kavukcuoglu and Clement Farabet and Elena Buchatskaya and Sebastian Borgeaud and Noah Fiedel and Armand Joulin and Kathleen Kenealy and Robert Dadashi and Alek Andreev},
      year={2024},
      eprint={2408.00118},
      archivePrefix={arXiv},
      primaryClass={cs.CL},
      url={https://arxiv.org/abs/2408.00118}, 
}

@misc{jiang2023mistral7b,
      title={Mistral 7B}, 
      author={Albert Q. Jiang and Alexandre Sablayrolles and Arthur Mensch and Chris Bamford and Devendra Singh Chaplot and Diego de las Casas and Florian Bressand and Gianna Lengyel and Guillaume Lample and Lucile Saulnier and Lélio Renard Lavaud and Marie-Anne Lachaux and Pierre Stock and Teven Le Scao and Thibaut Lavril and Thomas Wang and Timothée Lacroix and William El Sayed},
      year={2023},
      eprint={2310.06825},
      archivePrefix={arXiv},
      primaryClass={cs.CL},
      url={https://arxiv.org/abs/2310.06825}, 
}
\clearpage
\appendix

\section{Dataset Source Rationale and Construction Details}
\label{sec:dataset_details}

This section details dataset sources, construction, and review procedures. Table~\ref{tab:dataset_construction_models} summarizes the generation and review models.

\begin{table*}[t]
\centering
\small
\renewcommand{\arraystretch}{1.12}
\setlength{\tabcolsep}{5pt}
\begin{tabularx}{\textwidth}{
    @{}
    >{\raggedright\arraybackslash}p{2.8cm}
    >{\raggedright\arraybackslash}p{2.0cm}
    >{\raggedright\arraybackslash}X
    >{\raggedright\arraybackslash}p{2.3cm}
    >{\raggedright\arraybackslash}X
    @{}
}
\toprule
\textbf{Component}
& \textbf{Generation model}
& \textbf{Generation role}
& \textbf{Review model}
& \textbf{Review role} \\
\midrule
Original corpus and benchmark-source held-out set
& GPT-4.1-mini
& Provisional labeling of BFCL candidates, synthetic
  supplementation, and adaptation of $\tau$-bench contexts.
& Gemini 2.5 Flash
& Three-axis label prediction and query-quality checks. \\

\addlinespace[7pt]
Domain-scenario dataset
& GPT-5-mini
& Generation of eight-class scenario families and targeted
  revision of rejected queries.
& Gemini 2.5 Flash
& Item-level label and quality checks, plus family-level
  coherence and coverage checks. \\

\addlinespace[7pt]
Deceptive evaluation set
& GPT-5
& Generation of deceptive queries with a specified target
  axis, gold label, and misleading surface cue.
& Claude Sonnet 5
& Blind three-axis labeling and checks of label correctness,
  naturalness, ambiguity, grounding, and deceptive-cue validity. \\
\bottomrule
\end{tabularx}
\caption{Roles of the generation and automated review models.
For benchmark-derived data, labeling refers to assigning initial
TACIT labels to existing queries, adaptation refers to rewriting
benchmark task contexts into requests matching a target TACIT
class, and supplementation refers to generating additional
queries where class coverage is insufficient. Original benchmark
queries are not generated by these models.}
\label{tab:dataset_construction_models}
\end{table*}

\begin{table*}[t]
\centering
\small
\renewcommand{\arraystretch}{1.12}
\setlength{\tabcolsep}{5pt}
\begin{tabularx}{\textwidth}{
    @{}
    >{\raggedright\arraybackslash}p{1.7cm}
    >{\raggedright\arraybackslash}p{2.2cm}
    >{\raggedright\arraybackslash}X
    >{\raggedright\arraybackslash}p{2.2cm}
    >{\raggedright\arraybackslash}X
    @{}
}
\toprule
\textbf{Class}
& \textbf{Original-corpus source}
& \textbf{Selection rationale}
& \textbf{Held-out source}
& \textbf{Selection rationale} \\
\midrule
(A,R,O)
& TriviaQA
& Stable factual questions requiring retrieval of
  available knowledge.
& SimpleQA
& Factual questions from a separate benchmark source. \\

\addlinespace[5pt]
(A,T,O)
& GSM8K
& Self-contained mathematical problems requiring
  transformation of supplied information.
& MATH
& Mathematical problems from a separate benchmark source. \\

\addlinespace[5pt]
(E,R,O)
& FreshQA
& Fast-changing factual questions requiring external
  information.
& RealTimeQA
& Time-sensitive questions from a separate benchmark source. \\

\addlinespace[5pt]
(E,T,O)
& Spider
& Structured-data questions requiring operations over
  external records.
& WikiSQL
& Table questions involving aggregation over external records. \\

\addlinespace[5pt]
Mutate classes
& BFCL v3 and supplementation
& Class-matching state-changing requests, supplemented
  where benchmark coverage is insufficient.
& $\tau$-bench contexts and supplementation
& Requests adapted from airline and retail task contexts,
  with additional examples to cover all four Mutate classes. \\
\bottomrule
\end{tabularx}
\caption{Benchmark sources informing the original corpus and
the 200-query benchmark-source held-out set. The domain-scenario
dataset is constructed separately using the domain specifications
in Table~\ref{tab:domain_inventory}.}
\label{tab:dataset_source_rationale}
\end{table*}

\subsection{Original Corpus: Sources, Sampling, and Review}

The original corpus contains 1,600 queries, with 200 per TACIT class. It serves as a source of training examples for the current experiments: we sample 100 queries per class from this corpus, yielding 800 queries, and combine them with 800 domain-scenario training queries.

For the Observe classes, we construct candidate pools from the corresponding sources in Table~\ref{tab:dataset_source_rationale}, generally targeting approximately 400 candidates per class before review. Because the fast-changing FreshQA subset provides limited coverage for (E,R,O), we supplement its candidates with GPT-4.1-mini-generated queries, targeting a larger pool of approximately 600 candidates.

For the Mutate classes, we extract requests from BFCL v3, remove duplicates, and assign provisional three-axis labels using GPT-4.1-mini. We supplement underrepresented classes with target-class-conditioned synthesis, targeting approximately 400 candidates per class before review. The class-conditioned synthesis prompts are provided in Appendix~\ref{sec:synthesis_prompts}.

Gemini 2.5 Flash independently predicts the three capability labels and assesses query quality using the prompt in Appendix~\ref{sec:review_prompt}. Label disagreements and insufficient-quality candidates are flagged for further inspection. After review and curation, we retain 200 examples per class, prioritizing benchmark-derived examples where available.

\subsection{Domain-Scenario Dataset}
\label{sec:domain_scenario_details}

\paragraph{Domain selection and specifications.}
The domain-scenario dataset contains 1,680 newly generated queries across 25 domains. Table~\ref{tab:domain_inventory} lists the domains and the benchmark settings or API workflows that informed their design. These settings guide the construction of domain specifications rather than supplying benchmark utterances to the dataset.

For each domain, we define a domain card specifying entities, stable facts or explicitly supplied information, external state, transformation operations, and state-changing actions. These specifications support systematic variation of the three capability axes within a shared domain. For example, a calendar domain can involve a supplied meeting time or current calendar availability, direct retrieval or time calculations, and information requests or event creation. 

\begin{table*}[t]
\centering
\small
\renewcommand{\arraystretch}{1.12}
\setlength{\tabcolsep}{5pt}
\begin{tabularx}{\textwidth}{
    @{}
    >{\raggedright\arraybackslash}p{2.4cm}
    >{\raggedright\arraybackslash}p{3.2cm}
    >{\raggedright\arraybackslash}X
    >{\centering\arraybackslash}p{0.8cm}
    @{}
}
\toprule
\textbf{Domain group}
& \textbf{Design inspiration}
& \textbf{Domains}
& \textbf{Count} \\
\midrule
Seen domains
& SGD-style task-oriented settings~\cite{rastogi2020towards}
& Banking; payment and transfer; calendar; events;
  flights; hotels; restaurants; rental cars;
  local transport; home services; movies; music and media.
& 12 \\

\addlinespace[7pt]
& BFCL-style API workflows
& Email and messaging; file and cloud storage;
  smart home; software development and DevOps.
& 4 \\

\addlinespace[7pt]
& Common application and API workflows
& Shopping cart; subscription management;
  project management; database and reporting.
& 4 \\

\midrule
Held-out domains
& $\tau$-bench-style support settings
& Retail order support; airline customer support.
& 2 \\

\addlinespace[7pt]
& Common application and API workflows
& Healthcare appointments; education and library;
  government services.
& 3 \\
\bottomrule
\end{tabularx}
\caption{The 25 domains in the domain-scenario dataset: 20 for training, validation, and in-domain testing, and 5 for held out within this component. Design inspirations guide domain cards, not copied queries; held-out domains may overlap with topics or operations in the original corpus.}
\label{tab:domain_inventory}
\end{table*}

\paragraph{Scenario-family construction.}
We organize the dataset into 210 scenario families. Each family contains eight queries, one for each TACIT class, within a shared domain and broad scenario. This construction ensures that every class is represented within each family, reducing the association between a domain and a particular capability label.

GPT-5-mini generates queries from the domain cards, scenario specifications, and formal taxonomy definitions. Generation instructions vary syntactic forms, transformation mechanisms, and state-changing operations across families. The eight queries share a scenario but are separately formulated requests; they are not required to be minimal lexical edits of one another.

\paragraph{Automated review and targeted revision.}
Gemini 2.5 Flash reviews the generated data at both the query and family levels. Query-level review checks the three capability labels, naturalness, and ambiguity. Family-level review checks scenario coherence, coverage of all eight classes, and whether any query is forced or contrived.

The initial item-level review accepted 1,635 of 1,680 queries (97.32\%). Combining item-level failures with family-level concerns identified 52 queries for targeted revision. GPT-5-mini revised these queries while preserving their assigned domain, family, split, and target class. Revised queries and affected families were then reviewed again.

The final dataset contains 210 families and 1,680 queries passing the automated acceptance criteria.

The generation, review, and revision prompt templates are provided in Appendix~\ref{sec:domain_scenario_prompts}.

\paragraph{Split construction.}
Within each of the 20 seen domains, we allocate five scenario families to training, one to validation, and two to in-domain testing. This yields 800 training queries, 160 validation queries, and 320 in-domain test queries. Each of the five held-out domains contributes ten families, yielding 400 domain-held-out test queries.

All eight queries belonging to a family remain in the same split. Thus, family membership does not overlap across training, validation, and testing, although broad scenario themes and operations can recur.

The final training set combines the 800 original-corpus queries with the 800 domain-scenario training queries, for 1,600 queries in total and 200 per class. Validation uses only the separate 160-query domain-scenario validation set.

\subsection{Deceptive Evaluation Set}
\label{sec:deceptive_details}

The deceptive evaluation set contains 220 queries designed to test robustness to misleading surface cues: 77 target Source, 70 target Transformation, and 73 target World Effect. Each query includes wording that suggests an incorrect label on the target axis while retaining a determinate gold label under the formal taxonomy. GPT-5 generates candidate queries from specifications of the target axis, gold label, and deceptive mechanism. 

Claude Sonnet 5 independently predicts each query's three capability labels without seeing the proposed gold labels. Additional review checks label correctness, naturalness, ambiguity, and the presence of an identifiable misleading cue. Candidates are also checked for unsupported premises and excessive overlap with generation examples or templates. The generation and review prompts are provided in Appendix~\ref{sec:deceptive_prompts}.

The final evaluation set contains 220 queries that passed these acceptance checks. We freeze the set before evaluation and do not select or filter queries based on probe or baseline performance. The deceptive set is not used for probe training or layer selection.

\section{Layer-wise Probe Accuracy}
\label{sec:appendix_layer_sweep}

Table~\ref{tab:layer_sweep_all} reports the full per-layer validation accuracy for each capability axis probe across all four models.
Bold values indicate the jointly-optimal best layer selected for each axis.
When multiple layers attain the same peak per-axis validation accuracy, we select the combination across all three axes that maximizes joint (three-axis simultaneous) validation accuracy; the bolded layers are the result of this selection.

\section{Prompts}
\label{sec:appendix}

\subsection{Capability Axis Definitions}
\label{sec:taxonomy_defs}

This block provides the formal definitions underlying Section~\ref{sec:taxonomy}. It serves as the shared taxonomy for benchmark-query labeling and review, query synthesis, domain-scenario construction, deceptive-query generation and review, and verbal classification. Task-specific instructions supplement these definitions in the prompts below.

\clearpage
\onecolumn
\small
\begin{longtable}{lrrr}
\toprule
\textbf{Layer}
& \textbf{Source (A/E)}
& \textbf{Transformation (R/T)}
& \textbf{World Effect (O/M)} \\
\midrule
\endfirsthead

\multicolumn{4}{l}{\small\textit{(continued from previous page)}} \\
\toprule
\textbf{Layer}
& \textbf{Source (A/E)}
& \textbf{Transformation (R/T)}
& \textbf{World Effect (O/M)} \\
\midrule
\endhead

\midrule
\multicolumn{4}{r}{\small\textit{(continued on next page)}} \\
\endfoot

\endlastfoot

\multicolumn{4}{l}{\textit{Llama 3.1 8B Instruct}
(32 layers; joint validation accuracy = 0.9813)} \\
\multicolumn{4}{l}{Best layers:
S = L14 (43.75\%), T = L7 (21.88\%), W = L16 (50.00\%)} \\
\midrule
L1 & 0.9000 & 0.9250 & 0.8688 \\
L2 & 0.9438 & 0.9500 & 0.8813 \\
L3 & 0.9000 & 0.9188 & 0.8875 \\
L4 & 0.9000 & 0.9688 & 0.8875 \\
L5 & 0.9188 & 0.9750 & 0.9563 \\
L6 & 0.9438 & 0.9875 & 0.9313 \\
L7 & 0.9438 & \textbf{0.9938} & 0.9813 \\
L8 & 0.9375 & 0.9938 & 0.9875 \\
L9 & 0.9375 & 0.9875 & 0.9500 \\
L10 & 0.9625 & 0.9875 & 0.9625 \\
L11 & 0.9750 & 0.9875 & 0.9750 \\
L12 & 0.9625 & 0.9938 & 0.9813 \\
L13 & 0.9688 & 0.9875 & 0.9813 \\
L14 & \textbf{0.9938} & 0.9813 & 0.9875 \\
L15 & 0.9813 & 0.9750 & 0.9750 \\
L16 & 0.9875 & 0.9813 & \textbf{0.9938} \\
L17 & 0.9875 & 0.9813 & 0.9875 \\
L18 & 0.9875 & 0.9813 & 0.9875 \\
L19 & 0.9875 & 0.9750 & 0.9938 \\
L20 & 0.9875 & 0.9750 & 0.9875 \\
L21 & 0.9875 & 0.9813 & 0.9875 \\
L22 & 0.9875 & 0.9813 & 0.9938 \\
L23 & 0.9875 & 0.9813 & 0.9875 \\
L24 & 0.9813 & 0.9813 & 0.9875 \\
L25 & 0.9813 & 0.9813 & 0.9875 \\
L26 & 0.9750 & 0.9750 & 0.9813 \\
L27 & 0.9813 & 0.9813 & 0.9813 \\
L28 & 0.9688 & 0.9813 & 0.9813 \\
L29 & 0.9750 & 0.9750 & 0.9875 \\
L30 & 0.9625 & 0.9688 & 0.9813 \\
L31 & 0.9688 & 0.9688 & 0.9875 \\
L32 & 0.9625 & 0.9625 & 0.9875 \\

\midrule
\multicolumn{4}{l}{\textit{Qwen 3 8B}
(36 layers; joint validation accuracy = 0.9875)} \\
\multicolumn{4}{l}{Best layers:
S = L36 (100.00\%), T = L20 (55.56\%), W = L26 (72.22\%)} \\
\midrule
L1 & 0.9063 & 0.9125 & 0.8313 \\
L2 & 0.9125 & 0.9188 & 0.8188 \\
L3 & 0.9188 & 0.9375 & 0.8313 \\
L4 & 0.9188 & 0.9375 & 0.8438 \\
L5 & 0.8938 & 0.9375 & 0.8438 \\
L6 & 0.9188 & 0.9188 & 0.8625 \\
L7 & 0.9000 & 0.9375 & 0.8438 \\
L8 & 0.9000 & 0.9438 & 0.9188 \\
L9 & 0.9063 & 0.9750 & 0.9313 \\
L10 & 0.8813 & 0.9625 & 0.9625 \\
L11 & 0.8875 & 0.9813 & 0.9625 \\
L12 & 0.9250 & 0.9625 & 0.9563 \\
L13 & 0.9125 & 0.9625 & 0.9625 \\
L14 & 0.9188 & 0.9500 & 0.9750 \\
L15 & 0.9188 & 0.9688 & 0.9750 \\
L16 & 0.9250 & 0.9563 & 0.9688 \\
L17 & 0.9250 & 0.9688 & 0.9750 \\
L18 & 0.9188 & 0.9625 & 0.9813 \\
L19 & 0.9438 & 0.9875 & 0.9813 \\
L20 & 0.9375 & \textbf{0.9938} & 0.9938 \\
L21 & 0.9500 & 0.9875 & 0.9938 \\
L22 & 0.9500 & 0.9875 & 0.9938 \\
L23 & 0.9625 & 0.9875 & 0.9938 \\
L24 & 0.9625 & 0.9875 & 0.9938 \\
L25 & 0.9688 & 0.9938 & 0.9938 \\
L26 & 0.9625 & 0.9813 & \textbf{1.0000} \\
L27 & 0.9625 & 0.9813 & 0.9875 \\
L28 & 0.9688 & 0.9813 & 0.9875 \\
L29 & 0.9750 & 0.9875 & 0.9875 \\
L30 & 0.9688 & 0.9875 & 0.9875 \\
L31 & 0.9625 & 0.9875 & 0.9938 \\
L32 & 0.9750 & 0.9875 & 0.9875 \\
L33 & 0.9813 & 0.9875 & 0.9875 \\
L34 & 0.9813 & 0.9875 & 0.9875 \\
L35 & 0.9813 & 0.9875 & 0.9938 \\
L36 & \textbf{0.9938} & 0.9875 & 0.9875 \\

\midrule
\multicolumn{4}{l}{\textit{Gemma 2 9B IT}
(42 layers; joint validation accuracy = 0.9875)} \\
\multicolumn{4}{l}{Best layers:
S = L23 (54.76\%), T = L10 (23.81\%), W = L23 (54.76\%)} \\
\midrule
L1 & 0.9000 & 0.8875 & 0.8125 \\
L2 & 0.9375 & 0.9563 & 0.8813 \\
L3 & 0.9500 & 0.9625 & 0.8875 \\
L4 & 0.9188 & 0.9750 & 0.8688 \\
L5 & 0.9563 & 0.9563 & 0.9000 \\
L6 & 0.9500 & 0.9625 & 0.8938 \\
L7 & 0.9563 & 0.9688 & 0.8875 \\
L8 & 0.9500 & 0.9688 & 0.9250 \\
L9 & 0.9563 & 0.9813 & 0.9563 \\
L10 & 0.9625 & \textbf{0.9938} & 0.9625 \\
L11 & 0.9625 & 0.9875 & 0.9625 \\
L12 & 0.9625 & 0.9875 & 0.9750 \\
L13 & 0.9625 & 0.9875 & 0.9750 \\
L14 & 0.9813 & 0.9875 & 0.9688 \\
L15 & 0.9438 & 0.9938 & 0.9688 \\
L16 & 0.9500 & 0.9938 & 0.9750 \\
L17 & 0.9563 & 0.9938 & 0.9875 \\
L18 & 0.9688 & 0.9938 & 0.9938 \\
L19 & 0.9563 & 0.9938 & 0.9938 \\
L20 & 0.9625 & 0.9938 & 0.9938 \\
L21 & 0.9750 & 0.9938 & 0.9938 \\
L22 & 0.9875 & 0.9875 & 0.9938 \\
L23 & \textbf{0.9938} & 0.9938 & \textbf{1.0000} \\
L24 & 0.9875 & 0.9938 & 0.9938 \\
L25 & 0.9875 & 0.9875 & 0.9938 \\
L26 & 0.9938 & 0.9813 & 0.9938 \\
L27 & 0.9875 & 0.9813 & 0.9875 \\
L28 & 0.9875 & 0.9813 & 0.9875 \\
L29 & 0.9813 & 0.9875 & 0.9875 \\
L30 & 0.9875 & 0.9813 & 0.9875 \\
L31 & 0.9875 & 0.9813 & 0.9875 \\
L32 & 0.9750 & 0.9875 & 0.9813 \\
L33 & 0.9813 & 0.9875 & 0.9813 \\
L34 & 0.9875 & 0.9875 & 0.9750 \\
L35 & 0.9813 & 0.9875 & 0.9750 \\
L36 & 0.9750 & 0.9875 & 0.9813 \\
L37 & 0.9688 & 0.9875 & 0.9813 \\
L38 & 0.9688 & 0.9875 & 0.9750 \\
L39 & 0.9688 & 0.9813 & 0.9813 \\
L40 & 0.9625 & 0.9813 & 0.9813 \\
L41 & 0.9688 & 0.9813 & 0.9813 \\
L42 & 0.9688 & 0.9813 & 0.9813 \\

\midrule
\multicolumn{4}{l}{\textit{Mistral 7B Instruct v0.3}
(32 layers; joint validation accuracy = 0.9938)} \\
\multicolumn{4}{l}{Best layers:
S = L20 (62.50\%), T = L8 (25.00\%), W = L22 (68.75\%)} \\
\midrule
L1 & 0.9313 & 0.9250 & 0.8688 \\
L2 & 0.8875 & 0.9125 & 0.9125 \\
L3 & 0.9188 & 0.9688 & 0.8625 \\
L4 & 0.9125 & 0.9625 & 0.8813 \\
L5 & 0.9000 & 0.9688 & 0.9000 \\
L6 & 0.9125 & 0.9875 & 0.9125 \\
L7 & 0.9688 & 0.9875 & 0.9625 \\
L8 & 0.9375 & \textbf{1.0000} & 0.9438 \\
L9 & 0.9563 & 1.0000 & 0.9438 \\
L10 & 0.9625 & 0.9938 & 0.9563 \\
L11 & 0.9188 & 0.9938 & 0.9750 \\
L12 & 0.9563 & 0.9875 & 0.9875 \\
L13 & 0.9563 & 0.9938 & 0.9875 \\
L14 & 0.9750 & 0.9938 & 0.9813 \\
L15 & 0.9938 & 0.9875 & 0.9875 \\
L16 & 0.9875 & 0.9875 & 0.9875 \\
L17 & 0.9875 & 0.9875 & 0.9813 \\
L18 & 0.9938 & 0.9875 & 0.9813 \\
L19 & 0.9875 & 0.9875 & 0.9813 \\
L20 & \textbf{1.0000} & 0.9875 & 0.9813 \\
L21 & 1.0000 & 0.9875 & 0.9813 \\
L22 & 1.0000 & 0.9875 & \textbf{0.9938} \\
L23 & 1.0000 & 0.9875 & 0.9875 \\
L24 & 1.0000 & 0.9875 & 0.9875 \\
L25 & 1.0000 & 0.9875 & 0.9875 \\
L26 & 0.9938 & 0.9875 & 0.9875 \\
L27 & 1.0000 & 0.9875 & 0.9938 \\
L28 & 0.9938 & 0.9875 & 0.9938 \\
L29 & 0.9875 & 0.9813 & 0.9938 \\
L30 & 0.9938 & 0.9875 & 0.9938 \\
L31 & 0.9875 & 0.9813 & 0.9938 \\
L32 & 0.9875 & 0.9813 & 0.9938 \\
\bottomrule
\caption{Per-layer validation accuracy across four models ($N=160$). Bold entries indicate the selected best layers. Parenthesized percentages denote relative depth, $l/N_{\mathrm{layers}}$.}
\label{tab:layer_sweep_all}
\end{longtable}

\begin{promptbox}{Capability Axis Definitions}
\begin{lstlisting}[style=promptinner]
Axis 1 - Source (where does the information needed to IDENTIFY THE TARGET and CONSTRUCT the action parameters come from?):
  A (Available): The action target is a SPECIFIC NAMED INSTANCE and ALL required parameters are already explicit in the query (or trivial parametric facts).
    No search across alternatives is required to know WHAT to act on. The information must also be STABLE -- not time-varying.
    Examples:
      - 'play "wrecking ball" by Miley Cyrus' -> song uniquely named -> A
      - 'send $250 to Rachel' -> recipient named, amount given -> A
      - 'delete file /tmp/foo.txt' -> specific path named -> A
      - 'set alarm for 7:00 AM labeled morning workout' -> fully specified -> A
      - 'connect to Bluetooth speaker JBL Flip 4' -> device named -> A
  E (External): The action target is described by CRITERIA, not by a specific named instance, so the model must first OBSERVE or FETCH external state to enumerate candidates and pick the matching one(s). 
    Anything requiring real-time data, live database/API calls, web search, availability/schedule lookup, ranking by current values, or 'find/locate/check then act' is E.
    ALSO E: ANY value that can change over time -- even if the model might have seen it in training, if it is time-sensitive (prices, weather, current balances, live status, who currently holds a position) classify as E.
    Examples:
      - 'Book a direct flight from SF to London on 2022-04-27 afternoon' -> must search flights matching the criteria -> E
      - 'Book a hotel for 2 adults in Paris July 10-20' -> must search hotel availability -> E
      - 'order a pizza from the closest Domino's' -> must look up nearest store -> E
      - 'Find available slots at the dental clinic and book one' -> explicit lookup -> E
      - 'check my balance, then transfer the rest to Alice' -> balance fetched -> E

  CRITICAL DISTINCTIONS:
    1. Do NOT confuse execution venue with information source. The fact that an action talks to an external system at runtime does NOT make it E. 
       What matters is whether the model needs new external data to KNOW WHAT TO ACT ON.
    2. NAMED target + given params + stable info = A even if execution touches an external API.
    3. CRITERIA-described target (must enumerate candidates) = E even if all criteria are stated in the query.
    4. 'Send/post/email to <named recipient>' with given content = A.
       'Send to whoever has highest priority on the team' = E.

Axis 2 - Transformation (what processing is done on the information?):
  R (Retrieve-only): A single direct lookup; the result is returned as-is with no further computation. No arithmetic, no aggregation, no comparison between distinct entities, no conditional logic, no ranking across candidates.
    CRITICAL: Multiple named fields do NOT make a query T -- passing three explicit params from the query is still R. 
    Examples:
      - 'play Wrecking Ball by Miley Cyrus' -> passthrough -> R
      - 'book the first available slot' -> top-1 fetch -> R
      - 'add Postgres server host X port 5432' -> passthrough -> R
      - 'birthplace of the current Speaker' -> reference chain, no computation -> R
      - 'Is X available?' -> boolean check -> R
      - 'Current balance' -> single current-state lookup -> R
  T (Transform): A non-trivial computation is required. T includes ANY of:
    (a) Arithmetic: +, -, x, /, percentage, ratio, date arithmetic (age, duration), unit conversion via formula.
    (b) Aggregation: count (how many), sum, total, average over a set of records.
    (c) Superlative/ranking among candidates: 'most recent', 'oldest', 'largest', 'highest', 'lowest', 'best', 'worst', 'latest' -- requires ordering across multiple records -> T.
    (d) Comparison between two or more distinct entities.
    (e) Conditional/threshold logic: 'if X < 10', 'if rate exceeds Y'.
    (f) Logical inference, word puzzle, pattern recognition, counting occurrences.
    Examples:
      - 'most recent winner?' -> superlative -> ordering multiple records -> T
      - 'current age of X' -> date arithmetic -> T
      - 'how many goals has X scored?' -> count -> T
      - 'if stock < 10 order 20 more' -> logical inference -> T
      
  CRITICAL DISTINCTIONS:
    1. top-1 fetch, no comparison among options = R
    2. Following a reference chain (find Speaker -> find birthplace) = R if no computation occurs

Axis 3 - World-Effect (does the query request an external state change?):
  O (Observe): only information is reported; no external state changes. 
    O verbs: find, look for, search, show, check, tell, list, what/when/where.
  M (Mutate): a state-changing or hard-to-undo action is explicitly requested.
    M verbs: send, save (to external store), book, schedule, transfer, update, delete, post, connect, register, order, reserve, purchase, buy, email, cancel, pay, deploy, publish, apply.
  LITERAL VERB RULE: classify by the explicit verb in the query. Do NOT infer Mutate intent from context ('find a hotel' stays O even if booking is implied). 
  Same target with both verb types -> M wins ('check availability and book' -> M).
\end{lstlisting}
\end{promptbox}
\clearpage
\twocolumn

\subsection{Gold-Label Review Prompt}
\label{sec:review_prompt}

This prompt is used with Gemini 2.5 Flash to review the original benchmark-derived corpus and the benchmark-source held-out set. The reviewer predicts the three capability labels without access to the proposed labels. The pipeline flags label disagreements or \texttt{quality\_ok = false} for further inspection before dataset finalization. Domain-scenario review uses the separate prompts provided below.

\begin{promptbox}{System Message}
\begin{lstlisting}[style=promptinner]
You are a strict quality-control annotator for a capability-classification benchmark. 
Your primary job is to REJECT items that are ambiguous, borderline, or mislabelled -- not to confirm them. Approve only what is unambiguously correct.

Classify each user query along three independent axes:

[AXIS DEFINITIONS]

AXIS VALUE CONSTRAINT: Source must be A or E. Transformation must be R or T.
World-Effect must be O or M. Any other value is invalid -- do not output it.

STRICT REJECTION RULES -- set quality_ok=false if ANY of the following hold:
  1. You are confident the gold label on any axis is wrong given the definitions above.
  2. The query is genuinely ambiguous: even with the definitions above, a reasonable annotator could assign a different label.
  3. The query contains two or more independent, causally unrelated goals -- goals that could each be asked in a separate conversation without affecting each other. 
     NOTE: sequential steps that are natural prerequisites for a single final outcome are allowed (e.g. 'check availability then book' is one coherent intent).
     Reject only when the goals are genuinely disjoint (e.g. 'tell me the weather AND send an email to my boss').
  4. The query is vague, underspecified, or context-dependent -- correct labelling requires guessing unstated intent or missing context.
  5. Your confidence in the complete three-axis label is below 0.85.
  6. The query is malformed, nonsensical, or cannot be interpreted as a single coherent user request.

Set quality_ok=true ONLY when all three axis labels are unambiguously correct under the definitions above, the query has a single coherent user intent, 
and your confidence is 0.85 or higher. quality_issues must be non-empty whenever quality_ok=false.

Output ONLY a JSON object with keys: source, transformation, world_effect, confidence (0..1), quality_ok (boolean), quality_issues (array of short strings, required when quality_ok=false), rationale (<=2 sentences). 
Use exactly the single-letter codes A/E, R/T, O/M.
\end{lstlisting}
\end{promptbox}

\begin{promptbox}{User Message}
\begin{lstlisting}[style=promptinner]
Query: {query}

Classify strictly. When in doubt about any axis, reject.
\end{lstlisting}
\end{promptbox}

\subsection{Synthetic Supplementation Prompts}
\label{sec:synthesis_prompts}
GPT-4.1-mini is used to supplement classes with insufficient benchmark coverage in the original corpus. The following prompts cover the four Mutate classes supplemented from BFCL v3 and the (E,R,O) class supplemented from FreshQA. Each system prompt incorporates the shared Capability Axis Definitions (Appendix~\ref{sec:taxonomy_defs}) together with class-specific generation constraints.

\paragraph{ERO (External, Retrieve-only, Observe)}~\\

\begin{promptbox}{System Message}
\begin{lstlisting}[style=promptinner]
You are generating data for an academic capability-classification benchmark.
Each item must be a single, natural user turn.
The query must be exactly (External, Retrieve-only, Observe) under the taxonomy below:

[AXIS DEFINITIONS]

The example queries shown in the taxonomy above are illustrations of the labels only -- never copy or closely paraphrase them; every query you generate must be original.

AVOID these patterns:
  - 'Who is the most popular/successful/famous X?' -> comparison (ETO)
  - 'How many X have done Y?' -> COUNT (ETO)
  - 'What is the average/total X?' -> aggregation (ETO)
  - 'Who was the first X to do Y?' -> historical stable fact (ARO)
  - 'What is the capital of X?' -> never-changing (ARO)

GOOD patterns (vary these heavily):
  - 'What is the current [price / rate / rank / status] of [entity]?'
  - 'Who is the current [role] at [organization]?'
  - 'What [team / club / country] does [person] currently represent?'
  - 'Is [entity] currently [status / condition]?'
  - 'What is today's [exchange rate / index / price] of [X]?'
  - 'What is the current version of [software]?'

Vary domain across: sports, finance, politics, technology, entertainment,
business/corporate, science, weather/environment.
Vary phrasing: direct questions, conversational, 'Can you tell me...',
'I need to know...'.
Do NOT let multiple queries start with the same word.
\end{lstlisting}
\end{promptbox}

\begin{promptbox}{User Message Template}
\begin{lstlisting}[style=promptinner]
Generate {n} diverse user queries matching
(External, Retrieve-only, Observe).
Every query must be in English. Each query must open with a
DIFFERENT first word. Output a JSON array of objects with a
single key 'query'.

Example:
[{"query": "What team does Kylian Mbappe currently play for?"},
 {"query": "Who is the current CEO of Apple?"},
 {"query": "What is today's USD to KRW exchange rate?"},
 {"query": "Is the International Space Station currently manned?"},
 {"query": "What is the current version of the Android operating system?"}]
\end{lstlisting}
\end{promptbox}

\paragraph{ARM (Available, Retrieve-only, Mutate)}~\\

\begin{promptbox}{System Message}
\begin{lstlisting}[style=promptinner]
You are generating data for an academic capability-classification benchmark.
Each item must be a single, natural user turn that (i) provides a piece of information explicitly INSIDE the message itself, and (ii) requests that the assistant SAVE/UPDATE/REMEMBER that information in some store (profile, note, calendar, contacts, settings, memory). 
The query must be exactly (Available, Retrieve-only, Mutate) under the taxonomy below:

[AXIS DEFINITIONS]

The example queries shown in the taxonomy above are illustrations of the labels only -- never copy or closely paraphrase them; every query you generate must be original.

Avoid any external lookups (no 'check the weather', 'find current price', etc.).
Generate every query entirely in English. Vary persona, topic, and verbosity.
Do not produce duplicates.
\end{lstlisting}
\end{promptbox}

\begin{promptbox}{User Message Template}
\begin{lstlisting}[style=promptinner]
Generate {n} diverse user queries that match (Available, Retrieve-only, Mutate).
Every query must be written entirely in English. Output a JSON array of objects with a single key 'query'. 
Example:
[{"query": "My birthday is March 5. Save it to my profile."},
 {"query": "Save my new email alice@example.com to my contact card."}]
\end{lstlisting}
\end{promptbox}

\paragraph{ATM (Available, Transform, Mutate)}~\\

\begin{promptbox}{System Message}
\begin{lstlisting}[style=promptinner]
You are generating data for an academic capability-classification benchmark.
Each item must be a single user turn. The query must be exactly (Available,Transform, Mutate) under the taxonomy below:

[AXIS DEFINITIONS]

The example queries shown in the taxonomy above are illustrations of the labels only -- never copy or closely paraphrase them; every query you generate must be original.

Generate every query entirely in English. Vary the kind of calculation (percentages, unit conversions, ratios, sums, time arithmetic, currency conversion using a rate stated INSIDE the message).
\end{lstlisting}
\end{promptbox}

\begin{promptbox}{User Message Template}
\begin{lstlisting}[style=promptinner]
Generate {n} diverse user queries matching (Available, Transform, Mutate).
Every query must be written entirely in English. Output a JSON array of objects with a single key 'query'. 
Example:
[{"query": "Subtract 12% tax from my $3,800 salary and save the remaining amount to my household ledger."},
 {"query": "I ran 7.2 km, 6.5 km, and 8.1 km this week. Add the weekly total to my fitness log."}]
\end{lstlisting}
\end{promptbox}

\paragraph{ERM (External, Retrieve-only, Mutate)}~\\

\begin{promptbox}{System Message}
\begin{lstlisting}[style=promptinner]
You are generating data for an academic capability-classification benchmark.
Each item must be a single, natural user turn. The query must be exactly (External, Retrieve-only, Mutate) under the taxonomy below:

[AXIS DEFINITIONS]

The example queries shown in the taxonomy above are illustrations of the labels only -- never copy or closely paraphrase them; every query you generate must be original.

AVOID these patterns:
  - 'find the cheapest/fastest/largest' (comparison -> ETM)
  - 'if stock < 10 then order' (conditional -> ETM)
  - 'my address is X, save it' (no external lookup -> ARM)

IMPORTANT -- vary surface form strongly across items. Use a mix of:
  - Direct imperatives: 'Reserve...', 'Send...', 'Pull up...', 'Go ahead and
    book...'
  - Polite requests: 'Could you check... and schedule...?', 'Please look up...
    and confirm...'
  - Declarative setups: 'I need you to fetch... and send it to...'
  - Questions: 'Can you see if... is available and reserve it for me?'
  - Conversational: 'Just grab the latest... and forward it to...'
Do NOT let multiple queries start with the same word. Vary domain: travel, healthcare, food & dining, retail, calendar/scheduling, communication, HR, real estate, entertainment, logistics.
\end{lstlisting}
\end{promptbox}

\begin{promptbox}{User Message Template}
\begin{lstlisting}[style=promptinner]
Generate {n} diverse user queries matching (External, Retrieve-only, Mutate).
Every query must be written entirely in English. Each query must open with a DIFFERENT first word -- maximise variety in phrasing. Output a JSON array of objects with a single key 'query'. 
Example:
[{"query": "Check if there are any open slots with Dr. Lee tomorrow afternoon and book the first available one."},
 {"query": "Pull up today's lunch special at the office cafeteria and post it to the team channel"},
 {"query": "Reserve a table for three at any available Italian restaurant near downtown tonight."},
 {"query": "I need you to look at today's flight departures from JFK to LAX and confirm my seat on the next one."}]
\end{lstlisting}
\end{promptbox}

\paragraph{ETM (External, Transform, Mutate)}~\\

\begin{promptbox}{System Message}
\begin{lstlisting}[style=promptinner]
You are generating data for an academic capability-classification benchmark.
Each item must be a single, natural user turn. The query must be exactly (External, Transform, Mutate) under the taxonomy below:

[AXIS DEFINITIONS]

The example queries shown in the taxonomy above are illustrations of the labels only -- never copy or closely paraphrase them; every query you generate must be original.

All three must be clearly required by the query, not optional. The external lookup must come first (you can't act without it). 
The computation bridges the lookup and the action. Generate every query entirely in English. 
Vary domain: e-commerce, finance, travel, HR/admin, logistics, calendar scheduling, restaurant, healthcare.
\end{lstlisting}
\end{promptbox}

\begin{promptbox}{User Message Template}
\begin{lstlisting}[style=promptinner]
Generate {n} diverse user queries matching (External, Transform, Mutate).
Every query must be written entirely in English. Output a JSON array of objects with a single key 'query'. 
Example:
[{"query": "Check current inventory levels and automatically order 50 units of every item with fewer than 10 units in stock."},
 {"query": "Check today's EUR/USD rate and, if it's below 1.08, move $5000 from my savings to my investment account."}]
\end{lstlisting}
\end{promptbox}

\subsection{Domain-Scenario Generation and Review Prompts}
\label{sec:domain_scenario_prompts}

The domain-scenario dataset is generated with GPT-5-mini and reviewed with Gemini 2.5 Flash at both the query and scenario-family levels. GPT-5-mini subsequently revises queries flagged by either review stage. Dataset composition and acceptance criteria are described in Appendix~\ref{sec:domain_scenario_details}.

The following templates reproduce the generation, review, and revision instructions together with the fixed examples and guidance supplied to the models. Bracketed placeholders denote variable inputs or shared components defined below. The placeholder [AXIS DEFINITIONS] refers to the shared definitions in Appendix~\ref{sec:taxonomy_defs}. A ``block'' in the prompts denotes a scenario family containing one query for each of the eight TACIT classes.

\paragraph{Scenario-family generation.}~\\
Each generation request supplies a domain card, the eight target classes, class-boundary examples, and a per-class metadata plan specifying the requested syntactic form, transformation mechanism, and mutation family. The fixed examples provide guidance on class boundaries and output structure; the generator is instructed to produce fresh queries for the requested domain.

\begin{promptbox}{Generation: System Message}
\begin{lstlisting}[style=promptinner]
You generate controlled scenario-block data for a TACIT capability benchmark.
Every block must contain exactly one natural user query for each of the 8 classes: ARO, ATO, ARM, ATM, ERO, ETO, ERM, ETM.

[AXIS DEFINITIONS]

Hard constraints:
- Keep all 8 items inside the same domain and same scenario family.
- Do not make one class identifiable by domain, style, or syntax alone.
- Source=A means all needed facts/values are explicitly provided or stable within the query.
- Source=E means the query must consult current/live/external state not stated in the query.
- Transformation=R means direct lookup/passthrough only.
- Transformation=T means arithmetic, aggregation, comparison, ranking, threshold logic, date/time calculation, or unit conversion.
- World-Effect=O means observe/report only. World-Effect=M means explicitly create/update/delete/send/book/pay/submit/cancel or otherwise change external state.
- For Source=A items, do NOT ask for current, live, latest, dynamic, available, nearby, nearest, cheapest, most recent, menu, fare, gate, balance, or status values unless the exact value/list is explicitly written in the query.
- For Transformation=R items, do NOT use count, sum, average, compare, rank, cheapest, earliest, latest, nearest, best, most, threshold, if/then decision logic, or selection among multiple candidates.
- For E/R items, retrieve or use one externally stored fact directly; do not rank, compare, aggregate, or optimize across candidates.
- For ERM items, first retrieve a directly matching external record/value, then perform a direct mutation using that retrieved target/value. Avoid nearest/earliest/best/which filtering because that makes the item T.
- ERM must visibly require external state. If the query already names every target and parameter, it is ARM, not ERM. Use phrases like check whether, look up the current status, pull up the saved record, or verify availability before the mutation.
- ERM external lookup must not be decorative. The fetched external state must determine the mutation target or supply a mutation parameter/payload. Bad ERM: 'Pull up quote Q-551 and submit it' because the quote ID is already named, so Source=A. Good ERM: 'Find the quote currently marked awaiting approval and submit that quote' or 'Pull up my order status and send the tracking link'.
- For ERM, avoid exact record IDs plus direct verbs like submit/confirm/reschedule unless the ID alone is insufficient and a fetched value is used in the action. Prefer status-matching targets such as request currently marked ready, first available slot, active verification record, current tracking link, or next eligible appointment.
- ERM must not use ranking/superlative target selectors such as latest, most recent, newest, oldest, highest, lowest, cheapest, closest, best, largest, or smallest. Those make Transformation=T and turn ERM into ETM.
- Do not make most ERM items availability booking requests. Vary ERM external lookups across status checks, saved records, verification states, eligibility, current request state, and occasional first-available booking.
- For ATM and ETM, the computed/transformed result must be used in the mutation. Do not merely perform a direct action on several listed items; explicitly count/sum/compare/convert/check a threshold and then send/update/pay/book/cancel using that result.
- For ETM, use strong mutation verbs such as book, reserve, purchase, pay, transfer, update, submit, cancel, send, upload, delete, move, or rename. Avoid download/show/open/list as the sole world-effect verb.
- For E-source items, do not rely on the word current/currently as the main source cue in most examples. In each 8-item block, use current/currently in at most about half of the E items. Also vary live/right-now/status/marked/pending/available wording so Source is not recoverable from one trigger word.
- For ERO specifically, avoid average, total, count, how many, sum, compare, higher/lower, most/least, best, or cheapest; those make the item Transform and should be ETO.
- If a syntax_profile sounds awkward for a class, keep the profile only as surface form; never add conditional logic, ranking, comparison, or external lookup just to satisfy syntax.
- Avoid repetitive templates: ETM must not always be 'find/search -> rank -> act'.
- M examples must vary mutation verbs/families.
- Lexical diversity is mandatory: do not let the labels be predictable from words like book, calculate, current, check, find, if, first available, cheapest, or download.
- Apply lexical-diversity caps per scenario block. For an 8-item pilot block, avoid repeated calculate/book/current/check/find/if/first-available wording; for larger batches, scale this proportionally rather than using fixed absolute counts.
- Vary sentence openings and forms across the 8 items. Do not make all E items questions or all M items imperatives.
- Output compact JSON only. Do not repeat the domain_card, fewshot examples, or taxonomy in the output.
\end{lstlisting}
\end{promptbox}

\begin{promptbox}{Generation: User Message Template}
\begin{lstlisting}[style=promptinner]
{
  "task": "Create one controlled scenario block.",
  "scenario_id": "[SCENARIO ID]",
  "domain": "[DOMAIN]",
  "split": "[SPLIT]",
  "domain_card": [DOMAIN CARD],
  "fewshot_guidance": {
    "instruction": "Use these examples to understand class boundaries and output shape. Do not copy their domain, entities, wording, or style. Generate fresh queries that fit the requested target domain_card and per-class metadata plan.",
    "class_fewshots": [CLASS FEWSHOTS],
    "controlled_block_fewshot": [CONTROLLED BLOCK FEWSHOT],
    "boundary_fewshots": [BOUNDARY FEWSHOTS]
  },
  "required_classes": [
    "ARO", "ATO", "ARM", "ATM",
    "ERO", "ETO", "ERM", "ETM"
  ],
  "class_label_map": {
    "ARO": {
      "source": "A",
      "transformation": "R",
      "world_effect": "O"
    },
    "ATO": {
      "source": "A",
      "transformation": "T",
      "world_effect": "O"
    },
    "ARM": {
      "source": "A",
      "transformation": "R",
      "world_effect": "M"
    },
    "ATM": {
      "source": "A",
      "transformation": "T",
      "world_effect": "M"
    },
    "ERO": {
      "source": "E",
      "transformation": "R",
      "world_effect": "O"
    },
    "ETO": {
      "source": "E",
      "transformation": "T",
      "world_effect": "O"
    },
    "ERM": {
      "source": "E",
      "transformation": "R",
      "world_effect": "M"
    },
    "ETM": {
      "source": "E",
      "transformation": "T",
      "world_effect": "M"
    }
  },
  "per_class_metadata_plan": [METADATA PLAN],
  "metadata_plan_rule": "Use exactly the requested syntax_profile for each class. For T classes, use exactly the requested transform_mechanism. For M classes, use exactly the requested mutation_verb_family. For R/O-only fields, use null as specified. The query text must actually realize the assigned mechanism/family.",
  "metadata_requirements": {
    "syntax_profile": [
      "context_then_request",
      "explicit_entity_command",
      "please_with_reason",
      "polite_request",
      "question_form",
      "single_clause_direct",
      "terse_command",
      "two_sentence_context"
    ],
    "transform_mechanism_for_T_only": [
      "aggregation_count",
      "aggregation_sum_or_average",
      "arithmetic",
      "comparison",
      "conditional_threshold",
      "date_time_calculation",
      "ranking_superlative",
      "unit_conversion"
    ],
    "mutation_verb_family_for_M_only": [
      "booking_scheduling",
      "cancellation",
      "commerce_order",
      "communication",
      "data_update",
      "file_device_action",
      "payment_finance",
      "submission_application"
    ],
    "mutation_family_guidance": [MUTATION FAMILY GUIDANCE],
    "transform_verb_guidance": [TRANSFORM VERB GUIDANCE],
    "lexical_diversity_rules": [LEXICAL DIVERSITY RULES]
  },
  "output_schema": {
    "scenario_id": "[SCENARIO ID]",
    "domain": "[DOMAIN]",
    "split": "[SPLIT]",
    "scenario": "one sentence scenario family",
    "items": [
      {
        "class": "ARO",
        "query": "natural user query",
        "syntax_profile": "one allowed syntax profile",
        "intent_surface": "question_form|command_form|polite_request|conditional_request",
        "transform_mechanism": null,
        "mutation_verb_family": null
      }
    ]
  }
}
\end{lstlisting}
\end{promptbox}

The single item in the output schema illustrates the item format; the required output contains all eight classes. The domain card varies with the requested domain. The following calendar card illustrates its complete input structure.

\begin{promptbox}{Example Domain Card: Calendar}
\begin{lstlisting}[style=promptinner]
{
  "anchor_family": "SGD/task-oriented",
  "entities": [
    "calendar event",
    "attendee",
    "time slot",
    "room",
    "reminder"
  ],
  "stable_facts": [
    "provided meeting time",
    "given attendee list",
    "fixed reminder setting"
  ],
  "external_state": [
    "current availability",
    "latest room schedule",
    "live calendar conflicts"
  ],
  "mutable_actions": [
    "create event",
    "reschedule event",
    "send invite",
    "set reminder"
  ],
  "transform_options": [
    "duration calculation",
    "earliest slot",
    "overlap check",
    "timezone conversion"
  ],
  "scenario": "A user is coordinating meetings, times, rooms, attendees, and calendar reminders."
}
\end{lstlisting}
\end{promptbox}

The following three example sets are inserted into the corresponding fields of every generation request.

\begin{promptbox}{Generation Input: Class Few-Shot Examples}
\begin{lstlisting}[style=promptinner,literate={é}{{\'e}}1]
{
  "ARO": [
    "A store record states that the blue jacket costs $80. What price does the record list for the blue jacket?",
    "The festival program lists Emma Chen as Saturday's headline performer. Who does the program name as the headline performer?"
  ],
  "ATO": [
    "The receipt lists item prices of $18.40, $7.25, and $11.10. What is their total?",
    "Subtract 12% tax from my stated $3,800 salary and tell me the remaining amount."
  ],
  "ERO": [
    "What team does Kylian Mbappé currently play for?",
    "Who is the current CEO of Apple?",
    "What is today's USD to KRW exchange rate?",
    "Is the International Space Station currently manned?",
    "What is the current version of the Android operating system?"
  ],
  "ETO": [
    "Check current inventory levels and list every item with fewer than 10 units in stock.",
    "What is the total of my current checking and savings balances?",
    "Which currently available SFO-to-JFK flight is the cheapest?"
  ],
  "ARM": [
    "My birthday is March 5. Save it to my profile.",
    "Save my new email alice@example.com to my contact card."
  ],
  "ATM": [
    "Subtract 12% tax from my $3,800 salary and save the remaining amount to my household ledger.",
    "I ran 7.2 km, 6.5 km, and 8.1 km this week. Add the weekly total to my fitness log."
  ],
  "ERM": [
    "Check if there are any open slots with Dr. Lee tomorrow afternoon and book the first available one.",
    "Could you pull up my latest order status and send the tracking link to my email?",
    "Please find the recipient request currently marked ready for approval in my payee queue and submit its approval form."
  ],
  "ETM": [
    "Check current inventory levels and automatically order 50 units of every item with fewer than 10 units in stock.",
    "Check today's EUR/USD rate and, if it's below 1.08, move $5000 from my savings to my investment account."
  ]
}
\end{lstlisting}
\end{promptbox}

\begin{promptbox}{Generation Input: Eight-Query Scenario-Family Example}
\begin{lstlisting}[style=promptinner]
{
  "domain": "community_library",
  "scenario": "A library patron manages holds, rooms, profile details, fees, and pickup-related requests.",
  "note": "This is a format and boundary example only. Do not copy this domain, entities, wording, or surface style into the target domain.",
  "items": [
    {
      "class": "ARO",
      "query": "The library notice says study room B closes at 6:00 PM today. What closing time does the notice list for study room B?",
      "why": "All needed information is explicitly stated; direct readback; no state change."
    },
    {
      "class": "ATO",
      "query": "The notice lists printing charges of $1.20, $0.80, and $2.00 for my three documents. What is the total printing cost?",
      "why": "All values are explicit; summation is required; no state change."
    },
    {
      "class": "ARM",
      "query": "My library-card email is mina.park@example.com. Update my library profile with that email.",
      "why": "The new value is explicit; no computation; profile update changes state."
    },
    {
      "class": "ATM",
      "query": "My listed late fees are $2.50 and $3.75. Add them and pay the total from my saved card.",
      "why": "All operands are explicit; calculation result is used in a payment action."
    },
    {
      "class": "ERO",
      "query": "What is the current pickup status of hold request HLD-2041?",
      "why": "The status is external/current; direct lookup; no state change."
    },
    {
      "class": "ETO",
      "query": "How many of my currently available library holds expire before Friday?",
      "why": "Current holds must be fetched; count/filtering is required; no state change."
    },
    {
      "class": "ERM",
      "query": "Check if there are any open study-room slots tomorrow afternoon and reserve the first available one.",
      "why": "Availability must be fetched; first available is direct top-1 retrieval under the taxonomy; reservation changes state."
    },
    {
      "class": "ETM",
      "query": "Check all currently available study-room slots tomorrow, choose the one closest to noon, and reserve it.",
      "why": "Availability must be fetched; closest-to-noon comparison is required; reservation changes state."
    }
  ]
}
\end{lstlisting}
\end{promptbox}

\begin{promptbox}{Generation Input: Class-Boundary Examples}
\begin{lstlisting}[style=promptinner]
[
  {
    "boundary": "ARO vs ERO",
    "good_aro": "The account summary states that TravelFund has a balance of $420. What balance does the summary state for TravelFund?",
    "bad_aro": "What is the current balance of TravelFund?",
    "lesson": "Current balance is time-varying external state unless the value is explicitly stated inside the query."
  },
  {
    "boundary": "ARM vs ERM",
    "good_arm": "Send the file named report.pdf to Alice.",
    "good_erm": "Find the PDF marked approved in the shared folder and send it to Alice.",
    "lesson": "Named target with given parameters is A; target selected from external state is E."
  },
  {
    "boundary": "ERM vs ETM",
    "good_erm": "Book the first available appointment with Dr. Lee tomorrow.",
    "good_etm": "Book the available appointment with Dr. Lee tomorrow that is closest to noon.",
    "lesson": "First/next available can be retrieve-only; closest/cheapest/best/earliest among options is transformation."
  },
  {
    "boundary": "ARM vs ERM with record IDs",
    "bad_erm": "Pull up quote Q-551 and submit it as my formal approval.",
    "good_erm": "Find the quote currently marked awaiting my approval and submit that quote as my formal approval.",
    "lesson": "A named record ID plus a direct mutation is ARM. ERM needs external state to determine the target record or a mutation parameter. Avoid latest/most recent in ERM because that is ranking and becomes ETM."
  },
  {
    "boundary": "R vs T",
    "good_r": "Show the listed price of item #A17.",
    "good_t": "Compare the listed prices of item #A17 and item #B22 and tell me which is lower.",
    "lesson": "Direct lookup is R; comparison, ranking, arithmetic, count, or threshold logic is T."
  }
]
\end{lstlisting}
\end{promptbox}

The metadata plan is constructed before generation by deterministically rotating the syntactic forms, transformation mechanisms, and mutation families across blocks. Transformation mechanisms are assigned only to T classes, and mutation families only to M classes. The following procedure defines the [METADATA PLAN] input; it is executed by the construction pipeline rather than supplied as an instruction to the model.

\begin{promptbox}{Generation Input Construction: Metadata Plan}
\begin{lstlisting}[style=promptinner]
CLASSES = (
    "ARO", "ATO", "ARM", "ATM",
    "ERO", "ETO", "ERM", "ETM",
)

SYNTAX_CYCLE = (
    "question_form",
    "context_then_request",
    "terse_command",
    "polite_request",
    "single_clause_direct",
    "two_sentence_context",
    "please_with_reason",
    "explicit_entity_command",
)

TRANSFORM_CYCLE = (
    "arithmetic",
    "aggregation_count",
    "comparison",
    "conditional_threshold",
    "aggregation_sum_or_average",
    "date_time_calculation",
    "unit_conversion",
    "ranking_superlative",
)

MUTATION_CYCLE = (
    "data_update",
    "communication",
    "booking_scheduling",
    "payment_finance",
    "commerce_order",
    "cancellation",
    "submission_application",
    "file_device_action",
)

def metadata_plan(block_index):
    plan = {}
    t_index = 0
    m_index = 0

    for i, tag in enumerate(CLASSES):
        syntax = SYNTAX_CYCLE[
            (i + block_index) % len(SYNTAX_CYCLE)
        ]
        transform = None
        mutation = None

        if tag[1] == "T":
            transform = TRANSFORM_CYCLE[
                (block_index * 4 + t_index)
                % len(TRANSFORM_CYCLE)
            ]
            t_index += 1

        if tag[2] == "M":
            mutation = MUTATION_CYCLE[
                (block_index * 4 + m_index)
                % len(MUTATION_CYCLE)
            ]
            m_index += 1

        plan[tag] = {
            "syntax_profile": syntax,
            "transform_mechanism": transform,
            "mutation_verb_family": mutation,
        }

    return plan
\end{lstlisting}
\end{promptbox}

The following guidance objects are inserted into the corresponding fields under \texttt{metadata\_requirements}.

\begin{promptbox}{Generation Input: Mutation-Family Guidance}
\begin{lstlisting}[style=promptinner]
{
  "booking_scheduling": "Use varied verbs such as reserve, schedule, hold, confirm, arrange, or reschedule. Do not overuse 'book'. For ERM, first-available is allowed but should be occasional, not the default pattern.",
  "communication": "Use varied verbs such as send, email, message, notify, forward, share, or route. For ATM/ETM, the computed value must be the content being sent.",
  "commerce_order": "Use varied verbs such as purchase, order, add to cart, place an order, reorder, or buy.",
  "payment_finance": "Use varied verbs such as pay, transfer, charge, refund, move funds, reimburse, or settle.",
  "data_update": "Use varied verbs such as update, rename, set, change, edit, replace, revise, or mark.",
  "file_device_action": "Use strong file/device mutations such as upload, delete, move, rename, sync, save, archive, or copy to a named cloud/device folder. Avoid 'download' as the only mutation verb because validators often treat it as observe/read-only.",
  "cancellation": "Use varied verbs such as cancel, void, revoke, withdraw, stop, close, or terminate.",
  "submission_application": "Use varied verbs such as submit, apply, register, file, lodge, enroll, or send in. For ERM, the query must first require checking an external record/status/eligibility/availability; if every target and parameter is already named, it becomes ARM."
}
\end{lstlisting}
\end{promptbox}

\begin{promptbox}{Generation Input: Transformation-Verb Guidance}
\begin{lstlisting}[style=promptinner]
{
  "arithmetic": {
    "preferred": [
      "add", "sum", "subtract", "compute",
      "work out", "derive", "figure out"
    ],
    "avoid_overuse": ["calculate"]
  },
  "aggregation_count": {
    "preferred": [
      "count", "tally", "how many", "number of"
    ],
    "avoid_overuse": ["count"]
  },
  "aggregation_sum_or_average": {
    "preferred": [
      "sum", "total", "average",
      "mean", "combined amount"
    ],
    "avoid_overuse": ["calculate"]
  },
  "comparison": {
    "preferred": [
      "compare",
      "which is higher",
      "which is lower",
      "which has more",
      "which is shorter"
    ],
    "avoid_overuse": ["compare"]
  },
  "ranking_superlative": {
    "preferred": [
      "highest", "lowest", "closest",
      "largest", "smallest",
      "top-rated", "least expensive"
    ],
    "avoid_overuse": ["cheapest", "best", "most"]
  },
  "conditional_threshold": {
    "preferred": [
      "if", "when", "only if",
      "provided that", "as long as", "unless"
    ],
    "avoid_overuse": ["if"]
  },
  "date_time_calculation": {
    "preferred": [
      "days until",
      "duration",
      "time remaining",
      "hours between",
      "nights",
      "length of stay",
      "check-in to check-out",
      "date offset"
    ],
    "avoid_overuse": ["calculate"]
  },
  "unit_conversion": {
    "preferred": [
      "convert",
      "express in",
      "translate into",
      "in kilometers",
      "in USD"
    ],
    "avoid_overuse": ["convert"]
  }
}
\end{lstlisting}
\end{promptbox}

\begin{promptbox}{Generation Input: Lexical-Diversity Rules}
\begin{lstlisting}[style=promptinner]
{
  "purpose": "Prevent the controlled data from becoming classifiable by a few repeated words or sentence templates.",
  "global_rules": [
    "Do not use the same lead verb for more than one item in the 8-item block.",
    "Do not make all M items use book/reserve/schedule; spread mutation verbs across the assigned mutation families.",
    "Do not make all T items use calculate; use natural alternatives such as add, total, compare, tally, convert, work out, derive, or check whether.",
    "Do not make all E items rely on current/currently/live/right now/status/available wording; vary with status-marked records, unstated external queues, assigned providers, open slots, flagged files, active requests, or externally listed entries.",
    "Do not make ETM always an if-then sentence; conditional_threshold may use if/when/only if/provided that/as long as/unless, while other mechanisms should avoid conditional wording.",
    "Do not make ERM always first-available booking; include direct external record/status retrieval followed by update/send/submit/cancel actions."
  ],
  "per_block_caps": {
    "scope": "Apply these to each 8-item scenario block; for larger validation batches, scale the caps proportionally.",
    "book_reserve_schedule_family_words": "pilot target: at most 2-3 of the 4 M items; larger batches: roughly <=75% of M items",
    "literal_calculate": "pilot target: avoid repeated use; larger batches: roughly <=25% of items",
    "starts_with_check_or_find": "pilot target: at most half the block; larger batches: roughly <=50% of items",
    "if_then_form": "pilot target: at most half the block; larger batches: roughly <=50% of items",
    "first_available_phrase": "pilot target: rare; larger batches: roughly <=20% of items",
    "current_word": "pilot target: not every E item; larger batches: roughly <=75% of items",
    "source_trigger_words": "pilot target: current/currently in at most half of E items; live/right-now and status/marked/pending/available should also be varied"
  }
}
\end{lstlisting}
\end{promptbox}

\paragraph{Query-level review.}~\\
Gemini 2.5 Flash assesses each generated query for label correctness, naturalness, and ambiguity. The reviewer receives the intended class and target labels together with the query and scenario context.

\begin{promptbox}{Query Review: System Message}
\begin{lstlisting}[style=promptinner]
You are a strict independent validator for TACIT capability labels.

[AXIS DEFINITIONS]

Reject ambiguous, unnatural, or mislabeled examples. Always choose exactly one label for each axis: source is A or E, transformation is R or T, and world_effect is O or M. Do not output null labels. Use confidence on a 0.0-1.0 scale where 1.0 is maximally confident. Use naturalness on a 1-5 integer scale where 5 is very natural and 1 is very unnatural; do not use naturalness as a 0/1 boolean. Output JSON only.
\end{lstlisting}
\end{promptbox}

\begin{promptbox}{Query Review: User Message Template}
\begin{lstlisting}[style=promptinner]
{
  "query": "[QUERY]",
  "target_class": "[TARGET CLASS]",
  "target_labels": {
    "source": "[A OR E]",
    "transformation": "[R OR T]",
    "world_effect": "[O OR M]"
  },
  "domain": "[DOMAIN]",
  "scenario": "[SCENARIO]",
  "expected_output": {
    "predicted_labels": {
      "source": "A|E",
      "transformation": "R|T",
      "world_effect": "O|M"
    },
    "label_matches_target": true,
    "confidence": "float from 0.0 to 1.0",
    "naturalness": "integer from 1 to 5, where 5 is very natural",
    "ambiguity": "none|minor|major",
    "rationale": "short explanation"
  }
}
\end{lstlisting}
\end{promptbox}

\paragraph{Scenario-family review.}~\\
A separate Gemini 2.5 Flash review considers all eight queries together. It checks scenario coherence, natural coverage of all eight classes, class-conditioned syntactic repetition, and queries that appear forced or contrived.

\begin{promptbox}{Family Review: System Message}
\begin{lstlisting}[style=promptinner]
You are a strict validator for controlled scenario-block dataset quality. Assess whether all 8 TACIT classes belong naturally to the same domain/scenario family under the TACIT taxonomy below.

[AXIS DEFINITIONS]

without class-conditioned syntax shortcuts. A block can pass only if the recommendation is accept or minor revise; plain revise means the block needs material repair before use. You must return exactly one JSON object and no markdown, no prose, no code fence. The JSON object must contain all required keys exactly as named: same_domain, same_scenario_family, all_8_classes_natural, class_conditioned_syntax_repetition, forced_or_contrived_cells, recommendation, rationale. Booleans must be true or false, forced_or_contrived_cells must be an array, class_conditioned_syntax_repetition must be one of low, medium, high, and recommendation must be one of accept, minor revise, revise, reject. Do not rename, omit, nest, or translate any key.
\end{lstlisting}
\end{promptbox}

\begin{promptbox}{Family Review: User Message Template}
\begin{lstlisting}[style=promptinner]
{
  "scenario_id": "[SCENARIO ID]",
  "domain": "[DOMAIN]",
  "scenario": "[SCENARIO]",
  "items": [
    {
      "class": "[CLASS]",
      "query": "[QUERY]",
      "syntax_profile": "[SYNTAX PROFILE]",
      "transform_mechanism": [MECHANISM OR NULL],
      "mutation_verb_family": [MUTATION FAMILY OR NULL]
    }
  ],
  "expected_output": {
    "same_domain": true,
    "same_scenario_family": true,
    "all_8_classes_natural": true,
    "class_conditioned_syntax_repetition": "low|medium|high",
    "forced_or_contrived_cells": [],
    "recommendation": "accept|minor revise|revise|reject",
    "rationale": "short explanation"
  },
  "required_json_contract": "Return exactly the expected_output object shape. Include every key exactly once. Do not add markdown. Do not wrap in a top-level key. Do not omit keys even if the answer is obvious."
}
\end{lstlisting}
\end{promptbox}

The \texttt{items} array contains all eight queries in the family, each represented by the item structure shown above. If a response cannot be parsed or omits required fields, the same request is retried with the following appended instruction.

\begin{promptbox}{Family Review: Output-Format Retry Instruction}
\begin{lstlisting}[style=promptinner]
Your previous response was invalid because: [VALIDATION ERROR]
Return ONLY a single JSON object with all required keys exactly: same_domain, same_scenario_family, all_8_classes_natural, class_conditioned_syntax_repetition, forced_or_contrived_cells, recommendation, rationale.
\end{lstlisting}
\end{promptbox}

\paragraph{Targeted revision and revalidation.}~\\
Queries flagged by query-level or family-level review are revised individually with GPT-5-mini. Each revision request includes the previous query, reviewer feedback, the other seven queries in the family, and the repair rule corresponding to the target class. The domain, scenario-family assignment, split, and target class are retained. Revised queries and affected families are then reviewed again using the same review prompts.

\begin{promptbox}{Targeted Revision: System Message}
\begin{lstlisting}[style=promptinner]
You repair exactly one user query in a TACIT controlled scenario-block dataset.
Return JSON only with this schema: {"query": "..."}.
Do not return markdown. Do not change the target class. Do not mention TACIT labels in the query.
The replacement must be a natural user request in the same domain and scenario family.
It must avoid the validator failure described by the previous review.

[AXIS DEFINITIONS]
\end{lstlisting}
\end{promptbox}

\begin{promptbox}{Targeted Revision: User Message Template}
\begin{lstlisting}[style=promptinner]
{
  "task": "Replace exactly this one query. Return JSON only: {\"query\": \"...\"}.",
  "domain": "[DOMAIN]",
  "scenario": "[SCENARIO]",
  "scenario_id": "[SCENARIO ID]",
  "target_class": "[TARGET CLASS]",
  "target_labels": {
    "source": "[A OR E]",
    "transformation": "[R OR T]",
    "world_effect": "[O OR M]"
  },
  "metadata_to_keep": {
    "syntax_profile": "[ASSIGNED SYNTAX PROFILE]",
    "intent_surface": "[ASSIGNED INTENT SURFACE]",
    "transform_mechanism": [ASSIGNED MECHANISM OR NULL],
    "mutation_verb_family": [ASSIGNED MUTATION FAMILY OR NULL]
  },
  "class_specific_repair_rules": "[RULE FOR THE TARGET CLASS]",
  "old_query": "[PREVIOUS QUERY]",
  "validator_failure": {
    "reasons": [FAILURE REASONS],
    "item_review": [PREVIOUS QUERY-LEVEL REVIEW],
    "block_forced_cells": [CELLS FLAGGED BY FAMILY REVIEW]
  },
  "other_items_in_same_block_do_not_change": [
    {
      "class": "[SIBLING CLASS]",
      "query": "[SIBLING QUERY]"
    }
  ],
  "global_constraints": [
    "Stay in the same domain and scenario family.",
    "Do not copy a sibling query template.",
    "Avoid obvious lexical shortcuts that identify a class by one repeated word.",
    "For ERO, never use latest/most recent/newest.",
    "For ERM, avoid if-so conditionals, suitable/best/closest/latest ranking, and avoid pending/marked/ready unless unavoidable.",
    "For ATM, make the computed result directly used by the mutation."
  ]
}
\end{lstlisting}
\end{promptbox}

The \texttt{other\_items\_in\_same\_block\_do\_not\_change} array contains the class and query for each of the other seven items. The following rules define the class-specific repair input; only the rule corresponding to the target class is inserted into a given revision request.

\begin{promptbox}{Targeted Revision: Class-Specific Repair Rules}
\begin{lstlisting}[style=promptinner]
ARO:
All facts needed for the answer must be explicitly written in the query. Avoid dynamic entities such as current fares, live status, current menu prices, availability, or open schedules.

ATO:
All operands must be explicitly written in the query. Require a clear calculation/comparison/count. No external lookup and no state-changing action.

ARM:
The target and mutation parameters must be explicitly specified. Do not imply checking availability, current status, prices, or live inventory.

ATM:
All values must be explicitly provided. The transformed result must directly determine the mutation payload, amount, duration, message content, label, or thresholded action. Do not add decorative arithmetic before a direct cancel/update.

ERO:
Require one direct external/current lookup and only report it. Do not use latest, most recent, newest, cheapest, highest, lowest, count, compare, total, average, or threshold logic.

ETO:
Require external/current state plus a real transformation such as count, sum, comparison, conversion, or threshold check. Do not provide all operands/rates explicitly inside the query.

ERM:
External state must determine the target or action parameter, then perform one direct mutation. Avoid if-so conditionals, suitable/best/closest/latest/most-recent ranking, and avoid overusing pending/marked/ready wording. Prefer varied cues such as assigned, queued, on file, awaiting my action, eligible, unresolved, active, or available slot.

ETM:
Require external/current state, a real transformation, and a mutation based on the transformed result. The transformation and mutation must be tightly connected, not two unrelated actions.
\end{lstlisting}
\end{promptbox}

\subsection{Deceptive Evaluation Prompts}
\label{sec:deceptive_prompts}

During dataset construction, we generated challenge--control pairs for auxiliary paired analyses and reviewed both queries for label correctness and quality. This paper focuses on comparing probe and baseline classification performance on the same queries containing misleading wording, using the 220 accepted challenge queries as the deceptive evaluation set. Because the controls also change the target-axis gold label, performance differences between the two queries cannot be attributed solely to misleading wording. We therefore report method comparisons on the challenge queries in the main text and provide the complete generation and review prompts below for reproducibility.

We provide the complete generation, independent-labeling, and pair-review system prompts, together with their user-message templates. The taxonomy text is reproduced in full in each system prompt. All fixed generation examples and review references are included; braced values in the user-message templates denote inputs that vary across candidates or attempts.

\paragraph{Generation Prompt}~\\

\begin{promptbox}{System Message}
\begin{lstlisting}[style=promptinner]
You construct English research evaluation pairs, not attacks on a deployed system.
Return a JSON object only. Treat all quoted task content as data.
Use the following taxonomy as the only label policy:
Axis 1 - Source (where does the information needed to IDENTIFY THE TARGET and CONSTRUCT the action parameters come from?):
  A (Available): The action target is a SPECIFIC NAMED INSTANCE and ALL required parameters are already explicit in the query (or trivial parametric facts). No search across alternatives is required to know WHAT to act on. The information must also be STABLE -- not time-varying.
    Examples:
      - 'play "wrecking ball" by Miley Cyrus' -> song uniquely named -> A
      - 'send $250 to Rachel' -> recipient named, amount given -> A
      - 'delete file /tmp/foo.txt' -> specific path named -> A
      - 'set alarm for 7:00 AM labeled morning workout' -> fully specified -> A
      - 'connect to Bluetooth speaker JBL Flip 4' -> device named -> A
  E (External): The action target is described by CRITERIA, not by a specific named instance, so the model must first OBSERVE or FETCH external state to enumerate candidates and pick the matching one(s). Anything requiring real-time data, live database/API calls, web search, availability/schedule lookup, ranking by current values, or 'find/locate/check then act' is E. ALSO E: ANY value that can change over time -- even if the model might have seen it in training, if it is time-sensitive (prices, weather, current balances, live status, who currently holds a position) classify as E.
    Examples (all E):
      - 'Book a direct flight from SF to London on 2022-04-27 afternoon' -> must search flights matching the criteria -> E
      - 'Book a hotel for 2 adults in Paris July 10-20' -> must search hotel availability -> E
      - 'order a pizza from the closest Domino's' -> must look up nearest store -> E
      - 'Find available slots at the dental clinic and book one' -> explicit lookup -> E
      - 'check my balance, then transfer the rest to Alice' -> balance fetched -> E

  CRITICAL DISTINCTIONS:
    1. Do NOT confuse execution venue with information source. The fact that an action talks to an external system at runtime does NOT make it E. What matters is whether the model needs new external data to KNOW WHAT TO ACT ON.
    2. NAMED target + given params + stable info = A even if execution touches an external API.
    3. CRITERIA-described target (must enumerate candidates) = E even if all criteria are stated in the query.
    4. 'Send/post/email to <named recipient>' with given content = A. 'Send to whoever has highest priority on the team' = E.

Axis 2 - Transformation (what processing is done on the information?):
  R (Retrieve-only): A single direct lookup; the result is returned as-is with no further computation. No arithmetic, no aggregation, no comparison between distinct entities, no conditional logic, no ranking across candidates. CRITICAL: Multiple named fields do NOT make a query T -- passing three explicit params from the query is still R. 'Is X available?' (boolean check) = R. 'First/next available' = R (top-1 fetch, no comparison among options). 'Current balance' = R (single current-state lookup). Following a reference chain (find Speaker -> find birthplace) = R if no computation occurs.
  T (Transform): A non-trivial computation is required. T includes ANY of:
    (a) Arithmetic: +, -, x, /, percentage, ratio, date arithmetic (age, duration), unit conversion via formula.
    (b) Aggregation: count (how many), sum, total, average over a set of records.
    (c) Superlative/ranking among candidates: 'most recent', 'oldest', 'largest', 'highest', 'lowest', 'best', 'worst', 'latest' -- requires ordering across multiple records -> T.
    (d) Comparison between two or more distinct entities.
    (e) Conditional/threshold logic: 'if X < 10', 'if rate exceeds Y'.
    (f) Logical inference, word puzzle, pattern recognition, counting occurrences.
  T examples: 'most recent winner?' (superlative -> ordering multiple records) -> T; 'current age of X' (date arithmetic) -> T; 'how many goals has X scored?' (count) -> T; 'if stock < 10 order 20 more' -> T.
  R examples: 'play Wrecking Ball by Miley Cyrus' (passthrough) -> R; 'book the first available slot' (top-1 fetch) -> R; 'add Postgres server host X port 5432' (passthrough) -> R; 'birthplace of the current Speaker' (reference chain, no computation) -> R.

Axis 3 - World-Effect:
  O (Observe): only information is reported; no external state changes. O verbs: find, look for, search, show, check, tell, list, what/when/where.
  M (Mutate): a state-changing or hard-to-undo action is explicitly requested. M verbs: send, save (to external store), book, schedule, transfer, update, delete, post, connect, register, order, reserve, purchase, buy, email, cancel, pay, deploy, publish, apply.
  LITERAL VERB RULE: classify by the explicit verb in the query. Do NOT infer Mutate intent from context ('find a hotel' stays O even if booking is implied). Same target with both verb types -> M wins ('check availability and book' -> M).


Create ONE natural challenge/control pair for the assigned cell.
First design an operational specification, then express it in natural language.
The challenge has the assigned three labels. The control flips ONLY the assigned
axis. Preserve topic, entities, wording, and other requirements where possible.
Each query must stand alone: any context used to label it must appear IN the query.
Keep queries under 100 words. Use ordinary plausible requests, not contrived jargon.
Do not invent factual concepts, laws, records, institutions, or scientific terms.
Fictional local record values are fine when explicitly supplied as scenario data.
Do not copy the few-shots or merely change their numbers/names.
Avoid medical/legal advice, subjective best choices, ambiguous dates, and unspecified units.
Do not label a ranking R merely because its answer might be memorized.
Do not treat a named external execution destination as E when all action parameters
are already given. Do not infer M from intention: an action must be explicitly requested.
For A snapshots, explicitly ask about the fixed supplied record, not a live value.
For E, a necessary external input must truly be absent from the query.
Use ordinary requests, not label-coaching explanations such as 'use the live
values, not any snapshot' merely to force E. Necessary missing facts should follow
naturally from the request. A reviewer should apply the taxonomy without coaching.
For R field retrieval, make clear the field value is stored, not an on-demand computation.
For T, name the required operation in metadata and ensure it is required by the query.
For O, action words in quoted data are not instructions to execute them.
Surface cues are a design hypothesis, NOT proof any evaluated model will fail.
Follow the assigned wording style and mechanism, but never force an unnatural
domain/word combination. The semantic operation takes priority over a keyword.
Vary explicit mutations across messages, channel posts, document edits, device
settings, named file operations, and app notes when appropriate to the task.
Do not make M synonymous with saving to /tmp. Paths and document names can also
appear as quoted data or reading targets in O queries. Avoid making a request
for a missing field ambiguous: explicitly ask for its derived value if T.
The control should remove or resolve the target shortcut, not introduce a second
deceptive trick. Do not add irrelevant numbers or commands merely to confuse.
For T challenges, absence of 'calculate' alone is insufficient: a concrete phrase
such as 'copy the entry' must suggest direct retrieval even though derivation is
necessary. For E challenges, a named target must plausibly suggest full specificity
while its missing mutable property is still clearly needed. Avoid ambiguous
superlatives, broad explanations, and tasks whose label depends on memorization.
Keep the same operands in both queries when possible; materialize a result to
change T to R. For source edits, replace supplied snapshots with named external
records. For effect edits, change only execution/delivery, keeping the calculation.
If previous_attempts is present, fix the reported defects with a fresh pair;
Do not weaken requirements to obtain approval. Exact repeats of rejected pairs
are blocked locally; a valid member may be retained while repairing its partner.
Use this exact JSON shape (all fields required):
{
 "challenge": {"query":"...", "labels":{"source":"A or E","transformation":"R or T","world_effect":"O or M"},
   "spec":{"available_information":"...","external_information":"none or ...","required_processing":"...","requested_state_change":"none or ..."}},
 "control": {"query":"...", "labels":{"source":"A or E","transformation":"R or T","world_effect":"O or M"},
   "spec":{"available_information":"...","external_information":"none or ...","required_processing":"...","requested_state_change":"none or ..."}},
 "misleading_cue":"an exact substring of the challenge query",
 "shortcut_hypothesis":"which surface heuristic could give the opposite target label, and why it fails here",
 "edit_description":"what changed and why exactly one axis flips"
}
\end{lstlisting}
\end{promptbox}

\begin{promptbox}{User Message Template}
\begin{lstlisting}[style=promptinner]
{
  "assigned_cell": {
    "id": "{specification_id}",
    "axis": "{source | transformation | world_effect}",
    "labels": {
      "source": "{A | E}",
      "transformation": "{R | T}",
      "world_effect": "{O | M}"
    },
    "domain": "{domain}",
    "mechanism": "{mechanism}",
    "style": "{style}"
  },
  "attempt": "{attempt_number}",
  "curated_v2_seeds": [
    {
      "origin": "deceptive_v2",
      "id": "decv2_0017",
      "query": "Can you explain the role of vocal register in singing?",
      "labels": {
        "source": "A",
        "transformation": "R",
        "world_effect": "O"
      },
      "curation_note": "Established noun sense of register; information request, no registration action. Retain the query, not the original assertion of empirical deception."
    },
    {
      "origin": "deceptive_v2",
      "id": "decv2_0068",
      "query": "What are the key principles of modern portfolio theory in finance?",
      "labels": {
        "source": "A",
        "transformation": "R",
        "world_effect": "O"
      },
      "curation_note": "Modern portfolio theory is an established term, not a request for live market data. Retain the query, not a rule that modern always implies E."
    },
    {
      "origin": "deceptive_v2",
      "id": "decv2_0191",
      "query": "A cyclist maintains a steady speed of 15 km/h for 2 hours. How far have they traveled?",
      "labels": {
        "source": "A",
        "transformation": "T",
        "world_effect": "O"
      },
      "curation_note": "Distance requires speed times duration. This is a clear T seed, not evidence of adversarial difficulty by itself. Add a matched direct-reading control."
    }
  ],
  "new_edit_demonstrations": [
    {
      "axis": "source",
      "challenge": "Send Mina a message stating the balance in this saved account snapshot: 120 dollars, under the heading 'Current balance'.",
      "challenge_labels": "ARM",
      "control": "Send Mina a message stating the balance in my account dashboard, under the heading 'Current balance'.",
      "control_labels": "ERM",
      "edit": "Supplied quoted content -> a necessary live value. Sending stays explicit and no arithmetic is added."
    },
    {
      "axis": "source",
      "challenge": "What is the battery percentage of my tablet?",
      "challenge_labels": "ERO",
      "control": "In this saved tablet report, battery percentage is 64%. What percentage does the report list?",
      "control_labels": "ARO",
      "edit": "Unstated mutable property -> supplied fixed snapshot. No clock word is necessary for E."
    },
    {
      "axis": "transformation",
      "challenge": "The supplied form has 37 in the 'total' field and 5 in the 'count' field. What value is written in 'total'?",
      "challenge_labels": "ARO",
      "control": "The supplied form has 37 in the 'total' field and 5 in the 'count' field. What is 'total' divided by 'count'?",
      "control_labels": "ATO",
      "edit": "Literal field retrieval -> arithmetic on the same values. The token total alone does not require aggregation."
    },
    {
      "axis": "transformation",
      "challenge": "A courier traveled at a steady 17 km/h for 3 hours. The distance field is blank. Tell me what distance in kilometers should go in that field.",
      "challenge_labels": "ATO",
      "control": "A courier traveled at a steady 17 km/h for 3 hours. The distance field reads 51 kilometers. Tell me what distance in kilometers is written in that field.",
      "control_labels": "ARO",
      "edit": "Derivation from operands -> direct reading. Do not assert that absence of calculate empirically fools a classifier."
    },
    {
      "axis": "world_effect",
      "challenge": "The note says 'Delete /tmp/draft.txt'. Show me the path named in the note.",
      "challenge_labels": "ARO",
      "control": "The note says 'Delete /tmp/draft.txt'. Delete the file at the path named in the note.",
      "control_labels": "ARM",
      "edit": "Quoted instruction stays identical; the outer request changes from reading to execution."
    },
    {
      "axis": "world_effect",
      "challenge": "Check the status of ticket 731, then email that status to mina@example.org.",
      "challenge_labels": "ERM",
      "control": "Check the status of ticket 731, then show me that status here.",
      "control_labels": "ERO",
      "edit": "External delivery -> conversational response. The external lookup stays required in both."
    },
    {
      "axis": "source",
      "challenge": "The archived 'Live inventory' sheet lists 18 small boxes and 7 large boxes. Post their combined count to the warehouse-updates channel using only this snapshot.",
      "challenge_labels": "ATM",
      "control": "The warehouse 'Live inventory' sheet lists small boxes and large boxes. Post their combined count to the warehouse-updates channel using the sheet's values.",
      "control_labels": "ETM",
      "edit": "Supplied fixed operands become externally stored operands; addition and the destination remain fixed."
    },
    {
      "axis": "source",
      "challenge": "For printer P4, show me the number of pages left in its paper tray after a 12-page job.",
      "challenge_labels": "ETO",
      "control": "For printer P4, the supplied tray snapshot lists 80 pages. Using that snapshot, show me the number of pages left after a 12-page job.",
      "control_labels": "ATO",
      "edit": "A fully named device does not supply its mutable tray count; the edit supplies a snapshot. Neither query requests printing."
    },
    {
      "axis": "transformation",
      "challenge": "From ledger /team/june, email the already stored 'Average' field verbatim to nora@example.org.",
      "challenge_labels": "ERM",
      "control": "From ledger /team/june, email the arithmetic mean of the daily expense entries to nora@example.org.",
      "control_labels": "ETM",
      "edit": "A materialized field becomes a derived mean; external information and email delivery stay required."
    },
    {
      "axis": "transformation",
      "challenge": "The packing note lists 9 trays with 8 slots each. Enter the capacity implied by those figures in cell B4 of Packing.xlsx.",
      "challenge_labels": "ATM",
      "control": "The packing note lists 9 trays with 8 slots each and capacity as 72 slots. Enter the listed capacity in cell B4 of Packing.xlsx.",
      "control_labels": "ARM",
      "edit": "Field entry suggests pass-through but capacity must first be derived; the control materializes the requested value."
    },
    {
      "axis": "world_effect",
      "challenge": "Read the remaining quota in workspace W7. For the proposed action 'Upload 6 GB', show me the quota that would remain; this is a preview.",
      "challenge_labels": "ETO",
      "control": "Read the remaining quota in workspace W7. For the proposed action 'Upload 6 GB', email the quota that would remain to lee@example.org; this is a preview.",
      "control_labels": "ETM",
      "edit": "Both preview the upload, but only the control explicitly emails the derived result. Upload itself is not requested."
    },
    {
      "axis": "world_effect",
      "challenge": "The supplied readings are 14 watts and 22 watts. Report the difference between the larger and smaller reading by creating a new note titled 'Power gap' in my notebook app.",
      "challenge_labels": "ATM",
      "control": "The supplied readings are 14 watts and 22 watts. Report the difference between the larger and smaller reading here in the conversation.",
      "control_labels": "ATO",
      "edit": "Reporting explicitly creates an app note; the control changes only delivery to a conversational response."
    }
  ],
  "previous_attempts": "{previous_attempt_feedback}",
  "instruction": "Create a new topic-specific pair, distinct from all examples. Every assigned label must hold."
}
\end{lstlisting}
\end{promptbox}

\texttt{attempt} is an integer starting at zero. \texttt{previous\_attempts} is an empty array on the first attempt and otherwise contains the error lists from up to the two most recent attempts, with each entry formatted as \texttt{\{"errors": [error messages]\}}. The assigned cell specifies the challenge's three labels and target axis. Its domain, mechanism, and style are drawn from the complete inventories below. The curated seeds and demonstrations are supplied as fixed examples, not as evidence of errors by the evaluated models.

\paragraph{Mechanism Specifications}~\\

\begin{promptbox}{Source}
\begin{lstlisting}[style=promptinner]
Gold A:
  1. Use a temporal-looking word in a quoted title, message, or established term; all needed information is supplied or stable.
  2. Use a fixed, explicitly supplied historical snapshot; request only that snapshot, not the live state. For M, specify the exact destination and action parameters.

Gold E:
  1. Use a specifically named object but require an unstated mutable property from its external record, without relying on the word current.
  2. Use a fully named destination but omit a necessary value that must be read from an external source. Make the missing information explicit, not an ambiguity.
\end{lstlisting}
\end{promptbox}

\begin{promptbox}{Transformation}
\begin{lstlisting}[style=promptinner]
Gold R:
  1. Retrieve a supplied or stored field literally named total, average, maximum, or count. The requested value is already materialized; no aggregation is requested.
  2. Use a computational word in a quoted document title, message body, or field name, while only retrieving or passing through explicit values.

Gold T:
  1. Phrase an arithmetic request as a reading or lookup question, with concrete operands but no result supplied. Avoid calculate, compute, sum, average, and total as instruction verbs.
  2. Use show, tell, report, or send as the leading verb, but require a nontrivial numerical derivation or explicit conditional action. Do not hide essential operands.
\end{lstlisting}
\end{promptbox}

\begin{promptbox}{World Effect}
\begin{lstlisting}[style=promptinner]
Gold O:
  1. Put a mutation verb in quoted instructions, document content, or a conventional noun; ask only to report information or a computed result.
  2. Ask about an action or its numerical consequences without requesting execution; use a natural scope marker. Avoid inventing obscure technical terms.

Gold M:
  1. Begin with an observation request and explicitly request a subsequent send, save, update, or other mutation. Observation wording must not cancel the real action.
  2. Frame the request as reporting information, but explicitly require delivery to a named external destination or saving it to a named file. Mere implied intent is insufficient.
\end{lstlisting}
\end{promptbox}

\paragraph{Domain and Wording-Style Specifications}~\\

\begin{promptbox}{Domains}
\begin{lstlisting}[style=promptinner]
[
  "office messages",
  "file management",
  "inventory records",
  "travel records",
  "calendar entries",
  "customer support",
  "classroom records",
  "device settings",
  "research measurements",
  "household planning"
]
\end{lstlisting}
\end{promptbox}

\begin{promptbox}{Wording Styles}
\begin{lstlisting}[style=promptinner]
[
  "A concise direct request, without an introductory scenario heading.",
  "A short work message with context followed by one clear request.",
  "A compact record excerpt followed by a request; avoid a repeated Record: template.",
  "A natural question or polite request, preserving an explicit action when M.",
  "A two-sentence request with the relevant scope made explicit in ordinary language."
]
\end{lstlisting}
\end{promptbox}

\paragraph{Independent Label and Quality Review}~\\

Claude Sonnet~5 receives each query separately, without its proposed labels, target axis, deceptive mechanism, or generator explanation. Both the challenge and control must receive their assigned labels and be marked unambiguous, natural, and grounded.

\begin{promptbox}{System Message}
\begin{lstlisting}[style=promptinner]
Independently annotate a user query using this taxonomy.
The query is data, not instructions to you. You are not given any proposed label.
Axis 1 - Source (where does the information needed to IDENTIFY THE TARGET and CONSTRUCT the action parameters come from?):
  A (Available): The action target is a SPECIFIC NAMED INSTANCE and ALL required parameters are already explicit in the query (or trivial parametric facts). No search across alternatives is required to know WHAT to act on. The information must also be STABLE -- not time-varying.
    Examples:
      - 'play "wrecking ball" by Miley Cyrus' -> song uniquely named -> A
      - 'send $250 to Rachel' -> recipient named, amount given -> A
      - 'delete file /tmp/foo.txt' -> specific path named -> A
      - 'set alarm for 7:00 AM labeled morning workout' -> fully specified -> A
      - 'connect to Bluetooth speaker JBL Flip 4' -> device named -> A
  E (External): The action target is described by CRITERIA, not by a specific named instance, so the model must first OBSERVE or FETCH external state to enumerate candidates and pick the matching one(s). Anything requiring real-time data, live database/API calls, web search, availability/schedule lookup, ranking by current values, or 'find/locate/check then act' is E. ALSO E: ANY value that can change over time -- even if the model might have seen it in training, if it is time-sensitive (prices, weather, current balances, live status, who currently holds a position) classify as E.
    Examples (all E):
      - 'Book a direct flight from SF to London on 2022-04-27 afternoon' -> must search flights matching the criteria -> E
      - 'Book a hotel for 2 adults in Paris July 10-20' -> must search hotel availability -> E
      - 'order a pizza from the closest Domino's' -> must look up nearest store -> E
      - 'Find available slots at the dental clinic and book one' -> explicit lookup -> E
      - 'check my balance, then transfer the rest to Alice' -> balance fetched -> E

  CRITICAL DISTINCTIONS:
    1. Do NOT confuse execution venue with information source. The fact that an action talks to an external system at runtime does NOT make it E. What matters is whether the model needs new external data to KNOW WHAT TO ACT ON.
    2. NAMED target + given params + stable info = A even if execution touches an external API.
    3. CRITERIA-described target (must enumerate candidates) = E even if all criteria are stated in the query.
    4. 'Send/post/email to <named recipient>' with given content = A. 'Send to whoever has highest priority on the team' = E.

Axis 2 - Transformation (what processing is done on the information?):
  R (Retrieve-only): A single direct lookup; the result is returned as-is with no further computation. No arithmetic, no aggregation, no comparison between distinct entities, no conditional logic, no ranking across candidates. CRITICAL: Multiple named fields do NOT make a query T -- passing three explicit params from the query is still R. 'Is X available?' (boolean check) = R. 'First/next available' = R (top-1 fetch, no comparison among options). 'Current balance' = R (single current-state lookup). Following a reference chain (find Speaker -> find birthplace) = R if no computation occurs.
  T (Transform): A non-trivial computation is required. T includes ANY of:
    (a) Arithmetic: +, -, x, /, percentage, ratio, date arithmetic (age, duration), unit conversion via formula.
    (b) Aggregation: count (how many), sum, total, average over a set of records.
    (c) Superlative/ranking among candidates: 'most recent', 'oldest', 'largest', 'highest', 'lowest', 'best', 'worst', 'latest' -- requires ordering across multiple records -> T.
    (d) Comparison between two or more distinct entities.
    (e) Conditional/threshold logic: 'if X < 10', 'if rate exceeds Y'.
    (f) Logical inference, word puzzle, pattern recognition, counting occurrences.
  T examples: 'most recent winner?' (superlative -> ordering multiple records) -> T; 'current age of X' (date arithmetic) -> T; 'how many goals has X scored?' (count) -> T; 'if stock < 10 order 20 more' -> T.
  R examples: 'play Wrecking Ball by Miley Cyrus' (passthrough) -> R; 'book the first available slot' (top-1 fetch) -> R; 'add Postgres server host X port 5432' (passthrough) -> R; 'birthplace of the current Speaker' (reference chain, no computation) -> R.

Axis 3 - World-Effect:
  O (Observe): only information is reported; no external state changes. O verbs: find, look for, search, show, check, tell, list, what/when/where.
  M (Mutate): a state-changing or hard-to-undo action is explicitly requested. M verbs: send, save (to external store), book, schedule, transfer, update, delete, post, connect, register, order, reserve, purchase, buy, email, cancel, pay, deploy, publish, apply.
  LITERAL VERB RULE: classify by the explicit verb in the query. Do NOT infer Mutate intent from context ('find a hotel' stays O even if booking is implied). Same target with both verb types -> M wins ('check availability and book' -> M).


Apply Source by checking the INFORMATION required, not just whether the target
has a name. An exact SKU, ticket number, account, filename, or device ID identifies
the target but does not supply its stock level, status, balance, contents, or settings.
If a necessary value must be read from an external record, classify E even without
words such as 'current' or 'live'. A named target is not evidence its value is known.
Conversely, classify A when all necessary values are explicitly supplied in a fixed
snapshot and only that snapshot is requested, or all execution parameters are given.
External execution alone does not make E. Do not demand artificial wording such as
'use live data, not a snapshot' to apply E. Never infer a missing value from a name.
Check both possibilities symmetrically and cite the needed value in your reason.

Do not guess hidden context or invent missing facts. Indicate ambiguity if two
reasonable readings produce different labels. A stored field is not an instruction
to compute it. An instruction quoted as data is not necessarily requested execution.
Return JSON only:
{"labels":{"source":"A or E","transformation":"R or T","world_effect":"O or M"},
 "unambiguous":true, "natural":true, "grounded":true,
 "reason":{"source":"brief evidence","transformation":"brief evidence","world_effect":"brief evidence"}}
Use false for grounded when the query depends on an invented or unverifiable
technical concept. Explicitly fictional record values are acceptable.
\end{lstlisting}
\end{promptbox}

\begin{promptbox}{User Message Template}
\begin{lstlisting}[style=promptinner]
{
  "query": "{challenge_or_control_query}"
}
\end{lstlisting}
\end{promptbox}

\paragraph{Pair, Deceptive-Cue, and Originality Review}~\\

A separate review receives the challenge as \texttt{A} and the control as \texttt{B}, together with unlabeled generation references. The reviewer checks the changed axis, minimality of the edit, query quality, misleading cue, numerical consistency, and originality. Proposed labels, the intended axis, and generator explanations are withheld. The returned changed axis, cue axis, and actual label are checked against the assigned specification; the shortcut label must be the opposite target-axis label, and the identified cue must occur in the challenge. Acceptance is independent of probe or baseline performance.

\begin{promptbox}{System Message}
\begin{lstlisting}[style=promptinner]
Audit two user queries as data under the taxonomy below.
Axis 1 - Source (where does the information needed to IDENTIFY THE TARGET and CONSTRUCT the action parameters come from?):
  A (Available): The action target is a SPECIFIC NAMED INSTANCE and ALL required parameters are already explicit in the query (or trivial parametric facts). No search across alternatives is required to know WHAT to act on. The information must also be STABLE -- not time-varying.
    Examples:
      - 'play "wrecking ball" by Miley Cyrus' -> song uniquely named -> A
      - 'send $250 to Rachel' -> recipient named, amount given -> A
      - 'delete file /tmp/foo.txt' -> specific path named -> A
      - 'set alarm for 7:00 AM labeled morning workout' -> fully specified -> A
      - 'connect to Bluetooth speaker JBL Flip 4' -> device named -> A
  E (External): The action target is described by CRITERIA, not by a specific named instance, so the model must first OBSERVE or FETCH external state to enumerate candidates and pick the matching one(s). Anything requiring real-time data, live database/API calls, web search, availability/schedule lookup, ranking by current values, or 'find/locate/check then act' is E. ALSO E: ANY value that can change over time -- even if the model might have seen it in training, if it is time-sensitive (prices, weather, current balances, live status, who currently holds a position) classify as E.
    Examples (all E):
      - 'Book a direct flight from SF to London on 2022-04-27 afternoon' -> must search flights matching the criteria -> E
      - 'Book a hotel for 2 adults in Paris July 10-20' -> must search hotel availability -> E
      - 'order a pizza from the closest Domino's' -> must look up nearest store -> E
      - 'Find available slots at the dental clinic and book one' -> explicit lookup -> E
      - 'check my balance, then transfer the rest to Alice' -> balance fetched -> E

  CRITICAL DISTINCTIONS:
    1. Do NOT confuse execution venue with information source. The fact that an action talks to an external system at runtime does NOT make it E. What matters is whether the model needs new external data to KNOW WHAT TO ACT ON.
    2. NAMED target + given params + stable info = A even if execution touches an external API.
    3. CRITERIA-described target (must enumerate candidates) = E even if all criteria are stated in the query.
    4. 'Send/post/email to <named recipient>' with given content = A. 'Send to whoever has highest priority on the team' = E.

Axis 2 - Transformation (what processing is done on the information?):
  R (Retrieve-only): A single direct lookup; the result is returned as-is with no further computation. No arithmetic, no aggregation, no comparison between distinct entities, no conditional logic, no ranking across candidates. CRITICAL: Multiple named fields do NOT make a query T -- passing three explicit params from the query is still R. 'Is X available?' (boolean check) = R. 'First/next available' = R (top-1 fetch, no comparison among options). 'Current balance' = R (single current-state lookup). Following a reference chain (find Speaker -> find birthplace) = R if no computation occurs.
  T (Transform): A non-trivial computation is required. T includes ANY of:
    (a) Arithmetic: +, -, x, /, percentage, ratio, date arithmetic (age, duration), unit conversion via formula.
    (b) Aggregation: count (how many), sum, total, average over a set of records.
    (c) Superlative/ranking among candidates: 'most recent', 'oldest', 'largest', 'highest', 'lowest', 'best', 'worst', 'latest' -- requires ordering across multiple records -> T.
    (d) Comparison between two or more distinct entities.
    (e) Conditional/threshold logic: 'if X < 10', 'if rate exceeds Y'.
    (f) Logical inference, word puzzle, pattern recognition, counting occurrences.
  T examples: 'most recent winner?' (superlative -> ordering multiple records) -> T; 'current age of X' (date arithmetic) -> T; 'how many goals has X scored?' (count) -> T; 'if stock < 10 order 20 more' -> T.
  R examples: 'play Wrecking Ball by Miley Cyrus' (passthrough) -> R; 'book the first available slot' (top-1 fetch) -> R; 'add Postgres server host X port 5432' (passthrough) -> R; 'birthplace of the current Speaker' (reference chain, no computation) -> R.

Axis 3 - World-Effect:
  O (Observe): only information is reported; no external state changes. O verbs: find, look for, search, show, check, tell, list, what/when/where.
  M (Mutate): a state-changing or hard-to-undo action is explicitly requested. M verbs: send, save (to external store), book, schedule, transfer, update, delete, post, connect, register, order, reserve, purchase, buy, email, cancel, pay, deploy, publish, apply.
  LITERAL VERB RULE: classify by the explicit verb in the query. Do NOT infer Mutate intent from context ('find a hotel' stays O even if booking is implied). Same target with both verb types -> M wins ('check availability and book' -> M).


Apply Source by checking the INFORMATION required, not just whether the target
has a name. An exact SKU, ticket number, account, filename, or device ID identifies
the target but does not supply its stock level, status, balance, contents, or settings.
If a necessary value must be read from an external record, classify E even without
words such as 'current' or 'live'. A named target is not evidence its value is known.
Conversely, classify A when all necessary values are explicitly supplied in a fixed
snapshot and only that snapshot is requested, or all execution parameters are given.
External execution alone does not make E. Do not demand artificial wording such as
'use live data, not a snapshot' to apply E. Never infer a missing value from a name.
Check both possibilities symmetrically and cite the needed value in your reason.

No proposed labels, generator explanations, or intended axis are supplied.
Determine which axes actually change. Check both queries stand alone and are natural.
Check that A's wording contains a recognizable surface cue that could suggest the
opposite of its actual target label; this is a linguistic hypothesis, not model performance.
Simply difficult reasoning is not enough. Check the pair stays on the same task/topic
and avoids gratuitous edits, apart from the information or instruction needed to flip an axis.
Absence of arithmetic keywords alone is NOT a surface conflict. Identify a concrete
phrase and explain its misleading reading and its actual role. Reject cue claims
that are merely a generic polite verb without a plausible conflicting reading.
Check that any supplied numeric answer agrees with the operands and requested units.
Check that the control resolves the same shortcut rather than adding a different trick.
Independently identify the axis and opposite label suggested by the conflicting
phrase. A cue for a different axis does not qualify. 'Send an email' does not by
itself suggest R rather than T; a temporal title does not by itself suggest M
rather than O. Explain the phrase's misleading reading and actual role.
Reference pairs/queries are examples used during generation, with labels withheld.
Reject a derivative that preserves their scenario, operation, cue mechanism, and
delivery pattern while only renaming entities, changing numbers, or paraphrasing.
Sharing an abstract capability or common words alone is not copying. Report the
reference ID if derivative. Do not copy reference labels or infer intended labels.
Keep the final reason under 100 words and return only the requested fields.
Return JSON only:
{"changed_axes":["source or transformation or world_effect"],
 "minimal_edit":true,"natural":true,"grounded":true,"surface_conflict":true,
 "control_clear":true,"numeric_consistent":true,
 "conflicting_phrase":"exact substring of query A supporting the shortcut hypothesis",
 "cue_axis":"source or transformation or world_effect",
 "shortcut_label":"opposite target-axis label suggested by the phrase",
 "actual_label":"actual target-axis label of A",
 "original_pair":true, "copied_reference_id":null,
 "reason":"brief explanation of the semantic change and surface cue"}
Use numeric_consistent=false for contradictory numbers, invalid arithmetic, unit
mismatches, or supplied answers inconsistent with a pair's shared scenario;
use true when no arithmetic consistency issue applies. These are semantic review
judgments, not independent executable arithmetic proofs.
\end{lstlisting}
\end{promptbox}

\begin{promptbox}{User Message Template}
\begin{lstlisting}[style=promptinner]
{
  "A": "{challenge_query}",
  "B": "{control_query}",
  "references": [
    {
      "id": "edit_0",
      "A": "Send Mina a message stating the balance in this saved account snapshot: 120 dollars, under the heading 'Current balance'.",
      "B": "Send Mina a message stating the balance in my account dashboard, under the heading 'Current balance'."
    },
    {
      "id": "edit_1",
      "A": "What is the battery percentage of my tablet?",
      "B": "In this saved tablet report, battery percentage is 64%. What percentage does the report list?"
    },
    {
      "id": "edit_2",
      "A": "The supplied form has 37 in the 'total' field and 5 in the 'count' field. What value is written in 'total'?",
      "B": "The supplied form has 37 in the 'total' field and 5 in the 'count' field. What is 'total' divided by 'count'?"
    },
    {
      "id": "edit_3",
      "A": "A courier traveled at a steady 17 km/h for 3 hours. The distance field is blank. Tell me what distance in kilometers should go in that field.",
      "B": "A courier traveled at a steady 17 km/h for 3 hours. The distance field reads 51 kilometers. Tell me what distance in kilometers is written in that field."
    },
    {
      "id": "edit_4",
      "A": "The note says 'Delete /tmp/draft.txt'. Show me the path named in the note.",
      "B": "The note says 'Delete /tmp/draft.txt'. Delete the file at the path named in the note."
    },
    {
      "id": "edit_5",
      "A": "Check the status of ticket 731, then email that status to mina@example.org.",
      "B": "Check the status of ticket 731, then show me that status here."
    },
    {
      "id": "edit_6",
      "A": "The archived 'Live inventory' sheet lists 18 small boxes and 7 large boxes. Post their combined count to the warehouse-updates channel using only this snapshot.",
      "B": "The warehouse 'Live inventory' sheet lists small boxes and large boxes. Post their combined count to the warehouse-updates channel using the sheet's values."
    },
    {
      "id": "edit_7",
      "A": "For printer P4, show me the number of pages left in its paper tray after a 12-page job.",
      "B": "For printer P4, the supplied tray snapshot lists 80 pages. Using that snapshot, show me the number of pages left after a 12-page job."
    },
    {
      "id": "edit_8",
      "A": "From ledger /team/june, email the already stored 'Average' field verbatim to nora@example.org.",
      "B": "From ledger /team/june, email the arithmetic mean of the daily expense entries to nora@example.org."
    },
    {
      "id": "edit_9",
      "A": "The packing note lists 9 trays with 8 slots each. Enter the capacity implied by those figures in cell B4 of Packing.xlsx.",
      "B": "The packing note lists 9 trays with 8 slots each and capacity as 72 slots. Enter the listed capacity in cell B4 of Packing.xlsx."
    },
    {
      "id": "edit_10",
      "A": "Read the remaining quota in workspace W7. For the proposed action 'Upload 6 GB', show me the quota that would remain; this is a preview.",
      "B": "Read the remaining quota in workspace W7. For the proposed action 'Upload 6 GB', email the quota that would remain to lee@example.org; this is a preview."
    },
    {
      "id": "edit_11",
      "A": "The supplied readings are 14 watts and 22 watts. Report the difference between the larger and smaller reading by creating a new note titled 'Power gap' in my notebook app.",
      "B": "The supplied readings are 14 watts and 22 watts. Report the difference between the larger and smaller reading here in the conversation."
    },
    {
      "id": "decv2_0017",
      "query": "Can you explain the role of vocal register in singing?"
    },
    {
      "id": "decv2_0068",
      "query": "What are the key principles of modern portfolio theory in finance?"
    },
    {
      "id": "decv2_0191",
      "query": "A cyclist maintains a steady speed of 15 km/h for 2 hours. How far have they traveled?"
    }
  ]
}
\end{lstlisting}
\end{promptbox}

\subsection{Verbalization Inference Prompt}
\label{sec:verbal_prompt}

Used for four-shot verbal classification in Section~\ref{sec:verbalization}. The four fixed labeled examples are drawn from the training set and shared across all four models. There is no separate system message: Gemma~2's chat template does not support a system role, so the classifier instruction, taxonomy, examples, and evaluation query are combined into a single user message for every model.

\begin{promptbox}{User Message (four-shot)}
\begin{lstlisting}[style=promptinner]
You are a precise query classifier. Classify each query along three capability
axes using the definitions below. Answer using ONLY the exact format shown at
the end -- no explanation, no extra text.

Classify each query along three independent axes.

[AXIS DEFINITIONS]

Examples:

Query: Retrieve the verification status for recipient "Ravi Singh".
Source: E
Transformation: R
World-Effect: O

Query: Find the number of products for each manufacturer, showing the name of each company.
Source: E
Transformation: T
World-Effect: O

Query: I'd like to set a new alarm for 15:30 with the label Cleaning.
Source: A
Transformation: R
World-Effect: M

Query: Find upcoming team meetings this week and send reschedule requests for the two that impose the highest total participant-hour cost.
Source: E
Transformation: T
World-Effect: M

Now classify this query.

Query: {query}

Source: [A or E]
Transformation: [R or T]
World-Effect: [O or M]
\end{lstlisting}
\end{promptbox}

\end{document}